\documentclass{article} 
\usepackage{iclr2027_conference,times}

\usepackage{amsmath,amsfonts,bm}

\def\eqref#1{equation~\ref{#1}}

\def\1{\bm{1}}

\DeclareMathAlphabet{\mathsfit}{\encodingdefault}{\sfdefault}{m}{sl}
\SetMathAlphabet{\mathsfit}{bold}{\encodingdefault}{\sfdefault}{bx}{n}

\usepackage{hyperref}
\definecolor{BlueBlack}{RGB}{36, 113, 163}
\hypersetup{
    colorlinks,
    linkcolor={red},
    citecolor={BlueBlack},
    urlcolor={BlueBlack}
}

\usepackage{multirow}
\usepackage{array}
\usepackage{url}
\usepackage{booktabs}
\usepackage{longtable}
\usepackage{fontawesome5}
\usepackage{graphicx}
\usepackage[dvipsnames]{xcolor}
\usepackage[table]{xcolor} 
\usepackage{enumitem}
\usepackage{tabularx}
\usepackage[nameinlink]{cleveref}
\usepackage{changepage}
\usepackage{makecell}
\usepackage{xcolor}
\usepackage{pifont}
\usepackage{subcaption}

\PassOptionsToPackage{dvipsnames}{xcolor}
\usepackage[normalem]{ulem}
\usepackage{bbm}
\usepackage{anyfontsize}
\usepackage{xspace}
\usepackage[export]{adjustbox}
\usepackage{etoolbox}
\usepackage{wrapfig}
\usepackage[bb=dsserif]{mathalpha}
\usepackage{mathtools}

\usepackage{arydshln}
\usepackage{makecell}
\usepackage{lipsum}
\usepackage{soul}
\usepackage{tcolorbox}
\tcbuselibrary{breakable, skins}

\newcommand{\ours}{\texttt{RoboWorM}\xspace}
\newcommand{\myparagraph}[1]{\vspace{2pt}\noindent{\bf #1}}
\newcommand{\cmark}{\textcolor{green!60!black}{\ding{51}}}
\newcommand{\xmark}{\textcolor{red!75!black}{\ding{55}}}

\definecolor{oldgray}{RGB}{140,140,140}
\definecolor{newblue}{RGB}{0,70,180}
\definecolor{commentorange}{RGB}{220,120,0}

\newif\ifrevision
\revisionfalse           

\ifrevision
    \newcommand{\oldtext}[1]{{\color{oldgray}#1}}
    \newcommand{\newtext}[1]{{\color{newblue}#1}}
    
\else
    \newcommand{\oldtext}[1]{}
    \newcommand{\newtext}[1]{#1}
    
\fi

\usepackage{titlesec}

\titlespacing*{\section}{0pt}{6pt}{1pt}       
\titlespacing*{\subsection}{0pt}{6pt}{1pt}    
\titlespacing*{\subsubsection}{0pt}{6pt}{1pt} 

\title{\vspace{-1.2cm}\ours: Evaluating movement realism,\\ consistency and acceleration affects \\in Embodied Video \underline{Wor}ld \underline{M}odels \vspace{-0.5cm}}

\author{\textbf{Doanh Le Thien$^{1*}$, Viet-Thanh Nguyen$^{2*}$, Pham Tri Quang$^{1*}$, Nhi Pham$^{3}$, Christopher Wewer$^{3}$,}\\[0.05in]
\textbf{Minh-Loi Nguyen$^{1}$, Hoang H. Tran$^{1}$, Tho Le$^{2}$, Tung M. Luu$^{4}$, Dung D. Le$^{1}$, Vu N. Duong$^{1}$,}\\[0.05in]
\textbf{Jonas Fischer$^{3}$, Ngan Le$^{5}$, An Thai Le$^{1,2,12}$, Daniel Sonntag$^{6,7}$, Gim Hee Lee$^{8}$, James Zou$^{9}$,}\\[0.05in]
\textbf{Mathias Niepert$^{10,11}$, Jan Peters$^{6,12}$, Duy M. H. Nguyen$^{6,10,11\dagger}$, Ngo Anh Vien$^{1,2\dagger}$}
\\[0.2in]
$^{1}$Center for AI Research, VinUniversity\quad
$^{2}$VinRobotics\quad
$^{3}$MPI-INF \quad
$^{4}$KAIST \quad
$^{5}$University of Arkansas \\
$^{6}$German Research Center for Artificial Intelligence \quad
$^{7}$Oldenburg University \quad
$^{8}$NUS\\
$^{9}$Stanford University \quad
$^{10}$Universit\"at Stuttgart \quad
$^{11}$IMPRS-IS \quad
$^{12}$Technische Universit\"at Darmstadt \\
$^{*}$Equal Contributors \quad $^{\dagger}$Project Leads
}
\iclrfinalcopy %
\begin{document}

\maketitle

{
\begin{center}
\vspace{-0.3in}
\includegraphics[width=\linewidth]{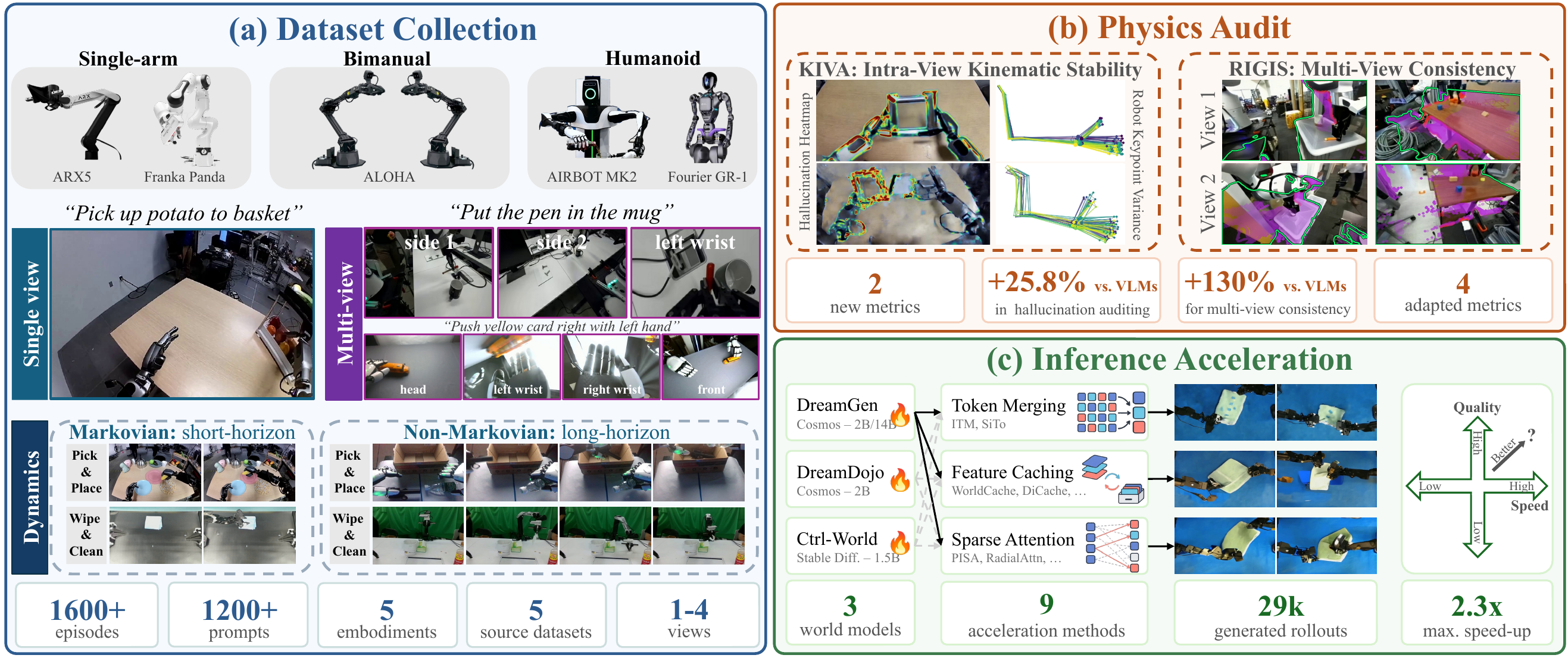}
\captionsetup{type=figure}
\vspace{-0.2in}
\caption{
\oldtext{We introduce \ours{}, a benchmark that evaluates whether robotic video world models remain physically consistent as they become faster.
\textbf{(a) Dataset Collection:} We cover 1,200+ prompts and 1,600+ episodes across 5 embodiments, 5 source datasets, 1--4 views, and both short- and long-horizon tasks.
\textbf{(b) Physics Audit:} Two VLM-free metrics measure complementary physical properties: \textbf{KIVA} for intra-view kinematic stability and embodiment hallucinations, and \textbf{RIGIS} for multi-view geometric consistency.
\textbf{(c) Inference Acceleration:} We evaluate 3 world models across 9 acceleration methods, achieving $2.3\times$ max speedup with comparable physical fidelity.}
\newtext{Overview of \textbf{\ours}, a deployment-aware benchmark auditing the physical fidelity and computational efficiency of robotic video world models. \textbf{(a) Dataset Collection:} 1,600+ real-world episodes and 1,200+ prompts across 5 embodiments (single-arm, bimanual, humanoid), 1–4 views, and diverse task horizons. \textbf{(b) Physics Audit:} We propose two deterministic, VLM-free metrics leveraging hardware priors: \textbf{KIVA} for kinematic stability and \textbf{RIGIS} for 3D geometric consistency. \textbf{(c) Inference Acceleration:} we evaluate 3 world models across 9 acceleration methods (29k rollouts), examining how speed-ups compromise deployment-critical physical constraints. Project page: \url{https://roboworm.github.io/roboworm/}.}
\vspace{0.2in}
}
\label{fig:teaser}
\end{center}
}

\begin{abstract}
\oldtext{In robotic settings with pre-defined embodiment setups, video world models have become increasingly useful as action planners and simulators to mitigate physical data scarcity in robot learning loops. However, their high generation latency often requires the use of acceleration inference methods, thus trading physical fidelity for efficiency. Existing benchmarks fail to capture these structural failure modes in open-loop generation: standard metrics enforce rigid references, while ungrounded vision-language models lack kinematic and cross-view geometric awareness. To address this gap, we present \ours{}, an orthogonal direction for measuring physical consistency without requiring a failure taxonomy or reference future, leveraging intrinsic robotic invariants to systematically audit structural and spatiotemporal fidelity. \ours{} aggregates $1600+$ episodes across $5$ distinct embodiments spanning $3$ major robot morphologies: single-arm, dual-arm, and humanoid systems, and covers both Markovian and non-Markovian task properties under single- and multi-view camera setups. We introduce two new metrics to complement existing benchmarks, namely (i) KIVA for intra-view kinematic stability, and (ii) RIGIS for multi-view geometric consistency. Compared to the strongest of 4 zero-shot frontier VLMs, KIVA yields a $+21.5\%$ relative improvement in segment-level overlap with human annotations in hallucination auditing, while RIGIS achieves a $+138\%$ relatively higher Spearman correlation with human preferences. \ours{} also provides unified benchmark to accelerate robotic world models while preserving physical reality, auditing 3 state-of-the-art video world models across 9 acceleration techniques. We will make our benchmark and code publicly available upon acceptance for further research.}
\newtext{Video world models (VWMs) are rapidly emerging as scalable neural simulators for robotic manipulation. However, most benchmarks only evaluate them as open-world video generators, using Vision-Language Models (VLMs) that lack strict 3D spatial and kinematic reasoning. To bridge this gap, we introduce \ours{}, a comprehensive, \emph{deployment-aware} benchmark that leverages known hardware priors such as robot kinematic trees and fixed camera configurations, to systematically examine structural and spatiotemporal fidelity. \ours{} curates 1,605 real-world episodes across single-arm, bimanual, and humanoid morphologies, spanning both Markovian and long-horizon non-Markovian tasks under single- and multi-view setups. To explicitly measure physical consistency, we propose two deterministic, VLM-free metrics: \textbf{KIVA} for intra-view kinematic stability, and \textbf{RIGIS} for multi-view geometric consistency. By exploiting intrinsic robotic invariants, KIVA improves hallucination detection alignment with human annotations by 25.8\% relative to the strongest frontier VLMs, while RIGIS achieves a 130\% higher Spearman correlation with human preferences. Furthermore, since real-world deployment requires low inference latency, we utilize \ours{} to conduct the first large-scale audit of the quality-efficiency trade-off. Evaluating 9 state-of-the-art acceleration techniques across 3 leading world models (over 29,000 generated rollouts), we demonstrate how these acceleration methods can retain superficial visual quality while silently compromising deployment-critical physical constraints. Using our metrics, we achieve 2.3$\times$ maximum inference speed-up with minimal quality loss.}
\end{abstract}

\vspace{-0.1in}
\section{Introduction}
Video world models (VWMs) have recently emerged as scalable neural simulators for robotic manipulation, mitigating physical data scarcity by generating synthetic trajectories, ranking policies, and enabling model-based planning~\citep{dreamgen,gao2026dreamdojo,guo2026ctrl}. However, the prevailing evaluation paradigms treat these models merely as open-world video generators. General-purpose benchmarks evaluate perceptual quality and text alignment~\citep{liu2023evalcrafter,huang2024vbench,li2026worldmodelbench}, while recent robotics-aligned benchmarks rely heavily on Vision-Language Models (VLMs) to score task plausibility~\citep{jiang2026robowm,li2026robotrustbench,yang2026mirabench,nguyen2026robogaze}. While valuable for assessing high-level semantic alignment, VLM evaluators suffer from optimism bias, struggle with strict 3D geometric reasoning, and evaluate physical AI in a vacuum. A generated robot video is not just a moving picture; it represents a physical system in motion. In reality, a robot's mechanical arms cannot arbitrarily change length, its gripper cannot float disconnected in mid-air, and multiple cameras viewing the same workspace must capture the exact same 3D layout.

To bridge this gap, we propose shifting physical evaluation toward a complementary \emph{deployment-aware} setting. In real applications, practitioners generating synthetic data already possess explicit prior knowledge of their setup, e.g., the robot's specific kinematic structure, task instructions, and fixed camera configurations. While evaluating world models as generic video generators is useful for open-world assessment, it leaves these critical structural priors unused. By actively leveraging these priors, we can construct deterministic, physics-grounded measurements that better align with the requirements of real-world robotics. Further, deployment introduces another critical constraint: \textit{inference efficiency}. The high generation latency of world models poses a severe throughput bottleneck, motivating acceleration techniques like caching, sparse attention, and token pruning~\citep{lv2025fastercache,bu2026dicache,xi2025sparse,yang2026sparse,li2026radial,li2026pisa,nawaz2026worldcache}. Since acceleration compresses structural representations, it preserves superficial perceptual appearance while silently violating strict kinematic constraints. Evaluating this quality-efficiency trade-off is thus an essential, yet largely unexplored, dimension of practical robotic deployment.

We introduce \ours, a comprehensive,\textbf{ deployment-aware benchmark} for evaluating both the structural fidelity and computational efficiency of robotic video world models. \ours{} spans 1,605 episodes curated from five source datasets, covering single-arm, bimanual, and complex humanoid morphologies across single- and multi-view setups as well as short- and long-horizontal tasks (Fig.\ref{fig:teaser}-a). To systematically localize physical failure modes, we propose two novel, \textit{VLM-free algorithmic evaluators}. While prior works attempt to measure physical properties by asking generic VLM judges if a video ``looks correct''~\citep{rong2026h2r,li2026robotrustbench,yang2026mirabench}, a method fundamentally lacking in strict spatial reasoning, we instead leverage known deployment priors to construct deterministic, physics-grounded measurements (Fig.~\ref{fig:teaser}-b).

First, we audit intra-view physical hallucinations through \textbf{KIVA}, which maps robot joint actions to keypoints to evaluate kinematic violations and physical plausibility. Since hallucinated videos yield out-of-distribution visual tokens, these estimates violate kinematic invariants, explicitly exposing the errors. Second, we verify cross-view consistency through \textbf{RIGIS}, a Robotic Inter-view Geometry \& Intrinsic-Extrinsic Stability framework, which ensures that synchronized views form a unified 3D scene with temporally stable camera extrinsics and intrinsics. Armed with these deterministic metrics, we utilize \ours{} to conduct a comprehensive audit of the quality-efficiency trade-off in robotic video generation. In summary, our core contributions are:
\begin{enumerate}
    \item \textbf{A Diverse Benchmark for Robot World Models:} We curate a rigorous evaluation suite of 1,605 episodes encompassing three major robot morphologies (single-arm, bimanual, humanoid). The benchmark spans both short Markovian interactions and long-horizon, non-Markovian tasks involving multiple subgoals and up to four synchronized camera views.
    \item \textbf{Deployment-Aware Physical Auditing:} We introduce KIVA and RIGIS, two deterministic, physics-grounded metrics leveraging known embodiment and geometric priors to evaluate kinematic stability and multi-view consistency. These metrics directly contrast with, and systematically outperform, VLM-based evaluators in human-preference alignment.
    \item \textbf{Auditing Inference Acceleration at Scale:} We systematically evaluate the quality-efficiency trade-off by auditing 9 state-of-the-art acceleration strategies (Fig.~\ref{fig:teaser}-c), spanning token merging, feature caching, and sparse attention, integrated across three leading world models \emph{DreamGen}~\citep{dreamgen}, \emph{DreamDojo}~\citep{gao2026dreamdojo}, and \emph{Ctrl-World} ~\citep{guo2026ctrl}. By assessing over 29,000 generated rollouts, we expose how computational shortcuts (yielding up to a 2.3$\times$ max. speed-up) silently compromise deployment-critical physical constraints that remain entirely invisible to standard visual-quality metrics.
\end{enumerate}

Our evaluation shows that visually convincing robotic videos can suffer substantial kinematic and multi-view inconsistencies that standard metrics and VLM judges miss. Such failures compounding over longer rollouts. \ours{} exposes these deployment-critical failures and quantifies the speed--integrity trade-off under acceleration, enabling evaluation beyond perceptual quality alone. We will make our benchmark and code publicly available upon acceptance for further research.

\section{Related Work}

\begin{table*}[!htb]
\vspace{-0.1in}
    \centering
    \caption{
    \textbf{Comparison of Video World Model Evaluation Benchmarks}. \textbf{\ours derives deterministic, deployment-aware metrics from dataset-native hardware priors}, scaling across multi-view setups, complex morphologies, and long-horizon tasks while auditing cross-view geometry and inference acceleration. Prior benchmarks rely on external VLMs or off-the-shelf models that do not exploit priors such as camera calibration, robot proprioception, and actions. \textit{Off. Alg.} denotes off-the-shelf algorithms; \textit{Hybrid} combines \textit{Off. Alg.} and \textit{VLM}; \textit{IK / Sim} denotes Inverse Kinematics and Simulation; ``\textit{Prior Alg.}'' denotes deterministic algorithms driven by dataset priors.}
    \label{tab:benchmark_comparison}

    \setlength{\tabcolsep}{2pt} 
    \renewcommand{\arraystretch}{1.25}
    
    \small

    \resizebox{\textwidth}{!}{%
    \begin{tabular}{l c c c c c c c c c}
        \toprule

        & 
        \multicolumn{5}{c}{\textbf{Data Diversity}} & 
        \multicolumn{4}{c}{\textbf{Evaluation Dimensions}} \\

        \cmidrule(lr){2-6}
        \cmidrule(lr){7-10}

        \textbf{Benchmark}
        & \makecell{\textbf{\#}\\\textbf{Samples}}
        & \makecell{\textbf{Data}\\\textbf{Sources}}
        & \makecell{\textbf{Robot}\\\textbf{Morphologies}}
        & \makecell{\textbf{Multi-View}\\\textbf{Setup}}
        & \makecell{\textbf{Long-Horizon}\\\textbf{Tasks}}
        & \makecell{\textbf{Evaluator}\\\textbf{Paradigm}}
        & \makecell{\textbf{Cross-View}\\\textbf{Geometry}}
        & \makecell{\textbf{Speedup}\\\textbf{Audit}} \\

        \midrule
        \multicolumn{10}{l}{\textit{\underline{General Video Evaluation}}} \\ 
        VBench
        & 800 & N/A & N/A & \xmark & \xmark
        & Hybrid & \xmark & \xmark \\

        VBench2.0
        & 1260 & N/A & N/A & \cmark & \xmark
        & Hybrid & \cmark & \xmark \\

        \midrule
        \multicolumn{10}{l}{\textit{\underline{VLM-Based Robotics Evaluation}}} \\
        
        WorldModelBench
        & 350 & 3 & 1 & \xmark & \xmark
        & Hybrid & \xmark & \xmark \\

        PAI-Bench
        & 2808 & 12 & 2 & \xmark & \xmark
        & VLM  & \xmark & \xmark \\

        H2R-Bench
        & 240 & 1 & 2 & \xmark & \xmark
        & VLM  & \xmark & \xmark \\

        RoboTrustBench
        & 1207 & 1 & 1 & \xmark & \cmark
        & VLM  & \xmark & \xmark \\

        MiraBench
        & 906 & 3 & 3 & \xmark & \cmark
        & VLM  & \xmark & \xmark \\

        \midrule
        \multicolumn{10}{l}{\textit{\underline{Sim/Analytical Robotics Evaluation}}} \\
        
        EWMBench
        & 100 & 1 & 1 & \xmark & \xmark
        & Off. Alg. & \xmark & \xmark \\

        RoboWM-Bench
        & 240 & 1 & 2 & \xmark & \cmark
        & IK / Sim & \xmark & \xmark \\

        \midrule
        \multicolumn{10}{l}{\textbf{\textit{\underline{Deployment-Aware (Ours})}}} \\
        
        \textbf{\ours}
        & \textbf{1,605}
        & \textbf{5}
        & \textbf{3 (inc. Humanoid)}
        & \cmark
        & \cmark
        & \textbf{Prior Alg.}
        & \cmark
        & \cmark \\

        \bottomrule
    \end{tabular}%
    }
\end{table*}

\paragraph{Evaluating VWMs for Robotics.}
Early evaluation frameworks for video generation, such as VBench~\citep{huang2024vbench} and EvalCrafter~\citep{liu2023evalcrafter}, assess perceptual quality, temporal consistency, and prompt alignment using standard visual metrics like Fréchet Video Distance (FVD)~\citep{unterthiner2018towards} or CLIP scores~\citep{radford2021learning}. While recent robotics suites like RoboTrustBench~\citep{li2026robotrustbench} and MiraBench~\citep{yang2026mirabench} shift toward physical tasks, they largely rely on subjective VLM judges or zero-shot algorithmic metrics applied without environment-specific adaptation. Conversely, analytical baselines like RoboWM-Bench leverage inverse kinematics (IK) in simulation to check physical consistency; however, such pipelines frequently struggle due to compounding errors during real-to-simulation conversion. In contrast, \ours{} introduces a deployment-aware paradigm that bypasses both brittle real-to-sim approximations and subjective VLM juries. By leveraging explicit priors such as kinematic URDFs, task instructions, and source videos, \textsc{KIVA} and \textsc{RIGIS} provide deterministic, prior-grounded measurements of intra-view kinematics and inter-view geometry (c.f. Table~\ref{tab:benchmark_comparison}).

\begin{figure}[t]
\centering
\vspace{-0.2in}
\includegraphics[width=0.88\linewidth]{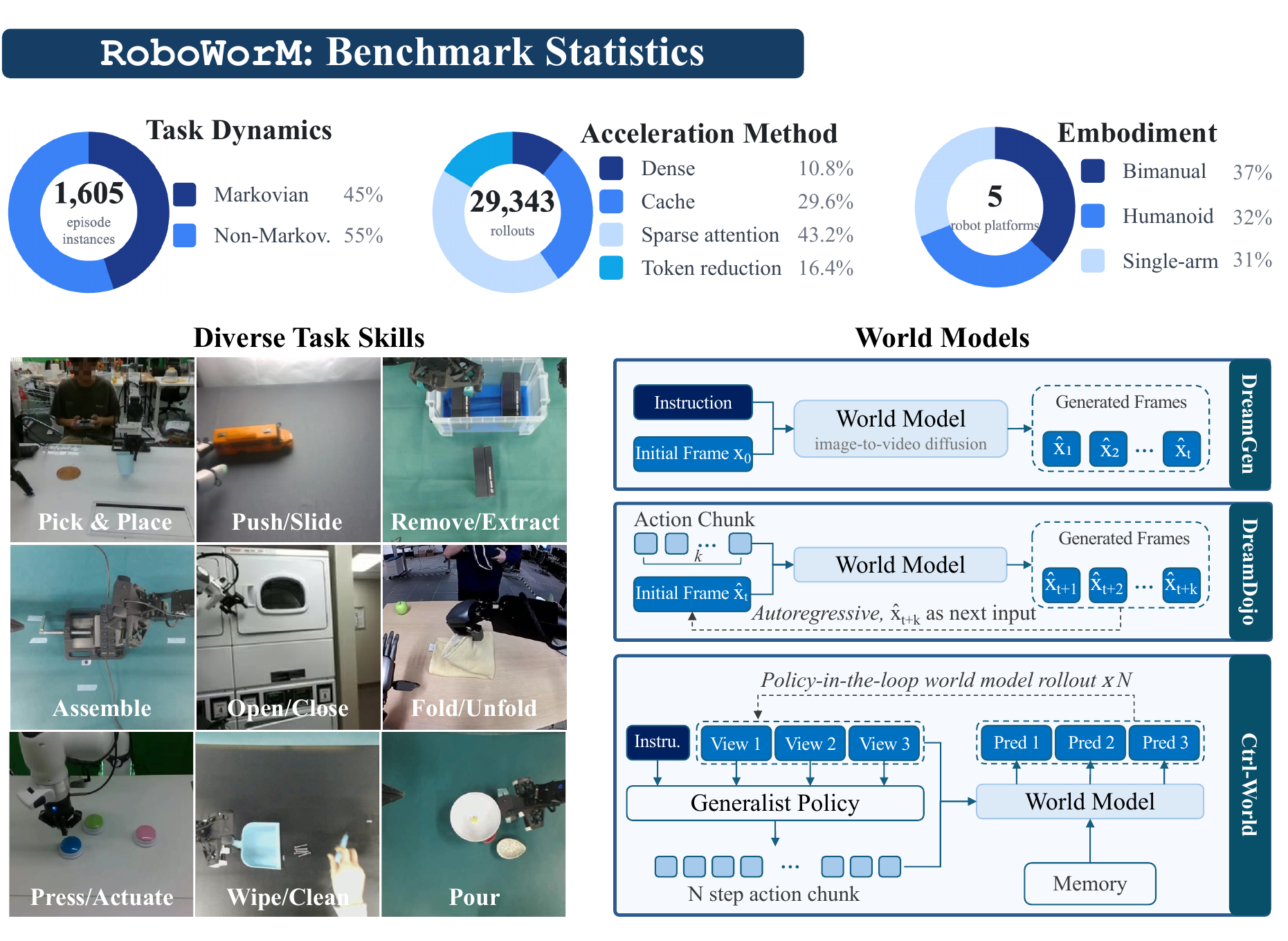}
\vspace{-0.1in}
\caption{
\newtext{\textbf{Benchmark Composition and Statistical Breakdown of \ours}. Data diversity spans a balanced mix of Markovian (short-horizon) and non-Markovian (long-horizon) task dynamics, task skills, and an even distribution across three robot morphologies. We benchmark 9 state-of-the-art acceleration methods on 29,000$+$ generated rollouts from three leading world models.}
\vspace{-0.15in}
}
\label{fig:composition}
\end{figure}

\paragraph{Multi-View and Long-Horizon Robotic Datasets.}
The robotics community has amassed large-scale datasets, most notably Open X-Embodiment~\citep{o2024open} and DROID~\citep{droid}, which have been instrumental in training generalized control policies and video world models. However, benchmark evaluations frequently inherit a narrow subset of this data, heavily biasing assessment toward single-arm, single-view tabletop tasks characterized by short, Markovian dynamics. While recent video world models like MultiWorld~\citep{wu2026multiworld} push the frontier of scalable multi-agent, multi-view generation, and diagnostic suites like MultiView-Bench~\citep{multiviewbench2026} construct multi-view VQA datasets for spatial understanding, their evaluation pipelines still heavily depend on VLMs as judges. The evaluation of complex embodiments and deterministic spatial reasoning in deployment settings remains vastly under-explored. 
\ours{} addresses this gap with a diverse benchmark of multi-view, long-horizon tasks across complex embodiments to rigorously evaluate unified 3D state representations (Fig.~\ref{fig:composition}).
\vspace{-0.1in}
\paragraph{Inference Acceleration in Generative Models.}
Diffusion models and autoregressive transformers suffer from severe inference latency, limiting their utility in closed-loop robotic control. A rich literature of acceleration techniques has emerged to address this, including token merging/pruning~\citep{bolya2022token,shao2026holitom,tran2024accelerating,zhang2025training,wu2025importance}, KV-caching~\citep{lv2025fastercache,bu2026dicache,nawaz2026worldcache}, and sparse attention mechanisms~\citep{xi2025sparse,yang2026sparse,li2026radial,li2026pisa}. While these methods report significant speed-ups for standard image and video synthesis, they are almost exclusively evaluated on perceptual metrics (e.g., FVD). Their impact on the physical fidelity of the generated output remains untested. As these methods aggressively compress spatial structure, they risk silently deteriorating kinematic linkage or geometric stability in robotic simulations. \ours{} provides the first systematic audit of inference acceleration methods specifically for robotic world models, explicitly mapping how computational shortcuts compromise deployment-critical physical constraints.
\section{\ours}
\vspace{-0.05in}
\subsection{Dataset Curation and Deployment Task Formulation}
\vspace{-0.05in}
\ours{} evaluates the physical consistency of generated rollouts across \emph{embodiments}, \emph{camera setups}, and \emph{temporal dynamics}. We curate our dataset from five public corpora: DROID~\citep{droid}, RoboChallenge~\citep{robochallenge}, AIST Bimanual
corpus~\citep{aist2025bimanip}, RoboCOIN~\citep{robocoin}, and NVIDIA GR00T
Teleop GR1~\citep{gao2026dreamdojo}, yielding six distinct configurations spanning five robot platforms and three morphology classes (c.f. Fig.~\ref{fig:composition}).

Each configuration consists of both Markovian and non-Markovian rollouts. For corpora with free-form instructions, temporal dynamics regimes are assigned via automated LLM classification (Qwen3-VL-30B) using sequencing and iterative markers, whereas multi-step instructions use manual task labeling. See Appendix~\ref{supp:dataset} for dataset-to-configuration mappings, and curation protocol.

\subsection{Video World Model Evaluation Suite and Rollout Generation}
\label{sec:vwm_configures}

\vspace{-0.05in}
To systematically examine generation capabilities, we evaluate three video world model families across three backbone architectures: DreamGen~\citep{dreamgen} (scalable general-purpose DiT, 2B/14B), DreamDojo~\citep{gao2026dreamdojo} (action-conditioned DiT, 2B), and Ctrl-World~\citep{guo2026ctrl} (spatio-temporal U-Net for synchronized multi-view generation). We evaluate 12 checkpoints matched to our setup: nine were explicitly fine-tuned on target domains, while three (Ctrl-World/DreamGen on single-arm/multi-view; DreamDojo on humanoid/single-view) reuse public weights. Configuration and dense-baseline details are deferred to the Appendix~\ref{supp:impl_finetuning}.

\subsection{Systematic Inference Acceleration Audit}
As robotic deployment heavily relies on real-time control, VWMs are increasingly paired with inference acceleration algorithms. 
We evaluate nine acceleration algorithms against a \textbf{dense} baseline, grouped into three categories: (1) \textbf{Cache Strategies} with FasterCache~\citep{lv2025fastercache}, DiCache~\citep{bu2026dicache}, World-Cache~\citep{nawaz2026worldcache}; (2) \textbf{Sparse Attention} with Sparse VideoGen 1 \& 2~\citep{xi2025sparse,yang2026sparse}, Radial Attention~\citep{li2026radial}, PISA~\citep{li2026pisa}; and (3) \textbf{Token Pruning} with SiTo~\citep{zhang2025training}, ITM~\citep{wu2025importance}.


\textbf{Implementation Constraints}: There are several architectural constraints that limit universal compatibility. For instance, token pruning requires DiT architectures and is incompatible with U-Nets. The full integration compatibility matrix is detailed in Fig.~\ref{fig:supp_coverage}. Overall, we rolled out approximately 29,000 samples for these acceleration techniques integrated into VWM in Sec.~\ref{sec:vwm_configures}.

\subsection{Deployment-Aware Physics Auditing with KIVA and RIGIS}
\begin{figure}[t]
    \centering
    \vspace{-0.15in}
\includegraphics[width=1.02\linewidth]{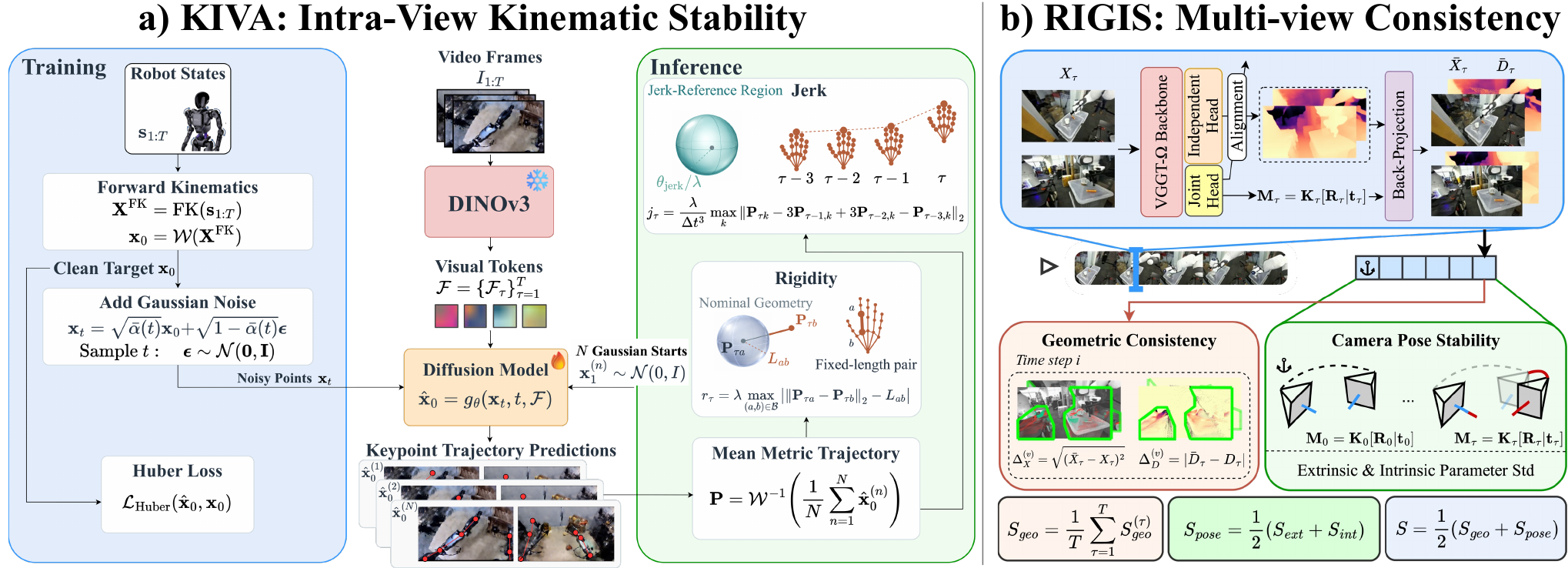}
    \caption{We introduce \textbf{KIVA for intra-view kinematics stability} and \textbf{RIGIS for multi-view consistency} in deployment-aware setting. \textbf{(a) KIVA.} A shared video-conditioned reader learns from forward-kinematics targets using Huber loss. At inference, five denoised  candidates are averaged before measuring rigidity and jerk, which are then calibrated  against real-video references. The hand examples and numerical values, including the reference thresholds, are illustrative rather than empirical results. \textbf{(b) RIGIS.} Multi-view inputs are processed via VGGT-$\Omega$ to evaluate geometric consistency $S_{\mathrm{geo}}$ and camera pose stability $S_{\mathrm{pose}}$, yielding the unified metric $S$.}
    \vspace{-0.2in}
    \label{fig:metrics}
\end{figure}
\subsubsection{KIVA: \texorpdfstring{\underline{K}inematic \underline{I}nvariant \underline{V}iolation \underline{A}ssessment}{Kinematic Invariant Violation Assessment}}
\label{sec:kiva}

\textbf{Intuition and Trajectory Estimation}. KIVA (Fig.~\ref{fig:metrics}-a) employs a diagnostic diffusion model $g_\theta$ to bridge visual features and 3D kinematics. During training on real-world videos, $g_\theta$ is supervised via Huber loss to reconstruct normalized coordinates derived from exact forward kinematics $\mathbf{X}^{\mathrm{FK}} = \operatorname{FK}(\mathbf{s}_{1:T})$, where $\mathbf{s}_{1:T}$ denotes the robot states obtained across $T$ frames in real-robot datasets and $\operatorname{FK}$ represents the forward kinematics equation, conditioned on DINOv3~\citep{simeoni2025dinov3} visual tokens $\mathcal{F}_\tau$ at each frame $\tau \in T$. This ground-truth supervision ensures $g_\theta$ yields precise, low-variance reconstructions for clean videos. Conversely, we observe that hallucinated videos produce out-of-distribution visual features. Because the diagnostic model is only trained on pristine, real-world data, these anomalous features cause its predictions to become highly unstable, leading to severe divergence across different noise seeds (as illustrated in Fig.~\ref{fig:teaser}-b, left). During inference, we average $N$ independent DDIM sampling runs ($N=5$ in our experiment) and apply an inverse workspace transformation. This maps the aggregated diffusion outputs back into the original metric space, yielding the deterministic, un-normalized 3D trajectory $\mathbf{P}_{\tau k}$. Definitions of the diffusion forward process, workspace transformations, and other parameters are deferred to Appendix~\ref{supp:metrics_hallucination}.

\textbf{Rigidity, Jerk, and Real-Video References.} We quantify physical violations directly on these estimated metric trajectories. Given structurally connected keypoint pairs $\mathcal{B}$, known nominal bone lengths $L_{ab}$ (in meters), and a frame interval $\Delta t$, we define the instantaneous \texttt{Rigidity} error $r_\tau$ (kinematic link deformation) and \texttt{Jerk} $j_\tau$ (abrupt temporal discontinuities, approximated via third-order backward finite differences):$$r_\tau = \lambda \max_{(a,b)\in\mathcal{B}} \left\vert{}\Vert{}\mathbf{P}_{\tau a} - \mathbf{P}_{\tau b}\Vert{}_2 - L_{ab}\right\vert{}, \quad j_\tau = \frac{\lambda}{\Delta t^3} \max_k \Vert{}\mathbf{P}_{\tau k} - 3\mathbf{P}_{\tau-1, k} + 3\mathbf{P}_{\tau-2, k} - \mathbf{P}_{\tau-3, k}\Vert{}_2$$where $\lambda$ aligns the metric errors into millimeter units. To contextualize these errors for a given video segment $S$, we compute the summary metrics $v_S^{\mathrm{rig}} = \operatorname{median}_{\tau\in S} r_\tau$ and $v_S^{\mathrm{jerk}} = \max_{\tau\in S} j_\tau$. We then report dimensionless ratios normalized against a reference set $\mathcal{R}$ of real-world videos:
\begin{equation*}
    s_S^d = \frac{v_S^d}{\theta_d}, \qquad \theta_d = Q_{0.95}\!\left(\{v_R^d : R \in \mathcal{R}\}\right), \quad d \in \{\mathrm{rig}, \mathrm{jerk}\},
\end{equation*}
where $Q_{0.95}$ denotes the 95th percentile over $\mathcal{R}$, computed by running the diagnostic model on real training videos. Since the model itself has a small reconstruction error, even real recordings yield non-zero rigidity and jerk, so raw values have no absolute meaning. Normalizing by $\theta_d$ anchors each score to real footage and makes it comparable across embodiments. The decision boundary is then a single scalar on $s_S=\max_d s_S^d$, which we calibrate per embodiment against human labels rather than fixing it at the real-video level $s_S=1$ (c.f. Appendix~\ref{supp:metrics_hallucination}).

\subsubsection{RIGIS: \texorpdfstring{\underline{R}obotic \underline{I}nter-view \underline{G}eometry \&{} \underline{I}ntrinsic-extrinsic \underline{S}tability} {Robotic Inter-view Geometry \& Intrinsic-Extrinsic Stability}}
\label{sec:rigis}

While KIVA audits intra-view kinematics, evaluating multi-view models requires verifying cross-camera spatial consistency. Existing metrics often fall short: VLMs lack explicit 3D reasoning, and standard 3D estimators inadvertently smooth over generated geometric hallucinations. We introduce RIGIS to enforce strict multi-camera physical constraints by anchoring its evaluation directly to the underlying spatial layout and dataset’s ground-truth priors (c.f. Fig~\ref{fig:metrics}-b).

\vspace{0.05in}
\textbf{Multi-View Alignment Pipeline.}
Rather than blindly estimating spatial parameters, RIGIS leverages the dataset’s known static camera configurations and workspace limits. For a generated rollout, it verifies that reprojected pixels form a unified 3D scene that does not drift from the true robotic setup. At each timestep $\tau \in \{1 \dots T\}$ across all active frames, the multi-view input frames $X_\tau$ are processed by a modified VGGT-$\Omega$ backbone \citep{wang2026vggtomega}. An independent prediction head estimates per-view depth maps, while a joint head predicts the camera projection matrix $\mathbf{M}_\tau = \mathbf{K}_\tau[\mathbf{R}_\tau \vert{} \mathbf{t}_\tau]$. These outputs are scale-aligned, fused, and back-projected to generate reconstructed renders $(\bar{X}_\tau, \bar{D}_\tau)$ within the visible intersection region.

\vspace{0.05in}
\textbf{Geometric Consistency.}
We compare the reprojected depth map against the scale-aligned depth map by measuring the per-view RGB root mean square error $\Delta_X^{(v)} = \sqrt{(\bar{X}_\tau - X_\tau)^2}$ and the mean absolute depth residual $\Delta_D^{(v)} = \vert{}\bar{D}_\tau - D_\tau\vert{}$. These errors are mapped via an exponential decay function to bounded scores $s_X^{(v)}, s_D^{(v)}$. The overall geometric consistency score $S_{\mathrm{geo}}$ is computed by averaging the joint RGB-depth fidelity across all $V$ views over the temporal sequence:
$$S_{\mathrm{geo}} = \frac{1}{T}\sum_{\tau=1}^{T} S_{\mathrm{geo}}^{(\tau)} = \frac{1}{T}\sum_{\tau=1}^{T} \prod_{v=1}^{V} \frac{1}{2}\left(s_X^{(v)} + s_D^{(v)}\right)\,.$$

\textbf{Camera Pose Stability.}
For a fixed robotic setup, the physical camera rig should not drift during generation. RIGIS tracks the temporal standard deviation of the predicted intrinsics ($\sigma_K$) and extrinsic relative poses ($\sigma_R, \sigma_t$) with respect to the initial conditioned frame at $\tau=1$. These deviations are mapped to stability scores:
$S_{\mathrm{ext}} = \frac{1}{2} \left(\mathrm{e}^{-\sigma_R/5} + \mathrm{e}^{-\sigma_t}\right), \quad S_{\mathrm{int}} = \frac{1}{V}\sum_{v=1}^{V} \mathrm{e}^{-\sigma_K^{(v)}}$
yielding the unified pose stability metric $S_{\mathrm{pose}} = \frac{1}{2}(S_{\mathrm{ext}} + S_{\mathrm{int}})$.
The final RIGIS score is defined as the unweighted mean of spatial fidelity and rig stability:
$S = \frac{1}{2}(S_{\mathrm{geo}} + S_{\mathrm{pose}}) \in [0, 1]$
where a higher score strictly indicates that the video world model successfully maintains coherent multi-view 3D state representations without spatial drift. See detailed formulations and calibration in Appendix~\ref{supp:metrics_multiview}.

\section{Experiments}
\subsection{Evaluation Protocol} 
\label{sec:evaluation_protocol}

\myparagraph{Reference Metrics and Baselines.} To complete the evaluation aspects, we incorporate foundational metrics, namely Visual Integrity~\citep{wang2023exploring}, Peak Signal-to-Noise Ratio (PSNR), Text Alignment~\citep{hessel2021clipscore}, and Fr\'{e}chet Video Distance (FVD)~\citep{unterthiner2018towards}. To evaluate automated judging, we use two closed-source (Gemini 3.6 Flash, Qwen 3.6 Plus) and two open-source (Gemma 4 31B, Qwen 3.6 35B) VLMs under zero-shot and CoT settings. See detail description in App.~\ref{supp:metrics_other} , and VLM judgement setup in Appendix~\ref{supp:vlm_setup}.

\myparagraph{Score Aggregation.}
KIVA scores are computed as the mean segment violation rate over single-view rollouts, while RIGIS scores represent the mean per-rollout score under identical pixel inputs. 
For KIVA, a segment is flagged as a violation when its rigidity or jerk error exceeds the 95th percentile threshold of real-video references ($\max(s^{\mathrm{rig}}, s^{\mathrm{jerk}}) > 1$). All scores are normalized relative to real video references. Further details are provided in Appendix~\ref{supp:metrics_hallucination} and \ref{supp:metrics_multiview}.

\myparagraph{Speedup Measurement.} 
Speedup is defined as the ratio of dense baseline to accelerated generation time per segment, aggregated across all episodes. Timings utilize internal timers to exclude constant overheads like model loading and VAE decoding. Measuring per segment isolates algorithmic efficiency by removing biases from variable episode lengths (c.f. Appendix~\ref{supp:impl_acceleration}).

\myparagraph{Human Study.}
We curate two diverse subsets balanced across robot morphologies, 291 single-view videos for embodied physics and 300 multi-view videos for multi-view consistency, graded on a three-point scale by two annotators with expert resolution. Full details are reported in Appendix~\ref{supp:human_vlm}.

\begin{figure*}[t]
  \centering
  \vspace{-0.2in}
  \includegraphics[width=\textwidth]{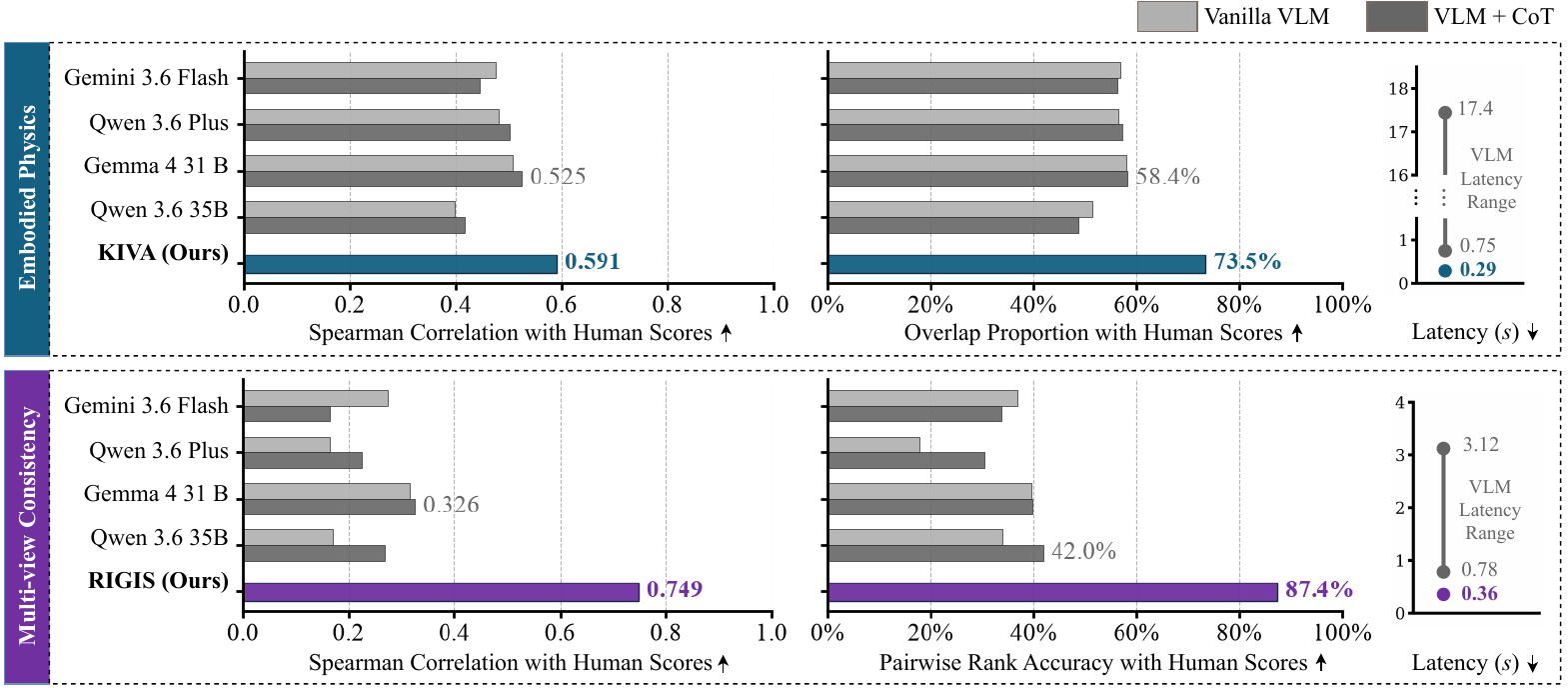}
  \caption{\textbf{Our metrics KIVA (top) and RIGIS (bottom) shows significantly higher human alignment correlation and accuracy, while achieving better processing latency.} 
  }
  \vspace{-0.3in}
  \label{fig:quantitative_human_alignment}
\end{figure*}

\subsection{Human Alignment and Cross-Metric Analysis}
\label{sec:result_rq1}
\noindent\textbf{RQ1: Do KIVA and RIGIS align better with human judgment than VLM judges?}

Fig.~\ref{fig:quantitative_human_alignment} shows that our kinematic- and geometry-grounded metrics outperform VLM judges: KIVA improves human-aligned hallucination detection by 25.8\% (58.4\% to 73.5\%) over frontier VLMs, and RIGIS achieves a 130\% higher Spearman correlation with human preferences (0.326 to 0.749). Even when equipped with fewshot understanding, the best-performing judge (Gemma~4~31B + CoT) remains below KIVA. Comprehensive results are provided in Appendices~\ref{supp:human_kiva} and~\ref{supp:human_rigis}.

\noindent\textbf{RQ2: How do KIVA and RIGIS compare to VLMs in evaluation stability and efficiency?}

VLM judges exhibit severe structural instability, with score standard deviations ranging from $0.076$ for embodied physics to $0.591$ for multi-view consistency, alongside massive computational overheads under chain-of-thought reasoning. Our algorithmic metrics, on the other hand, achieve significantly better latency as illustrated in Fig.~\ref{fig:quantitative_human_alignment}. We provided further detailed statistical distributions, variance analyses, and latency benchmarks in Appendix~\ref{supp:vlm_results_summary}.

\vspace{0.05in}
\noindent\textbf{RQ3: How do existing metrics correlate with KIVA and RIGIS?}

We compare KIVA against baselines like Motion Smoothness (MS) \citep{duan2025worldscore} and Motion Amplitude Score (MAS) \citep{deng2026rethinking}, observing they capture only surface transitions and are more sensitive to hallucinations. KIVA instead leverages embodied kinematic constraints to evaluate physical plausibility. Comparing RIGIS to feature matching (MVC) \citep{zheng2025vbench2} and SLAM RPE \citep{wu2026multiworld,wu2026geometry} reveals that conventional pipelines fail under large viewpoints; RIGIS instead evaluates holistic multi-view consistency, aligning closely with human judgment (Fig.~\ref{fig:qualitative_human_alignment}-a).

\subsection{Long-Horizon Drift in Video World Models}
\label{sec:result_rq4}
\newcommand{\rqna}{\textcolor{black!55}{--}}  
\newcommand{\rqns}{\textcolor{black!55}{n/s}}  
\newcommand{\rqn}[1]{\makebox[2.45em][r]{#1}}
\newcommand{\rqnc}[1]{\makebox[2.45em][c]{#1}}

\begin{table*}[t]
\centering
\small
\setlength{\tabcolsep}{1pt}
\renewcommand{\arraystretch}{1.12}
\vspace{-0.2in}
\caption{\textbf{Dense world-model results show a setting-dependent performance gap: Ctrl-World dominates multi-view tracks, while DreamDojo leads single-view tracks.} Median performance under Markovian and non-Markovian dynamics. \textbf{Bold} indicates the better model per embodiment; ``N/A'' denotes not applicable. FVD follows panel-specific protocols: multi-view rows share a per-camera reference, while single-view rows use per-configuration references and are read vertically.}
\vspace{-0.1in}
\label{tab:dense_baselines}
\vspace{2pt}
\adjustbox{max width=1.0\textwidth}{%
\begin{tabular}{@{}llccccccccccc}
\toprule
& & & \multicolumn{4}{c}{\textbf{Visual fidelity}} & \multicolumn{2}{c}{\textbf{Text alignment}$\uparrow$} & \multicolumn{2}{c}{\textbf{FVD}$\downarrow$} & \multicolumn{2}{c}{\textbf{Physics}} \\
& & & \multicolumn{2}{c}{Vis.\ integrity$\uparrow$} & \multicolumn{2}{c}{PSNR$\uparrow$} & & & & & \multicolumn{2}{c}{\scriptsize RIGIS $\uparrow$ or KIVA $\downarrow$}  \\
\cmidrule(lr){4-5}\cmidrule(lr){6-7}\cmidrule(lr){8-9}\cmidrule(lr){10-11}\cmidrule(lr){12-13}
\textbf{Morphology} & \textbf{Model} & \textbf{Size} & Mkv & Non-Mkv & Mkv & Non-Mkv & Mkv & Non-Mkv & Mkv & Non-Mkv & Mkv & Non-Mkv \\
\midrule
\multicolumn{11}{@{}l}{\textbf{(a) Multi-view tracks} \quad {\footnotesize shared per-camera real reference}} & \multicolumn{2}{c}{\footnotesize\emph{RIGIS $\uparrow$}} \\
\addlinespace[3pt]
\multirow{2}{*}{Bimanual}   & DreamGen         & 2B   & 27.79 & 25.28 & 14.59 & 12.80 & 24.9 & 25.3 & 233 & 162 & 0.785 & 0.767 \\
                            & Ctrl-World       & 1.5B & \textbf{33.63} & \textbf{32.36} & \textbf{23.51} & \textbf{22.23} & \textbf{25.4} & \textbf{25.9} & \textbf{29} & \textbf{25} & \textbf{0.803} & \textbf{0.779} \\
\addlinespace[3pt]
\multirow{2}{*}{Humanoid}   & DreamGen         & 2B   & 22.48 & 20.50 & 13.37 & 13.09 & 22.9 & \textbf{21.7} & 191 & 127 & 0.860 & 0.844 \\
                            & Ctrl-World       & 1.5B & \textbf{27.46} & \textbf{25.88} & \textbf{28.42} & \textbf{28.10} & \textbf{23.2} & 21.5 & \textbf{29} & \textbf{16} & \textbf{0.879} & \textbf{0.878} \\
\addlinespace[3pt]
\multirow{2}{*}{Single-arm} & DreamGen         & 14B  & 29.53 & 29.27 & 17.06 & 16.48 & 24.8 & \textbf{25.0} & 125 & 161 & 0.907 & 0.901 \\
                            & Ctrl-World       & 1.5B & \textbf{31.51} & \textbf{31.46} & \textbf{17.65} & \textbf{19.71} & \textbf{25.2} & 24.9 & \textbf{78} & \textbf{55} & \textbf{0.909} & \textbf{0.909} \\
\midrule
\multicolumn{11}{@{}l}{\textbf{(b) Single-view tracks} \quad {\footnotesize per-configuration real reference; FVD not comparable across models}} & \multicolumn{2}{c}{\footnotesize\emph{KIVA $\downarrow$}} \\
\addlinespace[3pt]
\multirow{2}{*}{Bimanual}   & DreamGen  & 2B & 36.14 & 32.82 & 18.68 & 16.28 & 24.2 & 26.0 & 206 & 146 & 23.6 & 23.2 \\
                            & DreamDojo & 2B & \textbf{39.52} & \textbf{35.20} & \textbf{21.43} & \textbf{19.33} & N/A & N/A & 55 & 41 & \textbf{6.1} & \textbf{0.0} \\
\addlinespace[3pt]
\multirow{2}{*}{Humanoid}   & DreamGen  & 2B & 40.59 & 40.67 & 12.60 & 12.78 & 32.3 & 33.0 & 103 & 108 & \textbf{4.7} & 12.3 \\
                            & DreamDojo & 2B & \textbf{42.87} & \textbf{41.08} & \textbf{16.75} & \textbf{16.35} & N/A & N/A & 81 & 356 & 13.0 & \textbf{3.3} \\
\addlinespace[3pt]
\multirow{2}{*}{Single-arm} & DreamGen  & 2B & \textbf{29.72} & \textbf{39.79} & 19.95 & 19.70 & 30.4 & 35.7 & 155 & 172 & 59.4 & 70.7 \\
                            & DreamDojo & 2B & 28.88 & 38.42 & \textbf{26.61} & \textbf{24.63} & N/A & N/A & 49 & 57 & \textbf{21.3} & \textbf{27.0} \\
\bottomrule
\end{tabular}}
\vspace{-0.28in}
\end{table*}

\textbf{RQ4: How do Video World Models perform across morphologies and long horizons?}

\begin{figure*}[thb]
  \centering
  \includegraphics[width=\textwidth]{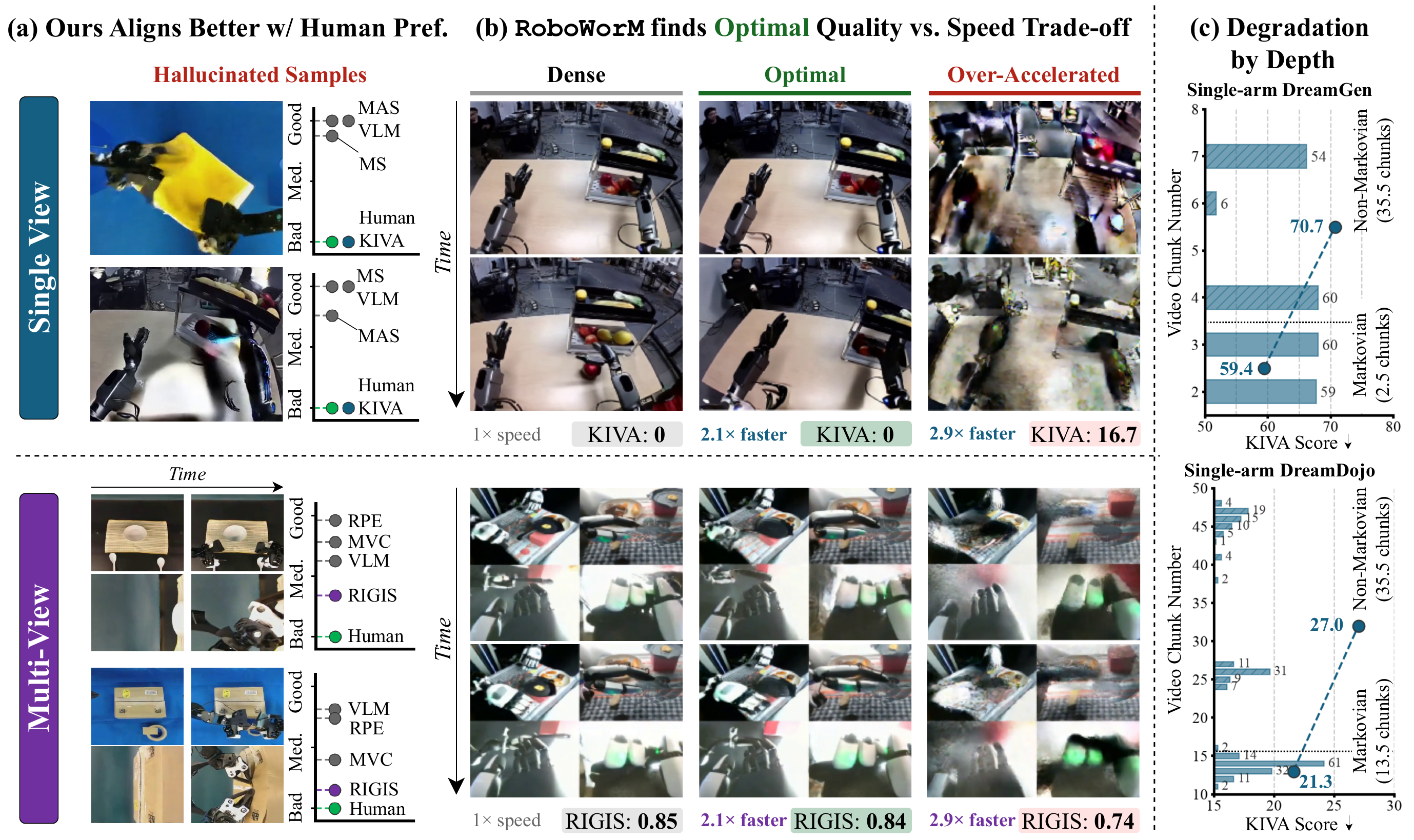}
  \caption{\textbf{\ours identifies optimal speedup configurations without quality trade-offs where standard metrics fail}. (a) KIVA and RIGIS accurately evaluate embodied hallucinations and geometry. (b) \ours systematically profiles the speedup-quality frontier. (c) Correlation between KIVA degradation and increased video chunk across Markovian and Non-Markovian tasks.}
  \vspace{-18pt}
  \label{fig:qualitative_human_alignment}
\end{figure*}

\myparagraph{Longer rollouts show stronger degradation.}
KIVA degrades mainly in the two deepest configurations of single-arm DreamGen and DreamDojo, where non-Markovian rollouts have higher rollout depth (i.e., more frames). Their supports are strictly disjoint, meaning regime and rollout depth are completely confounded (Fig~\ref{fig:qualitative_human_alignment}-c). This exposes their entanglement rather than isolating depth as the root cause, requiring our causal interpretation to rely instead on the preceding within-clip cascade.

\myparagraph{Architecture and cross-view consistency.}
Ctrl-World consistently outperforms DreamGen in RIGIS across all multi-view comparisons (c.f. Tab.~\ref{tab:dense_baselines}). This stems directly from architectural design: Ctrl-World applies explicit spatial-temporal processing, whereas DreamGen processes a $2\times2$ tiled frame via self-attention without native geometric constraints. Since RIGIS directly measures cross-view consistency, this structural difference naturally explains the observed performance gap.

\subsection{Acceleration Trade-offs}
\label{sec:result_rq5}
\textbf{RQ5: How does acceleration trade off speed and structural integrity in video world models?}

\begin{figure}[thb]
\centering
\vspace{-0.25in}
\includegraphics[width=\linewidth]{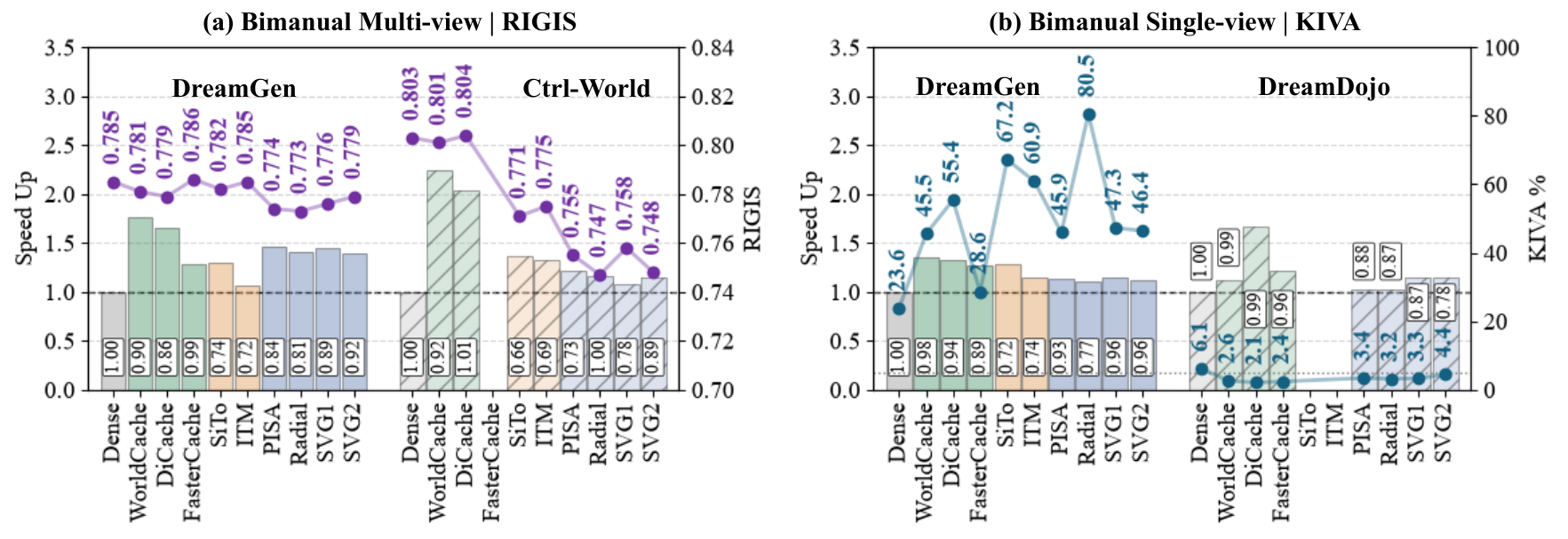}
\vspace{-0.15in}
\caption{\textbf{Acceleration mechanisms exhibit distinct speed--quality trade-offs.} Caching achieves the largest speedups with near-zero degradation; token pruning degrades most. Bars show speedup, the middle axis physical audit, and dots mean NR. See Fig.~\ref{fig:supp_accel_tradeoff_single_arm}, \ref{fig:supp_accel_tradeoff_humanoid} for other embodiments.}
 \vspace{-12pt}
\label{fig:accel_tradeoff}
\end{figure}

\myparagraph{Caching offers the most robust acceleration}, achieving a $1.7\times$
mean speedup with minimal quality loss ($-0.004$ RIGIS) and winning all $12$
configurations. WorldCache is its strongest member, at $1.8\times$ mean and a
benchmark-best $2.3\times$ on DreamGen humanoid and single-arm multi-view. Unlike
token pruning, which incurs higher error ($13.4\%$ NR loss) and slower speeds
($1.2\times$), or sparse attention (limited by low attention runtimes), caching
reuses computation safely across diverse backbone architectures.
\section{Conclusion and Discussion}

\vspace{-0.05in}
Our findings suggest that evaluating video world models requires a shift toward deployment-aware auditing. By grounding metrics in training data distributions and robot kinematics, \ours provides a complementary lens to standard visual assessment. Our analysis indicates that tracking kinematic violations, multi-view consistency, and the trade-offs of inference acceleration help reveal failure modes that are otherwise missed by generic metrics.

\textbf{Limitations.} Our \ours has several limitations: (i) KIVA and RIGIS do not verify contact-level physical validity, which would require integrating full physics simulators; (ii) they are currently limited to rigid-body motions and do not capture deformable materials or force-driven dynamics; and (iii) our acceleration audits measure output-level degradation without identifying which internal layers or layers of computation are responsible for losing physical fidelity. We leave these directions for future work.

\section*{AI USE STATEMENT}
We used large language models (LLMs) solely to improve the grammar, clarity, and readability of the manuscript. All research ideas, methods, experiments, analyses, and conclusions are the authors' own, and the authors reviewed and take full responsibility for all content of this paper.

\section*{Reproducibility Statement} To facilitate reproducibility, full details regarding the RoboWorM dataset collection, benchmark setup, and evaluation protocols are provided in the appendices. Code, evaluation scripts, and the complete benchmark dataset will be publicly released upon publication.

\bibliography{iclr2027_conference}
\bibliographystyle{iclr2027_conference}

\appendix

\renewcommand\thesection{\Alph{section}}
\numberwithin{equation}{section}
\numberwithin{figure}{section}
\numberwithin{table}{section}
\renewcommand{\thefigure}{\thesection\arabic{figure}}
\renewcommand{\thetable}{\thesection\arabic{table}}
\crefname{appendix}{Sec.}{Secs.}

{\onecolumn 
{\begin{center}
\Large\bf
{\ours: Evaluating Movement Realism, Consistency and Acceleration Affects in Embodied Video \underline{Wor}ld \underline{M}odels}\\[1em]
\large
Appendix
\end{center}
}

\newcommand{\tocsubgap}{0.7em}     
\newcommand{\tocsecgap}{1.4em}     
\newcommand{\subindent}{1.5em}

\newcommand{\additem}[2]{%
\item[\textbf{(\ref{#1})}] 
    \textbf{#2} \dotfill\makebox{\textbf{\pageref{#1}}}%
    \par\vspace{0.2em}%
}

\newcommand{\addsubitem}[2]{%
    \par\vspace{\tocsubgap}%
    \hspace*{\subindent}(\ref{#1})\hspace{.5em}#2 \dotfill\makebox{\pageref{#1}}\par%
}

\setlist[itemize]{noitemsep,leftmargin=*,topsep=0em}
\setlist[enumerate]{noitemsep,leftmargin=*,topsep=0em}

\vspace{0.3in}

\begin{adjustwidth}{0.15cm}{0.15cm}
\begin{enumerate}[label={({\arabic*})}, topsep=1em, itemsep=\tocsecgap]

    \additem{supp:dataset}{Benchmark Dataset Construction \& Dataset Details}
        \addsubitem{supp:dataset_sources}{Source Corpora \& Curation Protocol}
        \addsubitem{supp:dataset_processing}{Data Preparation}
        \addsubitem{supp:dataset_taxonomy}{Task Taxonomy \& Instruction Prompts}
        \addsubitem{supp:dataset_rollouts}{Distribution of the Generated Acceleration Data} 

    \additem{supp:metrics}{Metrics Formalism \& Implementation}
        \addsubitem{supp:metrics_hallucination}{Detailed Formulation of KIVA}
        \addsubitem{supp:metrics_multiview}{Detailed Formulation of RIGIS}
        \addsubitem{supp:metrics_other}{Additional Metrics Detail}

    \additem{supp:implementation_details}{World Model Training \& Acceleration Implementation}
        \addsubitem{supp:impl_finetuning}{Base World Model Finetuning Protocols}
        \addsubitem{supp:impl_acceleration}{Diffusion Acceleration Method Configurations}

    \additem{supp:human_vlm}{Human Assessment \& VLM Evaluation Protocols}
        \addsubitem{supp:human_protocol}{Human Assessment Protocol}
        \addsubitem{supp:vlm_setup}{VLM Judge Evaluation Setup}
        \addsubitem{supp:human_kiva}{Supplementary Results for KIVA}
        \addsubitem{supp:human_rigis}{Supplementary Results for RIGIS}

    \additem{supp:extension}{Using \& Extending \ours{}}
        \addsubitem{supp:extension_worldmodel}{Evaluating a New World Model}
        \addsubitem{supp:extension_embodiment}{Adding a New Embodiment or Dataset}
        \addsubitem{supp:extension_acceleration}{Evaluating a New Acceleration Method}
        \addsubitem{supp:extension_usecases_roboworm}{Additional Use Cases of RoboWorm}

\end{enumerate}
\end{adjustwidth}
}
\setlength{\parskip}{.5em}

\clearpage
\section{Benchmark Dataset Construction \& Dataset Details}
\label{supp:dataset}

\subsection{Source Corpora \& Curation Protocol}
\label{supp:dataset_sources}


\paragraph{Source corpora.}
Five publicly available corpora form the basis of the 12 evaluation settings: DROID (Franka Panda, three cameras) is used for the single-arm multi-view condition, whereas RoboChallenge (ARX-5 and Franka, one static camera) is used for the single-arm single-view condition; the AIST Bimanual corpus (ALOHA, $4$ cameras) for both bimanual conditions; RoboCOIN (AIRBOT MMK2) for the humanoid multi-view condition; and NVIDIA GR00T Teleop GR1 (GR1-T2 egocentric) for the humanoid single-view condition. Two correspondences are prone to misinterpretation. First, for each embodiment, the multi-view and single-view tracks are sourced from \emph{distinct} corpora (RoboChallenge vs.\ DROID; GR1 vs.\ RoboCOIN) and therefore contribute disjoint sets of episodes; only the two bimanual tracks share episodes. Second, although RoboCOIN logs tasks using two different camera rigs that are denoted by different view identifiers, these recordings constitute a single corpus rather than two separate datasets.

\paragraph{Temporal regimes.}
Each corpus is split into a \emph{Markovian} and a \emph{non-Markovian} subset.
A task is Markovian when the next observation follows from the current
observation and action (reaching, insertion, pushing, closing), and
non-Markovian when correct continuation requires history not visible in the
current frame  ordered multi-step sequences, deformable manipulation with
accumulated state, and hand-over coordination. Where instructions are free-form
per episode (DROID, GR1), we assign the label with an LLM classifier over the
instruction text (Qwen3-VL-30B at temperature $0$) under a fixed rubric:
sequencing connectives (\emph{then}, \emph{after}), ordinal markers, completion
quantifiers (\emph{all}, \emph{each}, \emph{remaining}) and iterative verbs
(\emph{repeatedly}, \emph{until}) mark non-Markovian, while a single
pick-and-place stays Markovian even when it names two locations; unlabelled
episodes are dropped rather than guessed. Where one instruction covers a whole
task (bimanual, RoboChallenge), the label is assigned manually per task  for
instance \emph{``put the cup on the coaster''} is Markovian, whereas
\emph{``press three buttons''} (prescribed order) and \emph{``wipe the table''}
(pull tissue $\rightarrow$ wipe $\rightarrow$ discard) are not. The bimanual
contrast in Table~\ref{tab:supp_dataset_taxonomy} follows: its Markovian subset is
dominated by gear, shaft and USB insertion, its non-Markovian subset by towel
folding and hand-over.

\paragraph{Episode selection.}
Evaluation settings are capped to keep evaluation cost comparable across $9$ acceleration
methods and $3$ generators, but the caps are not uniform. DROID and GR1 are
sampled to $130$ episodes per regime with a fixed seed; RoboChallenge takes
$60$ episodes per task, two tasks per regime. The bimanual and humanoid
multi-view settings are ingested exhaustively, so their counts are determined by
the source release rather than by a configured target. Two source pools substantially
exceed their caps: DROID contains $95{,}600$ episodes, while the humanoid
non-Markovian tasks provide $376$. Table~\ref{tab:supp_dataset_cells} reports
both the available pool and the selected subset to make the sampling ratio explicit.
The DROID pool is further restricted to the subset of shards downloaded for this study.

\begin{table}[t]
\centering
\scriptsize
\setlength{\tabcolsep}{4pt}
\caption{Composition of the twelve \ours{} evaluation settings, recomputed from the source metadata. \emph{Eps.} is the number of episodes we evaluate and \emph{Pool} the number available upstream; \emph{Views} is the synchronised cameras per episode and \emph{Instr.} the distinct instructions. Frame rate, resolution and view count are reported as found in the source, before the per-generator resampling and grid composition of Appendix~\ref{supp:dataset_processing}. Bimanual single-view is the head camera of the same episodes as bimanual multi-view, hence the identical counts.}
\label{tab:supp_dataset_cells}
\begin{tabular}{@{}lllrrrrrrr@{}}
\toprule
Embodiment & Track & Regime & Robot & Views & FPS & Resolution & Eps. & Pool & Instr. \\
\midrule
Single-arm & single & Mk. & RoboChallenge & 1 & 15 & 640$\times$480 & 120 & 120 & 2 \\
Single-arm & single & Non-mk. & RoboChallenge & 1 & 15 & 640$\times$480 & 120 & 120 & 2 \\
Single-arm & multi & Mk. & Franka & 3 & 15 & 320$\times$180 & 128 & 95,600 & 126 \\
Single-arm & multi & Non-mk. & Franka & 3 & 15 & 320$\times$180 & 129 & 95,600 & 128 \\
Bimanual & single & Mk. & ALOHA & 1 & 30 & 640$\times$480 & 150 & 150 & 34 \\
Bimanual & single & Non-mk. & ALOHA & 1 & 30 & 640$\times$480 & 150 & 150 & 17 \\
Bimanual & multi & Mk. & ALOHA & 4 & 30 & 640$\times$480 & 150 & 150 & 34 \\
Bimanual & multi & Non-mk. & ALOHA & 4 & 30 & 640$\times$480 & 150 & 150 & 17 \\
Humanoid & single & Mk. & GR1T2 & 1 & 20 & 1280$\times$800 & 130 & 130 & 118 \\
Humanoid & single & Non-mk. & GR1T2 & 1 & 20 & 1280$\times$800 & 130 & 130 & 87 \\
Humanoid & multi & Mk. & Airbot\_MMK2 & 4 & 30 & 640$\times$480 & 98 & 98 & 2 \\
Humanoid & multi & Non-mk. & Airbot\_MMK2 & 4 & 30 & 640$\times$480 & 150 & 376 & 6 \\
\midrule
\multicolumn{7}{@{}l}{\textbf{Total evaluation conditions (episode $\times$ camera track)}} & \textbf{1605} & & \\
\multicolumn{7}{@{}l}{\textbf{Distinct robot trajectories}} & \textbf{1305} & & \\
\bottomrule
\end{tabular}
\end{table}

\subsection{Data Preparation}
\label{supp:dataset_processing}

Preparation converts curated episodes into each generator's native input format.
The three generator families impose different spatial, temporal and
view-composition conventions, and we match each one's \emph{training}
configuration exactly, so that measured differences reflect the acceleration
method rather than an input-distribution shift.

\paragraph{Spatial resampling.}
Frames are resized to the resolution the corresponding checkpoint was trained
at: $320{\times}192$ per view for Ctrl-World, $384{\times}216$ per quadrant for
the DreamGen multi-view grid, $640{\times}480$ or $768{\times}432$ for
single-view DreamGen, and $640{\times}480$ for DreamDojo. Resizing is a direct
bilinear resample with no letterboxing or aspect-preserving crop, so source
aspect ratios are not preserved  GR1's $1280{\times}800$ egocentric frames
and DROID's $320{\times}180$ frames are both stretched into the target grid.
This is the transform the generators were fine-tuned under, but it means
absolute pixel-space metrics are not comparable \emph{across} settings of differing
source geometry. All comparisons in the paper are therefore made within a cell,
where every method sees pixel-identical inputs.

\paragraph{Temporal resampling.}
Source frame rates differ ($30$\,FPS for ALOHA and AIRBOT MMK2, $20$\,FPS for
GR1, $15$\,FPS for DROID and RoboChallenge). Each generator is fed at its
trained rate by integer frame striding, and the emitted clip declares the
resulting effective rate: $5$\,FPS for Ctrl-World, $16$\,FPS for DreamGen
($10$\,FPS for its $14$B DROID model), and $10$--$15$\,FPS for DreamDojo.
Integer striding cannot hit every target rate exactly from every source, so we
measure the realised frame rate, frame count and duration on the generated clips
rather than assuming them. Clips are generated to match their ground-truth
episode length, so duration varies within every cell
(Fig.~\ref{fig:supp_diversity}b).

\paragraph{Multi-view composition.}
None of the three generators natively emits $N$ independent synchronised
streams, so multi-view generation uses a single composed canvas. Ctrl-World
composes a horizontal $3{\times}1$ latent strip ($960{\times}192$), or a
$2{\times}2$ canvas ($1280{\times}192$) in its four-view variant. DreamGen
composes a $2{\times}2$ pixel grid ($768{\times}432$) with fixed slot
assignment: for bimanual ALOHA, the two external views on top and the left and
right wrist cameras below. When a cell offers only three active views, the
fourth quadrant is filled with a black frame and the prompt states that the
quadrant is inactive, so the model is not asked to hallucinate a camera; that
quadrant is excluded from all metrics. View subsetting is also a real reduction
 the Ctrl-World bimanual configuration drops one external view  so
\emph{Views} in Table~\ref{tab:supp_dataset_cells} upper-bounds the views actually
generated.

\paragraph{Instruction prompts.}
Instructions are neither authored by us nor model-generated: each is taken
verbatim from the source corpus metadata, preserving the string shown to the
teleoperator. Granularity is consequently heterogeneous  per-task corpora
carry one instruction per task ($34$ for bimanual Markovian, $2$ for humanoid
multi-view Markovian), whereas DROID and GR1 carry a near-unique instruction per
episode ($127$ and $118$). Upstream label noise is inherited as well: the GR1
subset substitutes scene identifiers and degenerate labels (\emph{invalid},
empty strings) for some instructions. We keep those episodes, since a world
model must still produce plausible motion under a degenerate prompt, but we
exclude the non-instruction strings when computing the taxonomy of
Appendix~\ref{supp:dataset_taxonomy}. Generators that expect a descriptive caption
receive the task string inserted into a fixed template declaring the grid
layout; the template is a deterministic substitution, with no per-episode
rewriting or VLM captioning.

\paragraph{Proprioception and actions.}
Action and state vectors are truncated to the dimensionality each checkpoint
expects ($14$-D joint-plus-gripper for bimanual ALOHA; $7$-D Cartesian or $8$-D
joint, plus gripper, for Ctrl-World single-arm) and strided by the same factor
as the frames, keeping action and observation indices aligned. Action-conditioned
models normalise proprioception with per-dimension $1$st/$99$th percentile
statistics, so every method within a cell shares one fixed normaliser. We note
that these percentiles are computed over the prepared episodes rather than held
out from them. This does not confound our comparisons, which are always between
methods sharing a normaliser, but absolute reconstruction numbers should not be
read as out-of-sample generalisation.

\paragraph{Filtering.}
An episode is included only when its trajectory data and all required view videos are
present and decodable. Episodes with missing streams are dropped rather than padded,
which is why per-setting video counts do not exactly equal episodes $\times$ views.
For example, three DROID episodes retain their trajectory files but contain only two
of the three required streams, leaving $128$ and $129$ episodes in the single-arm
multi-view settings rather than the $130$ initially sampled per regime. Trajectory
rows are re-aligned to the decoded frame count to prevent index drift.

We apply no success filter: DROID's success flag is never used as a selection
criterion, so failed demonstrations are retained. No minimum-length filter is applied,
yielding retained trajectories from $4$ to $3{,}000$ frames. A maximum-length filter
is applied only to RoboChallenge, where episodes exceeding a task-specific bound
($25$--$50$\,s) are excluded and the remainder are selected shortest-first, biasing
that setting toward shorter executions. Retaining failed and degenerate trajectories
is deliberate for an audit benchmark, as these cases are particularly informative
for evaluating failures in physical consistency.

\subsection{Task Taxonomy \& Instruction Prompts}
\label{supp:dataset_taxonomy}

\paragraph{Construction.}
\ours{} imposes no external ontology. We instead group the $509$ distinct
instructions present in the benchmark into canonical manipulation primitives by
splitting each instruction into clauses and mapping each clause's leading verb
onto a primitive. The lexicon is auditable: any unmatched verb is reported as
unmapped rather than silently bucketed, and the current lexicon leaves none of
the $509$ unmapped. The taxonomy is thus a grouping of \emph{measured strings},
not a label applied to episodes  $45$ distinct leading verbs collapse into
$14$ primitives.

\begin{figure}[t]
\centering
\includegraphics[width=\linewidth]{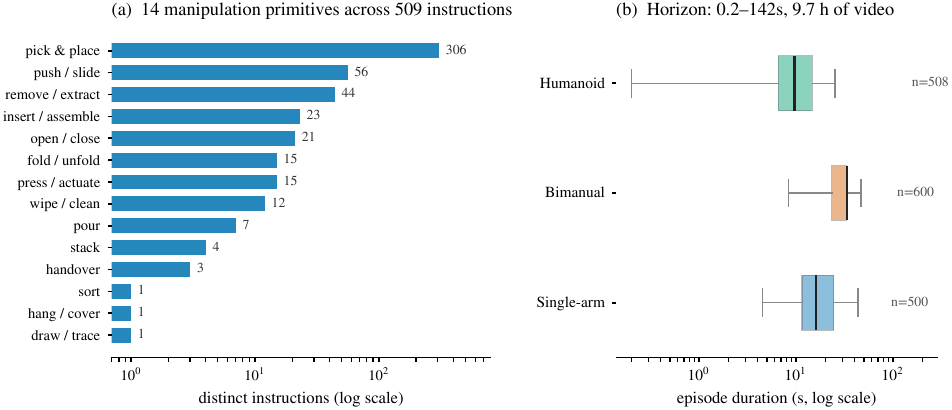}
\caption{\textbf{Instruction and horizon diversity.} (a) The $509$ distinct
instructions of \ours{}, grouped by leading verb into the $14$ manipulation
primitives of Appendix~\ref{supp:dataset_taxonomy}. The long tail is the point: pick \&
place alone covers $306$ instructions while six primitives have fewer than $10$,
so the taxonomy documents breadth and must not be read as a balanced factor.
(b) Episode duration per embodiment over all $1{,}608$ evaluation conditions,
computed as decoded frame count divided by \emph{source} frame rate ($n$
annotated per embodiment). Boxes span the interquartile range and outliers are
hidden, so the $0.2$--$142$\,s span in the panel title is wider than the
whiskers. Horizons vary by an order of magnitude \emph{within} every embodiment,
which is why every clip is generated to its own ground-truth length
(Appendix~\ref{supp:dataset_processing}) and all metrics are computed per episode rather
than over a fixed-length window.}
\label{fig:supp_diversity}
\end{figure}

\paragraph{Coverage.}
Counting each instruction by its first verb, the benchmark covers pick \& place
($306$), push/slide ($56$), remove/extract ($44$), insert/assemble ($23$),
open/close ($21$), fold/unfold ($15$), press/actuate ($15$), wipe/clean ($12$),
pour ($7$), stack ($4$), hand-over ($3$), and one instruction each for sort,
hang/cover and draw/trace. The distribution is strongly head-heavy
(Fig.~\ref{fig:supp_diversity}-a): pick \&
place alone accounts for $60\%$ of instructions and six primitives have fewer
than $10$. We therefore treat the taxonomy as evidence of coverage breadth
rather than as a balanced factor, and report no per-primitive metrics for the
tail. Since $12\%$ of instructions ($59/509$) request more than one primitive, a
multi-label count crediting every requested primitive totals $572$ mentions.

\paragraph{Compositional and scene structure.}
Compositional instructions concentrate in the non-Markovian regime by
construction, which is the intended contrast: \emph{``pull out a tissue, wipe
the stains on the table clean, then discard the tissue''} forces the model to
maintain phase state unrecoverable from the current frame. Scene diversity is
largest in humanoid single-view, where GR1 spans $38$ labelled scenes (retail,
bakery, electronics lab, gym and further indoor environments); the other corpora
are each recorded in one fixed table-top scene, so scene variation is not a
controlled factor of the grid.

\paragraph{Representative instructions.}
Appendix~\ref{tab:supp_dataset_taxonomy} lists all instructions for benchmark
settings defined at the task level. The remaining three settings use
episode-specific instructions. Representative examples include
\emph{``place the cup on the coaster''} and
\emph{``press the pink, blue, and green buttons in sequence''}
for RoboChallenge; \emph{``put the blue block in the green bowl''} and
\emph{``move the white mug to the left then use the green towel to wipe the
countertop''} for DROID; and \emph{``pick up yellow starfruit to bottom level of
wooden shelf''} for GR1.

{\small
\begin{longtable}{@{}l l >{\raggedright\arraybackslash}p{0.54\linewidth} r@{}}
\caption{Complete instruction set for benchmark settings whose instructions are defined at the task level, quoted verbatim from the source corpus. The two bimanual camera tracks share the same tasks and are therefore listed once. The remaining settings (DROID, RoboChallenge, and GR1) use episode-specific instructions and are summarized in Table~\ref{tab:supp_dataset_cells}.}
\label{tab:supp_dataset_taxonomy} \\
\toprule
Embodiment & Regime & Instruction & Eps. \\
\midrule
\endfirsthead
\multicolumn{4}{@{}l}{\emph{Table~\ref{tab:supp_dataset_taxonomy} continued from the previous page.}} \\
\toprule
Embodiment & Regime & Instruction & Eps. \\
\midrule
\endhead
\midrule
\multicolumn{4}{r@{}}{\emph{continued on the next page}} \\
\endfoot
\bottomrule
\endlastfoot
Bimanual & Markovian & close cardboard box & 1 \\
 &  & close toolbox & 7 \\
 &  & find hole and insert into gear & 4 \\
 &  & find insert large gear shaft & 6 \\
 &  & find insert small gear shaft & 4 \\
 &  & fit large gear shaft & 1 \\
 &  & fit small gear shaft & 2 \\
 &  & hit mark with hammer & 4 \\
 &  & hook cable 8pin & 1 \\
 &  & hook cable narrow 8pin & 1 \\
 &  & hook rubber shaft & 1 \\
 &  & insert gear onto shaft sliding & 6 \\
 &  & insert lan cable into the hub & 6 \\
 &  & insert large gear shaft & 8 \\
 &  & insert random shaft random medium gear & 11 \\
 &  & insert random shaft random small gear & 11 \\
 &  & insert rod board & 2 \\
 &  & insert rod moved board & 1 \\
 &  & insert shaft into gear & 1 \\
 &  & insert small gear shaft & 4 \\
 &  & insert usb cable fix parts & 2 \\
 &  & insert usb pcbox & 1 \\
 &  & insert usb port hub & 9 \\
 &  & insert washer shaft & 3 \\
 &  & insert washer shaft both hands & 4 \\
 &  & put cup moved red handle holder & 8 \\
 &  & remove box from case & 2 \\
 &  & remove box from case right arm & 1 \\
 &  & spoon red block bowl & 10 \\
 &  & spoon yellow block bowl & 16 \\
 &  & tape closed cardboard & 3 \\
 &  & turn towel over & 5 \\
 &  & wipe the desk with a blue cloth & 1 \\
 &  & wipe the desk with a sponge & 3 \\
\midrule
Bimanual & Non-Markovian & brush screws into dustpan human brush left hold & 6 \\
 &  & brush screws into dustpan left brush human hold & 2 \\
 &  & fold bath towel & 9 \\
 &  & fold big towel & 6 \\
 &  & fold blue towel & 23 \\
 &  & fold green towel & 13 \\
 &  & fold light blue towel & 8 \\
 &  & fold orange towel & 9 \\
 &  & fold towel assist & 7 \\
 &  & fold towel in random places & 11 \\
 &  & fold yellow towel & 11 \\
 &  & handover and wipe sponge & 13 \\
 &  & handover clear zip bag upright & 3 \\
 &  & handover metallic zip bag upright & 11 \\
 &  & place the chinese spoons on the cloth & 3 \\
 &  & put tape white case alternating arms & 8 \\
 &  & take out reel clear zip bag & 7 \\
\midrule
Humanoid & Markovian & push the pan to the edge of the table with your left hand and place it on the building block with your right hand. & 48 \\
 &  & push the yellow cart to the right with your left hand. & 50 \\
\midrule
Humanoid & Non-Markovian & Pick up the ballpoint pen and leave your handwriting on the paper. & 40 \\
 &  & pick up the pomegranate with left hand and put it in the storage box, and pick up the mango with right hand and put it in the storage box. & 33 \\
 &  & put the calculator box in the yellow bucket with your left hand and the power bank box in the yellow bucket with your right hand. & 56 \\
 &  & put the water bottle and handle into the cardboard box. & 21 \\
\end{longtable}
}

\subsection{Distribution of the Generated Acceleration Data}
\label{supp:dataset_rollouts}

\begin{figure}[thb]
\centering
\includegraphics[width=\linewidth]{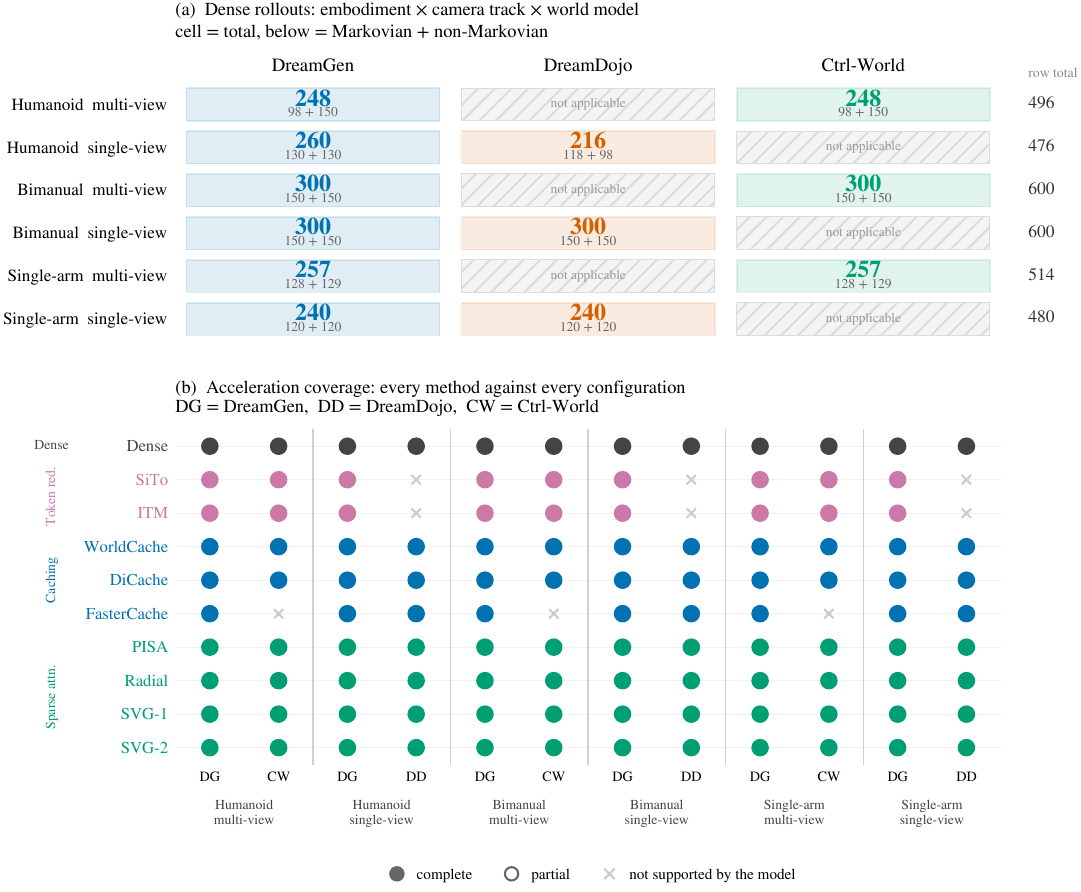}
\caption{\textbf{Distribution of generated rollouts.}
(a) Dense rollouts across embodiment, camera track, and world model, with each cell divided into Markovian and non-Markovian subsets.Hatched settings indicate model–camera combinations that are not supported: Ctrl-World is evaluated on multi-view tracks, DreamDojo on single-view tracks, and DreamGen on all six tracks, resulting in $12$ valid configurations.
(b) Coverage of the nine acceleration methods over the $12$ configurations. Crosses indicate unsupported backbone--method combinations.}
\label{fig:supp_coverage}
\end{figure}

The released benchmark includes both the curated evaluation episodes and the rollouts generated from them. Fig.~\ref{fig:supp_coverage} summarizes the coverage of these generated rollouts across model configurations and acceleration methods.

\paragraph{Configuration coverage.}
The benchmark contains $12$ valid model cell configurations and $3{,}166$ dense rollouts, including $1{,}605$ from DreamGen, $805$ from Ctrl-World, and $756$ from DreamDojo. The different per-model counts reflect their supported camera tracks rather than sampling imbalance.

\paragraph{Method coverage.}
Among the nine acceleration methods, six: WorldCache, DiCache, PISA, Radial Attention, and the two Sparse VideoGen variants cover all $12$ configurations, with $3{,}166$ rollouts each. SiTo and ITM exclude DreamDojo and contain $2{,}410$ rollouts each, while FasterCache excludes Ctrl-World and contains $2{,}361$ rollouts.

Overall, the benchmark contains $29{,}343$ generated rollouts: $3{,}166$ dense, $8{,}693$ caching-based, $12{,}664$ sparse-attention, and $4{,}820$ token-reduction rollouts. These totals describe dataset coverage rather than the relative performance of the acceleration families.

\clearpage
\section{Metrics Formalism \& Implementation}
\label{supp:metrics}

\subsection{Detailed Formulation of KIVA}
\label{supp:metrics_hallucination}
\textbf{Intuition and Overview.}
KIVA employs a diagnostic diffusion model trained to predict 3D robot keypoints from visual features. While real-world videos yield precise, low-variance keypoint reconstructions, hallucinated videos produce out-of-distribution features that cause diffusion predictions to diverge heavily across noise seeds. Averaging these estimates isolates the visual distortions into a measurable 3D trajectory. We then evaluate two physical properties on this trajectory: (i) \texttt{Rigidity}, which flags instances where the distance between predicted keypoints exceeds the robot's known hardware bone lengths, and (ii) \texttt{Jerk}, which measures anomalous acceleration to detect abrupt temporal discontinuities. These metrics are normalized against real-world video baselines and then thresholded at a level calibrated against human judgment.

\textbf{Clean targets and conditional denoising.}
Let $I_{1:T}$ denote a video sequence of $T$ frames. During training on real data, the recorded robot joint states $\mathbf{s}_{1:T}$ are processed via forward kinematics $\mathrm{FK}$(.) to produce exact 3D metric targets $\mathbf{X}^{\mathrm{FK}} = \operatorname{FK}(\mathbf{s}_{1:T}) \in \mathbb{R}^{T \times K \times 3}$ for $K$ keypoints. To bound the diffusion state space, we apply a fixed spatial affine transformation $\mathcal{W}(\cdot)$ defined by the physical limits of the robot's workspace, yielding the normalized clean coordinate tensor $\mathbf{x}_0 = \mathcal{W}(\mathbf{X}^{\mathrm{FK}}) \in [-1,1]^{T \times K \times 3}$. Concurrently, a frozen DINOv3 encoder~\citep{simeoni2025dinov3} extracts frame-level visual tokens $\mathcal{F}_\tau$ for each frame index $\tau \in \{1, \dots, T\}$.

We train a diffusion reader $g_\theta$ to denoise these coordinates conditioned on the visual features. For a sampled noise level $t$ and Gaussian noise $\boldsymbol\epsilon \sim \mathcal{N}(0, I)$, the forward process is defined as:
$$\mathbf{x}_t = \sqrt{\bar\alpha(t)}\,\mathbf{x}_0 + \sqrt{1-\bar\alpha(t)}\,\boldsymbol\epsilon; \textrm{ with } \hat{\mathbf{x}}_0 = g_\theta(\mathbf{x}_t, t, \mathcal{F}), $$ 
where $\bar\alpha(t)$ follows a cosine signal schedule. The network $g_\theta$ utilizes cross-attention to inspect each query's corresponding visual frame features $\mathcal{F}_\tau$, followed by a temporal encoder that shares context along the trajectory. The model is supervised by a Huber loss~\citep{huber1992robust}:
$$\mathcal{L}_{\mathrm{kp}} = \mathbb{E}_{\mathbf{x}_0, \mathcal{F}, t, \boldsymbol\epsilon} \left[\mathcal{L}_{\mathrm{Huber}}(\hat{\mathbf{x}}_0, \mathbf{x}_0)\right]$$
which averages the penalty over all coordinate entries without imposing explicit bone-length or smoothness constraints.

\textbf{Inference and candidate mean}. During evaluation on generated videos, the out-of-distribution visual features induce prediction variance. We perform $N$ independent DDIM sampling runs (with $N=5$), initialized from standard noise $\mathbf{x}_1^{(n)} \sim \mathcal{N}(0, I)$, while reusing the same conditioning representations. The mean clean estimate for keypoint $k$ at frame $\tau$ is averaged as: $\boldsymbol\mu_{\tau k} = \frac{1}{N}\sum_{n=1}^{N} \hat{\mathbf{x}}_{0, \tau k}^{(n)}.$ 

We then apply the inverse workspace transformation to map this mean trajectory back into un-normalized metric units, denoted as $\mathbf{P}_{\tau k}$, and compute downstream scores directly on these mean paths rather than relying on sample spread.

\textbf{Rigidity, jerk, and real-video references.}
We quantify physical violations directly on the estimated metric trajectories $\mathbf{P}_{\tau k}$. Given a set of structurally connected keypoint pairs $\mathcal{B}$, known nominal bone lengths $L_{ab}$ (in meters) for pairs $(a,b) \in \mathcal{B}$, and a frame interval $\Delta t$ (in seconds), we define the instantaneous rigidity error $r_\tau$ as the maximum kinematic link deformation:
$$r_\tau = \lambda \max_{(a,b)\in\mathcal{B}} \left|\|\mathbf{P}_{\tau a} - \mathbf{P}_{\tau b}\|_2 - L_{ab}\right|$$
To detect abrupt temporal discontinuities, we measure the maximum jerk (the third derivative of position with respect to time) across all keypoints $k$:
$$j(t) = \lambda \max_k \left\| \frac{d^3 \mathbf{P}_k(t)}{dt^3} \right\|_2 \approx \frac{\lambda}{\Delta t^3} \max_k \|\mathbf{P}_{\tau k} - 3\mathbf{P}_{\tau-1, k} + 3\mathbf{P}_{\tau-2, k} - \mathbf{P}_{\tau-3, k}\|_2$$

To contextualize these errors, for a given segment $S$, we compute the summary metrics $v_S^{\mathrm{rig}} = \operatorname{median}_{\tau\in S} r_\tau$ and $v_S^{\mathrm{jerk}} = \max_{\tau\in S} j_\tau$. We normalize these against a reference set $\mathcal{R}$ of real-video segments to report dimensionless ratios:
$$s_S^d = \frac{v_S^d}{\theta_d}, \qquad \theta_d = Q_{0.95}\!\left(\{v_R^d : R \in \mathcal{R}\}\right), \quad d \in \{\mathrm{rig}, \mathrm{jerk}\}$$
where $Q_{0.95}$ represents the 95th percentile. Crucially, the reference values $v_R^d$ are extracted by running the trained diffusion reader on the original real-world training data, establishing the baseline distribution of the model's natural reconstruction variance. Both ratios equal $1$ at the upper range of real recordings, so $s_S^d>1$ indicates that the error in $S$ exceeds the 95th percentile of the reader's values on real data. This level serves as the reference scale for the violation threshold below, rather than as the threshold itself.

\textbf{Threshold calibration.}
A segment's score is its largest detector ratio, $s_S=\max_{d\in\mathcal{D}_e} s_S^{d}$, with rigidity and jerk for single-arm and rigidity alone for bimanual and humanoid; $s_S=1$ is the real-video level.
Instead of fixed cuts at $1$ and $1.5$, we fit two cuts $\gamma_e^{1}\le\gamma_e^{2}$ per embodiment $e$, which map $s_S$ to low, medium or high.
The calibration set $\mathcal{C}_e$ holds nine rollouts of the human study, three from each of its low, medium and high strata, and every segment carries a three-point human label $h_S$ (two annotators, adjudicated on disagreement).
We choose the cuts that maximize exact agreement on the segments $\mathcal{S}_e$ of $\mathcal{C}_e$,
\begin{equation}
  (\gamma_e^{1},\gamma_e^{2}) = \arg\max_{\gamma^{1}\le\gamma^{2}} \frac{1}{|\mathcal{S}_e|}\sum_{S\in\mathcal{S}_e} \mathbb{1}\!\left[b(s_S;\gamma^{1},\gamma^{2})=h_S\right],
\end{equation}
which gives $(1.70, 2.10)$, $(1.00, 1.50)$ and $(0.32, 0.80)$ for single-arm, bimanual and humanoid.
Calibration rollouts are excluded from every reported agreement.

\myparagraph{Score aggregation.}
Results are reported per cell (model, camera track, embodiment, regime), with coverage given in Fig.~\ref{fig:supp_coverage}.
KIVA is computed on single-view rollouts over 16-frame segments (Fig.~\ref{sec:kiva}), and the cell score is the mean segment violation rate over rollouts (defined below).
Reference metrics are reported as the cell median with a 95\% bootstrap CI and valid count $n$.
Each accelerated rollout is compared with the dense rollout of the same episode, and for reference metrics a difference is considered supported only if the CI excludes zero, a Wilcoxon test gives $p<0.05$, and $n\geq 8$ (Sec.~\ref{supp:metrics_other}).
KIVA violation rates are compared directly without a significance test.
All comparisons are made within a cell, where every method receives pixel-identical inputs.

\myparagraph{Violation threshold.}
For KIVA, each segment $S$ receives a normalized error $s_S^d$ for $d\in\{\mathrm{rig},\mathrm{jerk}\}$, obtained by dividing its summary error by the 95th percentile $\theta_d$ of the reader's values on the real-video reference set $\mathcal{R}$, drawn from its training data; $s_S^d=1$ thus marks the upper range of real recordings.
We combine the two into a segment score $s_S=\max\big(s_S^{\mathrm{rig}}, s_S^{\mathrm{jerk}}\big)$.
Rather than flagging every segment above the real-video level, we calibrate a threshold $\gamma_e$ per embodiment $e$ against human judgment: three rollouts at low, medium, and high hallucination levels, defined by their segment scores relative to real video, are labeled by annotators, and $\gamma_e$ is set to best agree with these labels (Appendix~\ref{supp:metrics_hallucination}).
To summarize a rollout with $N$ segments $\{S_i\}_{i=1}^{N}$, we report its violation rate, the percentage of flagged segments,
\begin{equation}
  \mathrm{KIVA} = \frac{100}{N}\sum_{i=1}^{N} \mathbb{1}\!\left[s_{S_i}>\gamma_e\right],
\end{equation}
where $\mathbb{1}[\cdot]$ is the indicator function and a lower rate is better.
The cell score is the mean violation rate over the rollouts of the cell.
The same rate, computed with the same $\gamma_e$ on held-out real recordings of each embodiment, is reported alongside as the reference level.

\subsection{Detailed Formulation of RIGIS}
\label{supp:metrics_multiview}
This appendix provides the detailed mathematical formulation of \textbf{RIGIS} (\textbf{R}obotic \textbf{I}nter-view \textbf{G}eometry \& \textbf{I}ntrinsic-extrinsic \textbf{S}tability) expanding Sec.~\ref{sec:rigis}.

Under a rigid multi-camera setup, evaluating the cross-view physical consistency of a generated video rollout requires verifying two fundamental geometric requirements: (i)~all camera views must capture a single, unified 3D scene without structural hallucination, and (ii)~the camera calibration parameters must remain temporally invariant across the rollout. To assess these properties without reliance on ground-truth 3D geometry or VLM-based heuristics, RIGIS integrates two reference-free, bounded components in $[0, 1]$: \emph{geometric consistency} ($S_{\mathrm{geo}}$) and \emph{camera pose stability} ($S_{\mathrm{pose}}$).

All geometric primitives such as per-view depth maps and camera poses are estimated via a modified VGGT-$\Omega$ backbone \citep{wang2026vggtomega} (Fig.~\ref{fig:metrics}-b). Unless stated otherwise, we consider $V = 2$ synchronized camera views over $T$ timesteps $\tau \in \{1, \dots, T\}$ while frame $\tau=0$ serving as the pose reference anchor.

\paragraph{Notation.}
Let $\{X_\tau^{(v)}\}_{v=1}^{V}$ denote the RGB frames across $V$ views at timestep $\tau \in \{1, \dots, T\}$.
For each view $v$, the modified VGGT-$\Omega$ backbone outputs independent depth $D_\tau^{\mathrm{single},(v)}$, joint depth $D_\tau^{\mathrm{joint},(v)}$, and camera projection matrix $\mathbf{M}_\tau^{(v)} = \mathbf{K}_\tau^{(v)} [\mathbf{R}_\tau^{(v)} \mid \mathbf{t}_\tau^{(v)}]$, with intrinsics $\mathbf{K}_\tau^{(v)} \in \mathbb{R}^{3 \times 3}$ and extrinsics $[\mathbf{R}_\tau^{(v)} \mid \mathbf{t}_\tau^{(v)}] \in \mathbb{R}^{3 \times 4}$.
We write $\pi$ and $\pi^{-1}$ for pinhole projection and back-projection operations, respectively.
Geometric evaluation spans timesteps $\tau \in \{1, \dots, T\}$ while the initial conditioned frame at $\tau = 0$ serving as the pose reference anchor.

\noindent\textbf{Per-timestep geometry estimation}. 
VGGT-$\Omega$ routes tokens through an \emph{independent-attention} depth head and a \emph{joint-attention} depth head, plus a joint camera head (Fig.~\ref{fig:metrics}b, top).
Because independent and joint depths may disagree in absolute scale, we apply a per-view median log-ratio alignment on the center crop:
\begin{equation}
  \label{eq:scale_align}
  s_\tau^{(v)} = \exp\!\Bigl(\mathrm{median}\bigl(\log D_\tau^{\mathrm{joint},(v)} - \log D_\tau^{\mathrm{single},(v)}\bigr)\Bigr),
  \qquad
  D_\tau^{(v)} = s_\tau^{(v)}\, D_\tau^{\mathrm{single},(v)}.
\end{equation}
Aligned depths $D_\tau^{(v)}$ are used for unprojection and scoring.

\subsubsection{Geometric consistency}
\label{app:rigis_geo}
If a generated rollout depicts a structurally coherent 3D scene, pixels unprojected into world space and reprojected onto neighboring camera planes must remain consistent with the observed input frames.

\paragraph{Lift and reproject.}
Each aligned depth is unprojected to a world-coordinate point map
\begin{equation}
  \label{eq:unproject}
  \mathbf{P}_\tau^{(v)}(\mathbf{x}) = \pi^{-1}\!\bigl(\mathbf{x},\, D_\tau^{(v)}(\mathbf{x}),\, \mathbf{M}_\tau^{(v)}\bigr) \in \mathbb{R}^{3}.
\end{equation}
We discard flying pixels whose relative neighbor-depth discontinuity exceeds $0.1$.
Fused points are rasterized into every camera via a $z$-buffer, yielding rendered depth $\bar{D}_\tau^{(v)}$ and RGB $\bar{X}_\tau^{(v)}$.
An intersecting frustum mask $M_\tau^{(v)}$ retains pixels mutually visible across views.
Scoring uses the covisible ROI
\begin{equation}
  \label{eq:covisible_roi}
  \Omega_\tau^{(v)} = \bigl\{ \mathbf{x} \mid M_\tau^{(v)}(\mathbf{x}){=}1 \bigr\}.
\end{equation}

\paragraph{Per-view agreement.}
Within the overlap region $\Omega_\tau^{(v)}$, the RGB error $\Delta_X^{(v)}$ and depth residual $\Delta_D^{(v)}$ are computed as:
\begin{align}
  \Delta_X^{(v)} &= \sqrt{\tfrac{1}{|\Omega_\tau^{(v)}|}\sum_{\mathbf{x}\in\Omega_\tau^{(v)}} \|\bar{X}_\tau^{(v)}(\mathbf{x}) - X_\tau^{(v)}(\mathbf{x})\|_2^{2}} \label{eq:deltaX}\\
  \Delta_D^{(v)} &= \tfrac{1}{|\Omega_\tau^{(v)}|}\sum_{\mathbf{x}\in\Omega_\tau^{(v)}} \bigl|\bar{D}_\tau^{(v)}(\mathbf{x}) - D_\tau^{(v)}(\mathbf{x})\bigr| \label{eq:deltaD}
\end{align}
with depth scale reference $\tilde{d}^{(v)} = \mathrm{median}(D_\tau^{(v)})$.

Assuming RGB pixel values are normalized to $[0,1]$, the per-view color score $s_X^{(v)} \in [0,1]$ and depth score $s_D^{(v)} \in [0,1]$ are defined as:
\begin{equation}
  \label{eq:view_score}
  s_X^{(v)} = 1 - \Delta_X^{(v)},
  \qquad
  s_D^{(v)} = 1 - \left(\frac{\Delta_D^{(v)}}{\tilde{d}^{(v)}}\right)^{2},
  \qquad
  s^{(v)} = \tfrac{1}{2}\bigl(s_X^{(v)} + s_D^{(v)}\bigr),
\end{equation}
where $\tilde{d}^{(v)}$ denotes the depth scale of view $v$. The timestep score is obtained by taking the product over all views:
\begin{equation}
  \label{eq:sgeo_i_app}
  S_{\mathrm{geo}}^{(\tau)} = \prod_{v=1}^{V} s^{(v)} \in [0,1],
\end{equation}
ensuring that severe inconsistency in any single view penalizes the overall frame score.

\paragraph{Disjoint frames.}
A timestep $\tau$ with less than $2\%$ mutual frustum overlap (i.e., $|\Omega_\tau^{(v)}| / HW < 0.02$) in any view is marked disjoint, setting $S_{\mathrm{geo}}^{(\tau)} = 0$ and omitting it from pose-stability calculation.

\paragraph{Video-level geometric score.}
The overall geometric score $S_{\mathrm{geo}}$ is computed by averaging the frame-level scores across all $T$ timesteps over the episode:
\begin{equation}
  S_{\mathrm{geo}} = \frac{1}{T}\sum_{\tau=1}^{T} S_{\mathrm{geo}}^{(\tau)},
\end{equation}

\subsubsection{Camera pose stability}
\label{app:rigis_pose}

For physical robot setups with fixed camera rigs, intrinsic calibration parameters $\mathbf{K}_\tau^{(v)}$ and relative camera extrinsics $[\mathbf{R}_\tau^{(v)} \mid \mathbf{t}_\tau^{(v)}]$ must remain strictly stable over time. Temporal jitter or drift in estimated camera parameters reflects severe physical implausibility in the generated video.
Given camera parameter estimates $\{\mathbf{M}_\tau^{(v)}\}$ over non-disjoint timesteps, we evaluate the temporal variance of intrinsic and extrinsic parameters.

\paragraph{Intrinsic stability.}
For each view $v \in \{1, \dots, V\}$, we extract the intrinsic vector $\mathbf{k}_\tau^{(v)} = (f_x, f_y, c_x, c_y)_\tau^{(v)}$ and normalize it relative to the anchor frame: $\tilde{\mathbf{k}}_\tau^{(v)} = \mathbf{k}_\tau^{(v)} \oslash (|\mathbf{k}_1^{(v)}| + \delta)$, where $\oslash$ and $|\cdot|$ denote element-wise division and absolute value, respectively. Let $\sigma_K^{(v)}$ be the mean temporal standard deviation across the four normalized channels. The overall intrinsic stability is defined as:
\begin{equation}
  \label{eq:sint}
  S_{\mathrm{int}} = \frac{1}{V}\sum_{v=1}^{V} \exp\!\bigl(-\sigma_K^{(v)}\bigr).
\end{equation}

\paragraph{Extrinsic stability.}
Fixing the primary camera ($v{=}1$) at the origin, we evaluate relative camera motion for relocating views $v \in \{2, \dots, V\}$ with extrinsic matrices $\mathbf{E}_\tau^{(v)} = [\mathbf{R}_\tau^{(v)} \mid \mathbf{t}_\tau^{(v)}]$. The relative rotation with respect to the anchor frame is $\mathbf{R}_\tau^{\mathrm{rel}} = \mathbf{R}_\tau^{(v)} (\mathbf{R}_1^{(v)})^{\top}$, with geodesic rotation angle given by:
\begin{equation}
  \label{eq:theta}
  \theta_\tau^{(v)} = \frac{180}{\pi} \cdot \arccos\!\Bigl(\mathrm{clip}_{[-1,1]}\bigl(\tfrac{\mathrm{tr}(\mathbf{R}_\tau^{\mathrm{rel}})-1}{2}\bigr)\Bigr).
\end{equation}
Let $\sigma_R^{(v)} = \mathrm{std}(\{\theta_\tau^{(v)}\})$ and $\sigma_t^{(v)}$ be the mean temporal standard deviation of translation $\mathbf{t}_\tau^{(v)}$ across all three spatial axes. The extrinsic stability score averaged over all relocating views is:
\begin{equation}
  \label{eq:sext}
  S_{\mathrm{ext}} = \frac{1}{V-1}\sum_{v=2}^{V} \tfrac{1}{2}\left(\mathrm{e}^{-\sigma_R^{(v)}/5} + \mathrm{e}^{-\sigma_t^{(v)}}\right).
\end{equation}

\paragraph{Combined pose score.}
The total camera pose stability score aggregates intrinsic and extrinsic consistency:
\begin{equation}
  \label{eq:spose_app}
  S_{\mathrm{pose}} = \tfrac{1}{2}\bigl(S_{\mathrm{ext}}+S_{\mathrm{int}}\bigr) \in [0,1].
\end{equation}
If fewer than two non-disjoint frames remain in the rollout sequence, we set $S_{\mathrm{pose}} = 0$.

\subsubsection{RIGIS score}
\label{app:rigis_final}
For an individual video rollout, the raw RIGIS score is defined as $S = \frac{1}{2}(S_{\mathrm{geo}} + S_{\mathrm{pose}}) \in [0, 1]$. While individual scores remain unnormalized at the instance level, the aggregate benchmark score for a target robot morphology is normalized by the mean real-video baseline, i.e., $S_{\mathrm{target}} = \bar{S} / \bar{S}_{\mathrm{real}}$, where higher values indicate superior multi-view consistency. $\bar{S}_{\mathrm{real}}$ denotes the mean score of the real-video subset for the target morphology, mitigating passive camera setup variations on the attainable score range.


\subsection{Additional Metric Details}
\label{supp:metrics_other}

This section specifies the four reference-free metrics reported in
Sec. \ref{sec:evaluation_protocol}. We first define the common preprocessing pipeline shared across metrics, followed by the formulation, implementation, and validity
conditions of each metric. We additionally describe the aggregation protocol,
motion-confound analysis, and three auxiliary quantities that were computed
during development but excluded from the final benchmark.

\subsubsection{Visual Integrity.}
We measure frame-level visual quality using CLIP-IQA+~\citep{wang2023exploring}.
For each sampled frame, CLIP~\citep{radford2021learning} estimates the probability
of the prompt ''Good photo.'' relative to ''Bad photo.'', yielding a score in
$[0,1]$. Rather than using the temporal mean of the original metric, we report
the $10^{th}$ percentile across frames, multiplied by $100$.

\begin{figure}[h]
\centering
\includegraphics[width=\linewidth]{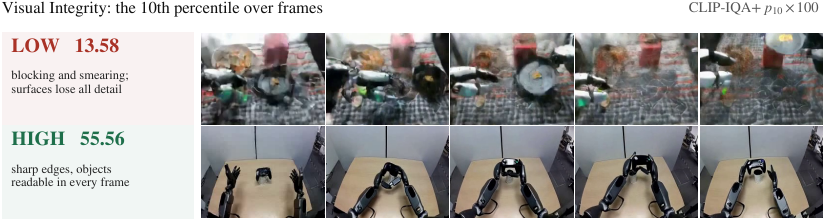}
\caption{\textbf{What Visual Integrity reacts to.} A low- and high-scoring clip are shown with five evenly spaced frames each; the value on the left is the score reported in the tables. Both examples are from DreamGen humanoid (low: ITM, \texttt{static}; high: dense, \texttt{head}). Because this metric fails the ground-truth-is-ceiling audit, it should not be used to rank models. The comparison is therefore intended only to illustrate the type of visual degradation captured by the metric.}
\label{fig:nr_intuition_vi}
\end{figure}

\myparagraph{Adaptation.}
The percentile departs from the source metric, which uses a temporal mean. A rollout that degenerates only briefly, a characteristic autoregressive failure, can still retain a healthy mean, whereas $p_{10}$ tests whether \emph{nearly all} generated frames remain usable. The mean is nevertheless retained as a diagnostic. Fig.~\ref{fig:nr_intuition_vi} illustrates the intuition: the low-quality clip has lost every surface in the scene while remaining a plausible ``video'', precisely the type of failure that temporal averaging would dilute.

\subsubsection{Text Alignment.}
\label{supp:metric_text}

\myparagraph{What it measures.} Whether the rollout still depicts the task that was
commanded, rather than drifting into a generically plausible continuation of the
conditioning frame. This is the only column that reads the instruction.

\myparagraph{Definition and backend.}
CLIPScore~\citep{hessel2021clipscore}, i.e.\ $\max(100\cdot\cos(E_I,E_C),0)$, averaged
over sampled frames, against the \textbf{task clause only}.

\myparagraph{Adaptation.}
Extracting this clause is important for our benchmark because DreamGen prompts append a description of the $2\times2$ grid layout, which is incompatible with a cropped single view and can push the task text beyond CLIP's $77$-token limit. This metric is applicable only to text-conditioned models. Image-only models never receive the instruction and are therefore marked as \texttt{None} rather than assigned a low score. Per-model real-video anchors are treated as text-conditioned because they correspond to real recordings of the same instructed task.

\begin{figure}[h]
\centering
\includegraphics[width=\linewidth]{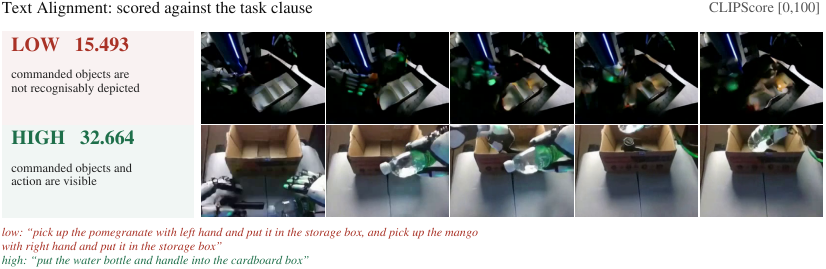}
\caption{\textbf{What Text Alignment reacts to.} A low- and high-scoring clip are shown, each evaluated against its own task clause printed below the frame strip. The two rows are not directly comparable because CLIPScore exhibits substantial caption-dependent bias. Accordingly, all ranking claims for this metric use paired deltas under the same prompt. Both examples are humanoid \texttt{static} views (low: Ctrl-World 4-view with SiTo; high: DreamGen dense).}
\label{fig:nr_intuition_text}
\end{figure}

\subsubsection{Fr\'{e}chet Video Distance}
\label{supp:metric_fvd}

Fr\'{e}chet Video Distance (FVD)~\citep{unterthiner2018towards} measures the distance
between the distributions of generated and real videos in I3D feature space and is
therefore evaluated at the set level rather than per rollout. We extract Kinetics-400
I3D features from consecutive $16$-frame windows resized to $224\times224$. For each
embodiment $\times$ camera $\times$ regime, all models and acceleration methods share
the same fixed real-video reference pool. Multi-view predictions are evaluated per
camera and averaged over cameras common to all configurations. Videos are sampled at a
common $5$\,fps before feature extraction, and duplicate reference clips are removed.
Because FVD is sensitive to sample count and evaluation protocol, we use it only for
comparisons sharing the same reference and sampling configuration, rather than for
absolute comparisons across benchmark settings.

\begin{figure}[thb]
\centering
\includegraphics[width=\linewidth, height=0.5\linewidth]{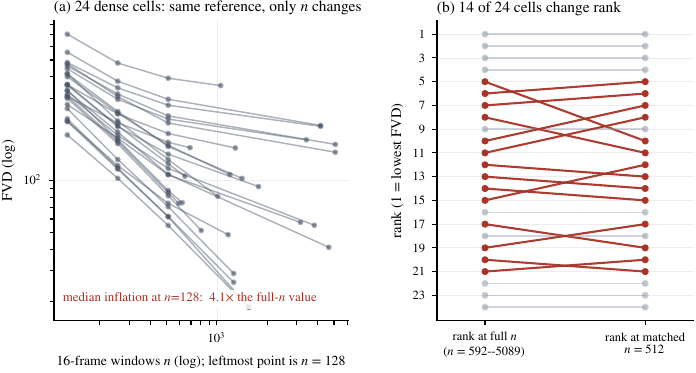}
\caption{\textbf{FVD sensitivity to sample size.}
(a) Dense settings re-scored against the same reference while varying the number of
windows $n$. (b) Matching $n$ changes the ranking of $14/24$ settings, indicating
substantial estimator sensitivity at small sample sizes.}
\label{fig:ref_intuition_fvd}
\end{figure}
\subsubsection{PSNR}
\label{supp:metric_psnr}

PSNR is used as the representative pixel-level fidelity metric in the aggregate
retention index, while SSIM and LPIPS are retained for per-cell analysis to avoid
over-weighting highly correlated appearance measures. Following prior acceleration
benchmarks, we report PSNR relative to the corresponding dense rollout rather than
treating real video as a PSNR ceiling. Because PSNR is logarithmic, degradation
severity is reported in dB when comparing across settings, while normalized retention is
used only within a shared dense baseline.

\clearpage
\section{World Model Training \& Acceleration Implementation}
\label{supp:implementation_details}

This appendix documents how the benchmark's generators were produced and
configured. Appendix~\ref{supp:impl_finetuning} covers the finetuning of the three base
world models, while Appendix~\ref{supp:impl_acceleration} covers the configuration of the nine
acceleration methods and the protocol under which their speedups were measured.
All reported quantities are taken from the configurations that produced the
released checkpoints and from direct measurement of the resulting corpora and
rollouts; quantities that neither source determines are marked as such rather
than replaced by defaults.

\subsection{Base World Model Finetuning Protocols}
\label{supp:impl_finetuning}

\paragraph{Scope and rationale.}
\ours{} evaluates acceleration methods within fixed base world models rather than comparing the base models themselves. Their training protocols are reported because the corresponding dense models define the reference for all speedup and quality-retention measurements. We use three families: Ctrl-World, an SVD image-to-video U-Net, and DreamGen and DreamDojo, both Cosmos-Predict2 diffusion transformers, spanning the $12$ configurations in Table~\ref{tab:supp_generation_cfg}.

\paragraph{Benchmark configuration selection.}
Ctrl-World generated additional configurations beyond those included in the reported benchmark grid. For bimanual and humanoid multi-view episodes, we generated both a $3{\times}1$ three-camera latent strip and a $2{\times}2$ four-camera latent canvas. We report only the four-view variant because it supersedes the three-view layout on the same episodes, while using an additional camera, a separate checkpoint, and a distinct acceleration profile. For example, WorldCache speedup on bimanual increases from $1.27\times$ to $1.81\times$. Reporting both layouts would duplicate the same episodes under two configurations of the same model. Single-arm retains the three-view strip because no four-view counterpart exists. Accordingly, the reported Ctrl-World layout is embodiment-specific and is identified explicitly in each table. The excluded rollouts and run scripts remain in the release for reproducibility, while the main benchmark uses the $12$ configurations in Table~\ref{tab:supp_generation_cfg}.

\paragraph{Training corpus construction.}
Each model family requires a different input representation, summarized in
Table~\ref{tab:supp_training_data}. Ctrl-World uses precomputed latents: $30$\,fps
multi-view recordings are subsampled to $5$\,fps, resized to $320{\times}192$ per
view, VAE-encoded, and paired with a $12$-D end-effector pose action. Although the
raw robot actions are $36$-D and $54$-D, the shared $12$-D pose interface enables
a common model across embodiments. DreamGen operates on RGB clips with text
conditioning. Its four camera streams are composed into a $2{\times}2$ grid,
allowing a single-view video backbone to process multi-view data without
architectural changes; text embeddings are computed per clip, and long recordings
are segmented into overlapping windows. DreamDojo consumes episodes directly and
uses conditioning from a frozen latent action model rather than proprioceptive
signals.

\paragraph{Adaptation regime.}
All three model families are fully finetuned without LoRA or other parameter-efficient adapters. The released checkpoints therefore contain full
model weights, with per-configuration identifiers reported in Table~\ref{tab:supp_generation_cfg}.

\paragraph{Hyperparameters.}
Table~\ref{tab:supp_finetune_cfg} gives the protocol of each family, read out of the
configuration that produced its checkpoints. The three recipes are \emph{not}
matched, and we do not claim they are: peak learning rates span an order of
magnitude ($1{\times}10^{-5}$ for Ctrl-World, $2^{-14.5}$ for DreamGen,
$1{\times}10^{-4}$ for DreamDojo), the U-Net trains in \texttt{fp16} while both
transformers train in \texttt{bf16}, and clip lengths and frame rates differ by
construction because each backbone inherits the released recipe of its upstream
model. We therefore compare acceleration methods down a column and never across
columns; a cross-family quality gap in this paper reflects the generators as
published, not a controlled comparison of architectures.
\paragraph{Compute and sampling configuration.}
Ctrl-World is trained with \textsc{Accelerate} data parallelism using a micro-batch size of $1$, while DreamGen and DreamDojo use FSDP across all ranks. DreamGen additionally uses context parallelism of degree $2$ during training, which is disabled for accelerated inference because the evaluated caching and sparse-attention methods do not support it. All rollouts and timing measurements are performed on a single A100-80GB (Appendix~\ref{supp:impl_acceleration}). Sampling settings, including denoising steps, guidance scale, and seeds, are reported in Table~\ref{tab:supp_generation_cfg}. DreamGen and DreamDojo use deterministic seed schedules, whereas Ctrl-World is unseeded; therefore, small differences between Ctrl-World methods may reflect sampling variance. We report single-run results without seed-wise error bars and interpret within-family comparisons accordingly.

\begin{figure}[t]
\centering
\includegraphics[width=\linewidth]{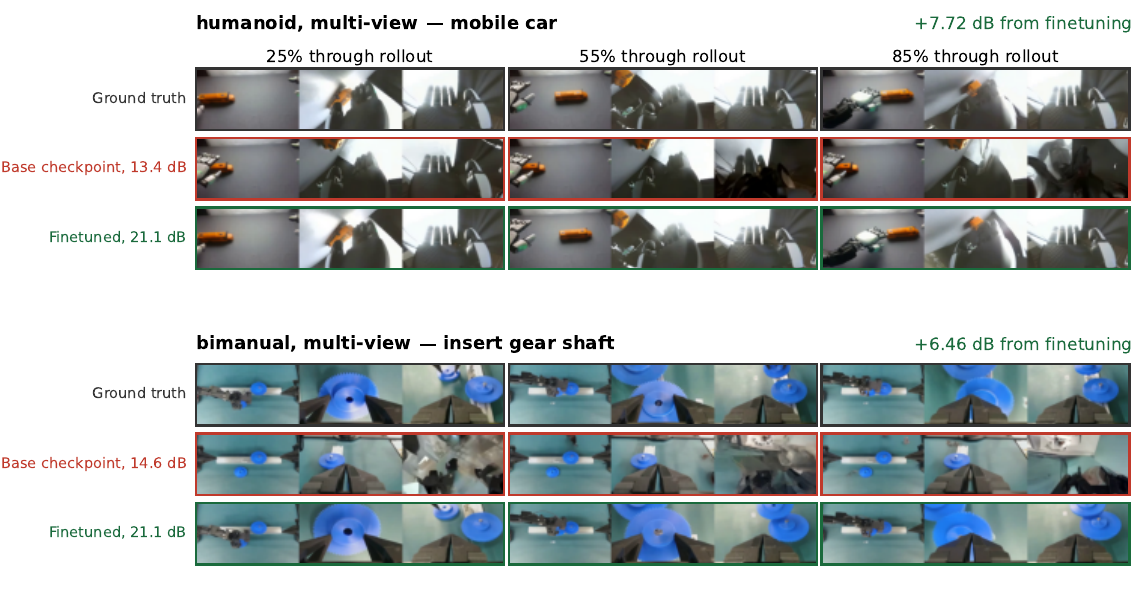}
\caption{\textbf{Effect of finetuning on matched episodes.} Ground truth, unfine-tuned, and finetuned rollouts are shown at matched time points with all camera views. The unfine-tuned model retains plausible appearance but progressively loses task-relevant structure, including object position and cross-view consistency. Finetuning mitigates these failures and therefore defines the dense reference used for quality-retention evaluation. Reported PSNR values correspond to the full episode.}
\label{fig:supp_finetune_qual}
\end{figure}

\begin{table}[t]
\centering
\small
\setlength{\tabcolsep}{4pt}
\caption{\textbf{Generation configurations of the $14$ dense benchmark settings.} Each row defines the dense baseline against which acceleration methods in the corresponding cell are evaluated. \emph{Size} denotes the backbone parameter count, \emph{Adaptation} indicates whether the model is finetuned for the corresponding robot setting or used from a released checkpoint, \emph{Steps} is the number of denoising steps, $g$ is the guidance scale, and $n$ is the number of released rollouts. Seed handling differs across model families: Ctrl-World is unseeded, DreamGen uses seed $0$ (offset by segment index for autoregressive multi-view rollouts), and DreamDojo uses a per-chunk seed.}
\label{tab:supp_generation_cfg}
\begin{tabular}{@{}llccrrlr@{}}
\toprule
Model & Cell & Size & Adaptation & Steps & $g$ & Seed & $n$ \\
\midrule

\textbf{Ctrl-World}
& single-arm MV
& 1.5B
& Released
& 50 & 1.0
& unseeded
& 257 \\

& bimanual MV
& 1.5B
& Fine-tuned
& 50 & 1.0
& unseeded
& 300 \\

& bimanual MV (4-view)
& 1.5B
& Fine-tuned
& 50 & 1.0
& unseeded
& 300 \\

& humanoid MV
& 1.5B
& Fine-tuned
& 50 & 1.0
& unseeded
& 248 \\

& humanoid MV (4-view)
& 1.5B
& Fine-tuned
& 50 & 1.0
& unseeded
& 248 \\

\midrule

\textbf{DreamGen}
& single-arm MV
& 14B
& Released
& 35 & 7
& 0
& 92 \\

& bimanual MV
& 2B
& Fine-tuned
& 35 & 7
& 0 + segment index
& 300 \\

& humanoid MV
& 2B
& Fine-tuned
& 35 & 7
& 0 + segment index
& 248 \\

& single-arm SV
& 2B
& Fine-tuned
& 35 & 7
& 0
& 240 \\

& bimanual SV
& 2B
& Fine-tuned
& 35 & 7
& 0
& 300 \\

& humanoid SV
& 2B
& Fine-tuned
& 35 & 7
& 0
& 260 \\

\midrule

\textbf{DreamDojo}
& single-arm SV
& 2B
& Fine-tuned
& 35 & 1.5
& per-chunk index
& 240 \\

& bimanual SV
& 2B
& Fine-tuned
& 35 & 1.5
& per-chunk index
& 300 \\

& humanoid SV
& 2B
& Released
& 35 & 1.5
& per-chunk index
& 216 \\

\bottomrule
\end{tabular}
\end{table}
\begin{table}[t]\centering\small
\setlength{\tabcolsep}{5pt}
\caption{\textbf{Training corpus construction for the finetuned DreamGen checkpoints.}
Each block traces a released checkpoint through its preprocessing pipeline. \emph{Clips} denotes generated video files, while \emph{usable} counts those with at least the $93$ frames required for training. The humanoid corpus retains only $132/248$ usable clips at $10$\,fps, whereas the bimanual corpus retains all clips at $30$\,fps. Overlapping $10$\,s windows with a $5$\,s stride and re-encoding to $16$\,fps increase usable samples by $2.8\times$ and $5.6\times$, respectively, while matching the base checkpoint frame rate.}
\label{tab:supp_training_data}
\begin{tabular}{@{}llrrrrr@{}}\toprule
Corpus & Stage & fps & Clips & Usable & \%\ usable & Median fr. \\
\midrule
humanoid MV & $2{\times}2$ grid stitch & $10$ & $248$ & $132$ & $53$ & $97$ \\
 & segment & $10$ & $416$ & $300$ & $72$ & $100$ \\
 & re-encode 16\,fps & $16$ & $373$ & $373$ & $100$ & $160$ \\
\addlinespace[2pt]
bimanual MV & $2{\times}2$ grid stitch & $30$ & $300$ & $300$ & $100$ & $1000$ \\
 & segment & $30$ & $1694$ & $1694$ & $100$ & $300$ \\
 & re-encode 16\,fps & $16$ & $1694$ & $1694$ & $100$ & $162$ \\
\addlinespace[2pt]
humanoid SV & transcode 16\,fps & $20$ & $260$ & $220$ & $85$ & $208$ \\
\addlinespace[2pt]
bimanual SV & transcode 16\,fps & $16$ & $1710$ & $1710$ & $100$ & $107$ \\
\addlinespace[2pt]
single-arm SV & transcode 16\,fps + segment & $16$ & $703$ & $703$ & $100$ & $102$ \\
\bottomrule
\end{tabular}
\end{table}

\begin{table*}[t]
\centering
\small
\setlength{\tabcolsep}{7pt}
\caption{\textbf{Finetuning configurations of the three base world models.}
We retain each model family's original training recipe rather than enforcing
matched hyperparameters across architectures. All adaptations use full-model
finetuning without parameter-efficient adapters.}
\label{tab:supp_finetune_cfg}

\resizebox{\textwidth}{!}{%
\begin{tabular}{@{}l p{4.2cm} p{5.5cm} p{4.5cm}@{}}
\toprule
Configuration & Ctrl-World & DreamGen & DreamDojo \\
\midrule

\multicolumn{4}{@{}l}{\textit{Model}} \\[2pt]

Backbone
& SVD image-to-video U-Net
& Cosmos-Predict2 2B / 14B DiT
& Cosmos-Predict2 2B DiT \\

Adaptation
& Full finetuning
& Full finetuning
& Full finetuning \\

Conditioning
& 12-D actions + embodiment ID
& Text prompt
& Latent actions \\

\addlinespace[3pt]
\multicolumn{4}{@{}l}{\textit{Optimization}} \\[2pt]

Optimizer
& AdamW
& AdamW
& FusedAdamW\\

Peak learning rate
& $1\times10^{-5}$
& $4.32\times10^{-5}$
& $1\times10^{-4}$ \\

Weight decay
& $0.01$
& $0.1$
& $0.1$ \\

LR schedule
& Constant
& LambdaLinear + 1k warm-up
& LambdaLinear \\

Precision
& fp16
& bf16
& bf16\\

\addlinespace[3pt]
\multicolumn{4}{@{}l}{\textit{Training setup}} \\[2pt]

Distributed training
& Accelerate DDP
& FSDP + context parallel $2$
& FSDP\\

Resolution
& $320\times192$
& $384\times216$ / $640\times480$ / $768\times432$
& $640\times480$ \\

Training clip
& $6$ history + $5$ prediction frames
& $93$ frames
& -- \\

Frame rate
& $5$ fps
& $10/16$ fps
& $10/15$ fps \\

Max. iterations
& $100$k
& $120$k
& $400$k \\

Training seed
& $42$
& $0$
& $42$ \\
\bottomrule
\end{tabular}%
}
\end{table*}

\subsection{Diffusion Acceleration Method Configurations}
\label{supp:impl_acceleration}

\paragraph{What is configured, and where.}
Nine methods are benchmarked in three families:
\emph{token reduction} (SiTo, ITM), which shrinks the token sequence entering
attention; \emph{step caching} (WorldCache, DiCache, FasterCache), which reuses a
previous step's residual and skips whole network evaluations; and \emph{sparse
attention} (PISA, SVG, SVG2, Radial), which restricts the attention operator
itself.

\paragraph{How a speedup is measured.}
A speedup is defined as the ratio between the generation time of a dense baseline and that of its accelerated counterpart, with seven choices fixing the measurement protocol. \emph{Clock} --- we use each model family's native generation-time instrumentation: per-episode \texttt{wall\_time\_s} for Ctrl-World, reported total inference time for DreamGen, and per-segment generation time for DreamDojo. We do not use end-to-end process wall clock, which additionally includes model loading, VAE decoding, metric computation, and other fixed overheads that would systematically push ratios toward $1$. \emph{Dense denominator} --- every speedup is normalized by a fresh dense run from the \emph{same cell}. Concretely, for each accelerated run we rerun the corresponding dense model on the same episode, from the same split, with the same backbone, embodiment, view setting, task-dynamics regime, and pinned generation configuration. Thus, if $T_{\mathrm{dense}}$ and $T_{\mathrm{method}}$ denote the paired generation times, we report
$S=T_{\mathrm{dense}}/T_{\mathrm{method}}$.
The denominator is therefore never a stored global constant, an average taken from another cell, or a previously recorded dense timing. This paired construction is important because dense generation time itself varies across models, embodiments, view configurations, and episode lengths; reusing a denominator from another setting would confound acceleration gains with baseline runtime differences. \emph{Isolation}: all runs use one GPU and are issued sequentially, with context parallelism disabled; the timing harness rejects a device with another resident process, since measurements under contention characterize the contention rather than the acceleration method. \emph{Sample}: accelerated and dense runs use the same episodes from both task-dynamics regimes, because generation cost scales with episode length and a Markovian-only sample could misrepresent methods whose savings are length-dependent. \emph{Warm-up}: the first segment or chunk is reported separately from the steady-state ratio $S_{\text{ss}}$, because several methods incur a one-off attention-profiling or $k$-means cost that would otherwise make the dense baseline appear artificially favorable; $S_{\text{ss}}$ better reflects the cost observed over a long rollout. \emph{Absence}: if either member of the paired timing is unavailable, no speedup is inferred: the entry is reported as \texttt{--} rather than interpolated from a neighbouring cell. \emph{Aggregation} --- the ratio is formed between two single generation-time figures of the same cell, $S=T_{\text{dense}}/T_{\text{method}}$, each reduced from that family's own repeated unit of work rather than from a per-episode total. Reducing timing at the unit level prevents episode length from contaminating the speedup ratio. We verified this empirically rather than assuming it: episode-level totals implicitly scale with the number of generated units and can therefore distort comparisons across clips of different lengths. For example, on Ctrl-World humanoid multi-view, the same WorldCache configuration yielded $0.64\times$ on a 37-frame clip but $1.71\times$ on a 61-frame clip, a difference attributable to episode length rather than to the caching mechanism itself. We therefore compute speedup from matched unit-level timings and apply no additional normalization by episode count or duration.

\paragraph{Timing provenance and coverage.}
The speedup column combines two distinct timing passes, which are reported separately rather than merged. Most ratios come from each method's per-cell tuning trial ($\dagger$), where timing is matched within a cell but the number of evaluated episodes varies across backbones: two episodes for Ctrl-World, one for DreamDojo, and two segments or chunks for DreamGen. Consequently, small differences across cells may partly reflect protocol variation, and these ratios should be interpreted as within-cell comparisons rather than as a globally calibrated cross-cell scale.

To provide a uniform reference, we additionally run a shared timing pass ($\ddagger$): one common episode per cell, with every method and its dense baseline timed back-to-back on the same idle GPU. This pass covers all $12$ dense baselines and $105$ method--cell scripts, corresponding to $117$ of the $120$ possible pairs; the remaining $3$ have no executable run script. The full pass is estimated to require $8.3$ GPU-hours. At the time of writing, it has completed for one cell, DreamDojo humanoid single-view. This result already illustrates why the two timing sources are kept separate: under the uniform protocol, DiCache reaches $1.56\times$, the first measured caching speedup on a DreamDojo backbone, whereas WorldCache reaches only $1.08\times$ and FasterCache is net-negative at $0.91\times$, an ordering not observed in the tuning records for this cell.

Timing coverage remains incomplete. Among the $56$ method--cell pairs with reported quality results, $6$ lack recorded generation times: the three caching methods on DreamDojo single-arm single-view, and WorldCache, DiCache, and ITM on Ctrl-World single-arm multi-view. These entries are therefore reported as unavailable rather than imputed, explaining why several rows contain quality-retention scores but no corresponding speedup.

\paragraph{Aggregate retention index.}
To summarize acceleration quality across metrics, we define
$\mathcal{K}=\{\text{Visual Integrity},\,\text{PSNR},\,\text{Text Alignment},\,\text{FVD}\}$.
For accelerator $a$ on backbone $b$, retention in metric $k$ is measured relative
to the dense rollout of the same backbone:
\begin{equation}
r_k(a,b)=\frac{s_k(a,b)}{s_k(\emptyset,b)}
\quad\text{for }k\neq\text{FVD},
\qquad
r_{\text{FVD}}(a,b)=\frac{s_{\text{FVD}}(\emptyset,b)}
{s_{\text{FVD}}(a,b)},
\end{equation}
where FVD is inverted because lower values are better. The aggregate retention is
\begin{equation}
Q(a,b)=\frac{1}{|\mathcal{K}|}\sum_{k\in\mathcal{K}} r_k(a,b).
\end{equation}
By construction, $Q(\emptyset,b)=1$, so $Q$ measures quality retention relative
to the computation replaced by the accelerator rather than to an external ideal.
Values above $1$ are valid when an accelerated model exceeds its dense baseline
on one or more metrics. We do not normalize against real video because PSNR has
no finite real-against-real reference. When a metric is unavailable, as for Text
Alignment on DreamDojo single-view, $Q$ is computed over the remaining metrics
and the omission is indicated explicitly.

\paragraph{Autoregressive rollout depth.}
\label{supp:metrics_chunk_depth}
Each episode is generated as a sequence of chunks, with only the first conditioned
on real frames. We define rollout depth as
$C=\lceil(T-T_{\mathrm{cond}})/\Delta\rceil$, where $\Delta$ is the generation
stride. We infer $\Delta$ from the empirical clip-length periodicity:
$(T-1)\equiv0\pmod{92}$ for all $799$ DreamGen clips,
$(T-1)\equiv0\pmod{12}$ for all $540$ DreamDojo bimanual and single-arm clips,
and $T\equiv0\pmod{49}$ for $210/216$ DreamDojo humanoid clips.
Because Markovian and non-Markovian regimes draw from different task pools,
rollout depth is not matched across regimes (Tab.~\ref{tab:chunk_depth}).
The two deepest non-Markovian configurations are also the two in which KIVA
degrades, although this cross-configuration association cannot isolate depth
from regime. Within clips, however, severity
$\max(s_S^{\mathrm{rig}},s_S^{\mathrm{jerk}})$ increases with chunk index in
$130/193$ multi-chunk clips (mean $\rho=+0.20$, sign-test $p<10^{-3}$),
consistent with error accumulation over rollout depth. The trend is not universal:
for single-arm single-view, whose first chunk is already $1.33$ above the
violation boundary, severity instead decreases with depth ($\rho=-0.10$).

\begin{table}[h]
\centering
\scriptsize
\caption{Median rollout depth $C$ per clip (range in brackets) against KIVA, the percentage of invalid $16$-frame segments, ordered by the non-Markovian/Markovian depth ratio.}
\label{tab:chunk_depth}

\centering
\small

\begin{tabular}{llcccccc}
\toprule
& & & \multicolumn{2}{c}{Depth $C$} & \multicolumn{3}{c}{KIVA (\%)} \\
\cmidrule(lr){4-5}
\cmidrule(lr){6-8}
Embodiment & Model & $\Delta$ & Mak. & Non-mak. & Mak. & Non-mak. & $\Delta$ \\
\midrule
Humanoid   & DreamGen  & 92 & 3 [2--3]     & 2 [2--3]     & 4.65  & 12.30 & $+7.65$ \\
Humanoid   & DreamDojo & 49 & 2 [1--14]    & 2 [1--29]    & 13.04 & 3.31  & $-9.73$ \\
Bimanual   & DreamDojo & 12 & 34 [10--124] & 41 [20--124] & 6.11  & 0.00  & $-6.11$ \\
Bimanual   & DreamGen  & 92 & 7 [2--18]    & 11 [3--18]   & 23.63 & 23.18 & $-0.45$ \\
Single-arm & DreamGen  & 92 & 3 [2--3]     & 5 [4--7]     & 59.40 & 70.65 & $+11.25$ \\
Single-arm & DreamDojo & 12 & 14 [11--15]  & 32 [16--48]  & 21.26 & 26.98 & $+5.72$ \\
\bottomrule
\end{tabular}

\end{table}

\begin{figure}[t]
\centering
\includegraphics[width=\linewidth]{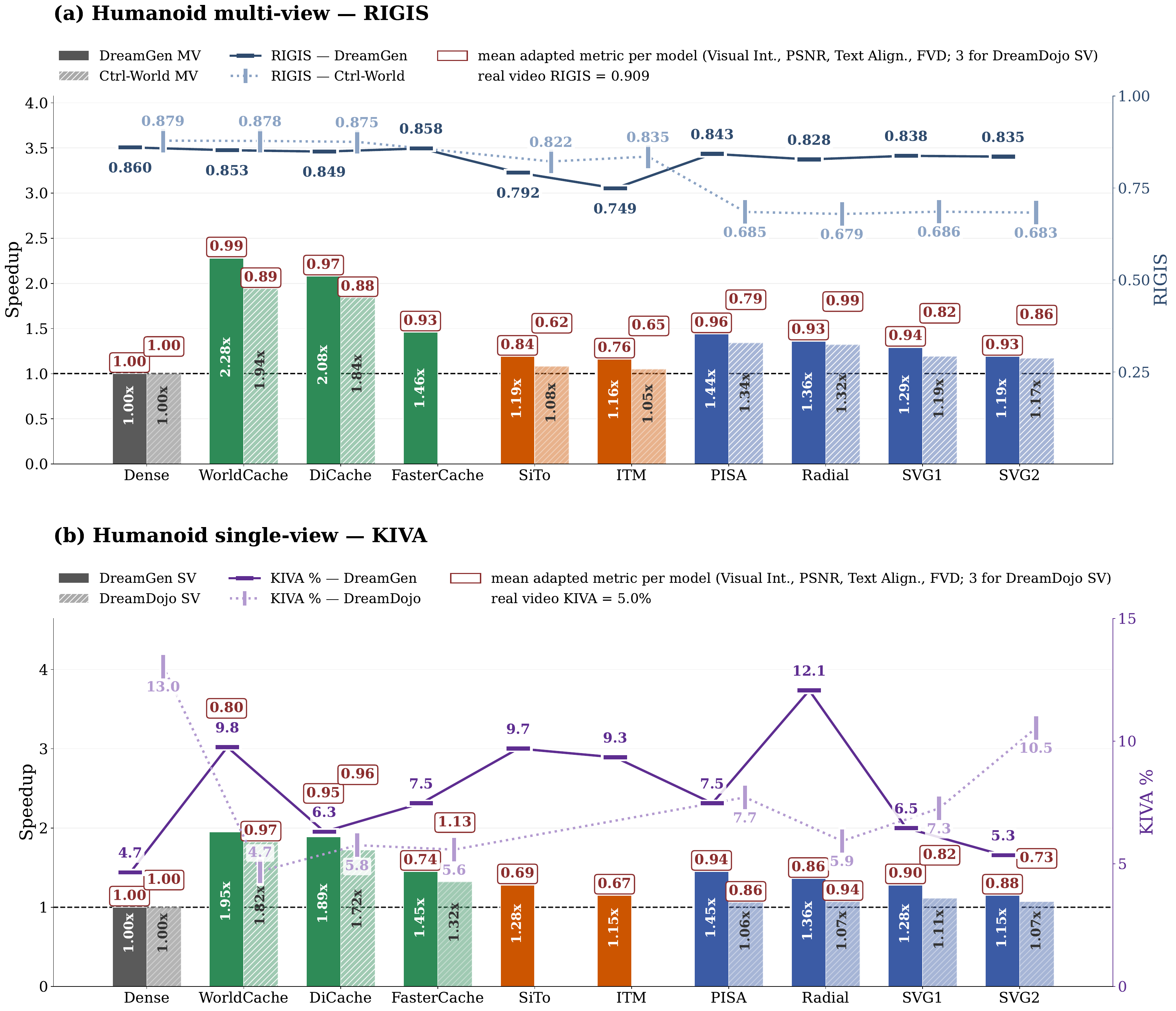}
\caption{\textbf{Acceleration trade-offs on humanoid morphology.}
Constructed as in Fig.~\ref{fig:accel_tradeoff}: bars show speedup over the
corresponding dense rollout, the middle axis reports the view-specific physical
audit, and boxed values denote aggregate retention $Q$ ($Q=1$ for dense).
On multi-view Ctrl-World, all sparse-attention methods reduce RIGIS from $0.879$
to $0.680$--$0.686$, whereas caching retains $0.875$--$0.878$ and token pruning
$0.822$--$0.835$. On single-view DreamDojo, all accelerated variants improve
KIVA relative to dense ($13.0\%$), with FasterCache reaching $Q=1.13$.}
\label{fig:supp_accel_tradeoff_humanoid}
\end{figure}

\begin{figure}[t]
\centering
\includegraphics[width=\linewidth]{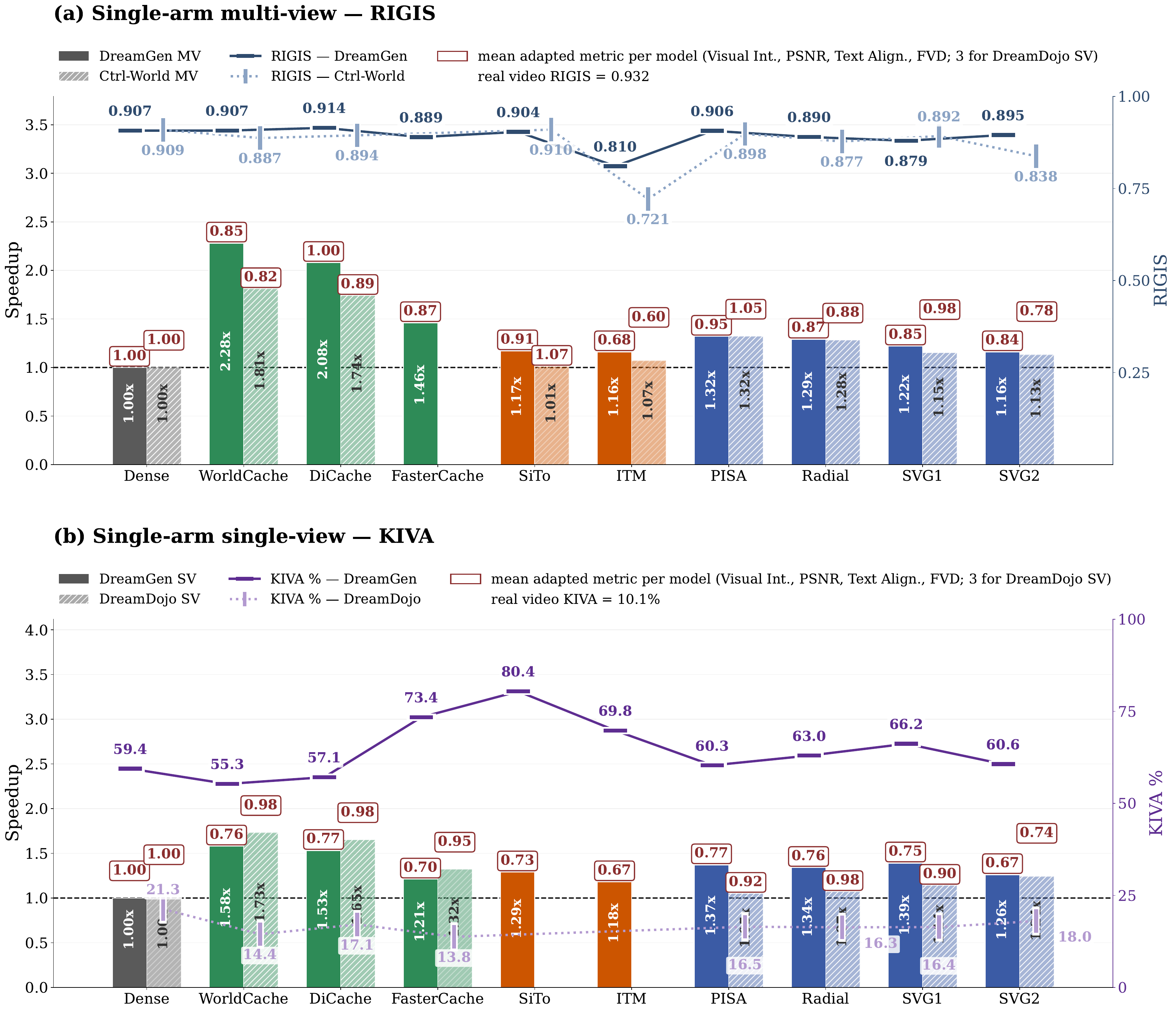}
\caption{\textbf{Acceleration trade-offs on single-arm morphology.}
Constructed as in Fig.~\ref{fig:accel_tradeoff}; see
Fig.~\ref{fig:supp_accel_tradeoff_humanoid} for the companion embodiment.
On multi-view, ITM produces the largest geometry degradation, reducing RIGIS to
$0.810$ on DreamGen and $0.721$ on Ctrl-World, compared with $0.838$--$0.915$
for the remaining methods. On single-view, the dense baseline already exhibits
high KIVA ($59.4\%$ versus $10.1\%$ for real video), so the $55$--$80\%$ range
observed under acceleration should not be interpreted as acceleration-induced
degradation. Missing entries indicate unmeasured configurations rather than zero.}
\label{fig:supp_accel_tradeoff_single_arm}
\end{figure}

\clearpage
\section{Human Assessment \& VLM Evaluation Protocols}
\label{supp:human_vlm}
This appendix provides comprehensive documentation of the evaluation protocols, dataset statistics, implementation setups, and extended empirical findings used to validate \textbf{KIVA}  and \textbf{RIGIS}. Specifically, Section \ref{supp:human_protocol} outlines the human annotation protocol, evaluated dataset preparation, evaluation rubrics, and inter-annotator consensus workflows. Section \ref{supp:vlm_config} details the automated VLM-as-a-judge infrastructure, including model configurations, multi-turn querying strategies, and prompt engineering templates. Finally, Sections \ref{supp:human_kiva} and \ref{supp:human_rigis} present supplementary results for KIVA and RIGIS, respectively, focusing on their alignment performance and correlation benchmarks against human evaluations.

\subsection{Human Assessment Protocol}
\label{supp:human_protocol}
\subsubsection{Prepared evaluation sets}
\label{supp:human_data}
To ensure a rigorous and unbiased assessment across diverse operational conditions, we construct two complementary evaluation corpora extracted systematically from \ours{}. Each set is carefully balanced across model quality tiers, robotic morphologies, and task dynamics.
Table~\ref{tab:supp_human_splits} summarizes the fine-grained data distribution across robot morphologies and task dynamics for both evaluation sets.

\begin{table}[htbp]
\centering
\caption{Data distribution of metric evaluation sets stratified by robot embodiment and task dynamics.}
\label{tab:supp_human_splits}
\subfloat[Embodied Physics Evaluation Set ($N{=}291$)]{%
\small
\begin{tabular}{lccc}
\toprule
Morphology & Markovian & Non-Markovian & Total \\
\midrule
Bimanual    & 50 & 50 & 100 \\
Humanoid    & 50 & 50 & 100 \\
Single-arm  & 50 & 41 & 91 \\
\midrule
\textbf{Total} & \textbf{150} & \textbf{141} & \textbf{291} \\
\bottomrule
\end{tabular}%
}%
\hspace{2em}%
\subfloat[Multi-view Consistency Evaluation Set ($N{=300}$)]{%
\small
\begin{tabular}{lccc}
\toprule
Morphology & Markovian & Non-Markovian & Total \\
\midrule
Bimanual    & 70 & 30 & 100 \\
Humanoid    & 43 & 57 & 100 \\
Single-arm  & 66 & 34 & 100 \\
\midrule
\textbf{Total} & \textbf{179} & \textbf{121} & \textbf{300} \\
\bottomrule
\end{tabular}%
}%
\end{table}

\subsubsection{Evaluation rubric \& scoring criteria}
\label{supp:human_rubric}

The assessment of kinematic physical plausibility and multi-view consistency demands distinct evaluation criteria and methodologies.

\paragraph{Embodied Physics Evaluation Rubric.}
The evaluation of embodied kinematic integrity is conducted at the segment level over 3-second windows; thus, a single episode involves multiple evaluation steps depending on its total duration. The detailed qualitative criteria, physical constraint requirements are provided in Box~\ref{box:supp_embodied_rubric}.

\begin{tcolorbox}[
  title={Embodied Physics Evaluation Rubric},
  colback=gray!4,
  colframe=black!40,
  fonttitle=\bfseries\small,
  breakable,
  boxrule=0.6pt,
  left=6pt, right=6pt, top=4pt, bottom=4pt,
]
\label{box:supp_embodied_rubric}
\small
You will watch robot video segments generated by a model and rate their hallucination level - artifacts that a video of a real robot would never contain. Robot quality covers whether the robot is deformed, blurred, color -bleeding, moving strangely, or moving in a way that violates ordinary physics.

Each video is split into short segments. Rate every segment on the scale:

\medskip
\noindent\textbf{Rating scale:} Good (3) - Medium (2) - Bad (1)
 
\medskip
\noindent\textbf{General principles.}
\begin{itemize}\setlength{\itemsep}{2pt}\setlength{\parsep}{0pt}
  \item \textbf{Evaluation focus.}
    Focus on the robot itself: Shape(is the robot deformed, stretched, broken apart, missing its arm/gripper, blurred, or dissolving?), Color / surface (any color bleeding, flickering, or abnormal color changes?), Motion (is it smooth and physically plausible like a real robot - or jittery, teleporting between positions, or bending joints in impossible ways?), Temporal consistency (is the robot in a later frame still the same robot as in earlier frames), Camera stability (ideally the camera is fully static; noticeable shaking or abnormal rotation/translation counts as an error.)
  \item \textbf{Rule for abnormal events.}
    Abnormal events happening to the robot itself count as errors, including but not limited to: the robot growing an extra arm or part; losing its arm/gripper; suddenly moving or jumping to another position; strange or non-physical behavior; deforming, disappearing / appearing suddenly. Only the robot is considered - not other objects in the frame.
  \item \textbf{What not to consider.}
    The robot's interaction with the environment (whether a grasp succeeds, whether the task is completed). Video compression quality or resolution.
\end{itemize}

\medskip
\noindent\textbf{Evaluation criteria.}
\begin{center}
\begin{tabular}{@{} p{1.5cm} p{10cm} @{}}
\toprule
\textbf{Rating} & \textbf{Multi-view Consistency \& Camera Stability} \\
\midrule
\textbf{Good} & 
Segment is sharp, robot arm moves with normal jitter, and the robot's action is clearly observable. Arm is not deformed. Robot is not blurry or smeared. \\
\midrule
\textbf{Medium} & 
Segment is relatively blurry but the robot arm is not overly deformed; or the robot arm shows slight deformation. Robot is still clearly recognizable and its action can still be followed. \\
\midrule
\textbf{Bad} & 
Segment is blurry or heavily smeared/warped; or the segment is sharp but the gripper is deformed; or the robot arm is heavily deformed; or strange components appear / the robot morphs into a different object. The action is hard or impossible to follow. \\
\bottomrule
\end{tabular}
\end{center}
\end{tcolorbox}

\paragraph{Multi-view Consistency Evaluation Rubric.}
The evaluation of multi-view consistency is conducted at the video-level, with the detailed qualitative criteria, camera stability requirements, and scoring thresholds provided in Box~\ref{box:supp_multiview_rubric}.

\begin{tcolorbox}[
  title={Multi-view Consistency Evaluation Rubric},
  colback=gray!4,
  colframe=black!40,
  fonttitle=\bfseries\small,
  breakable,
  boxrule=0.6pt,
  left=6pt, right=6pt, top=4pt, bottom=4pt,
]
\label{box:supp_multiview_rubric}
\small
You will evaluate the synchronization and consistency between two views of generated
videos depicting a robot performing actions. In addition to spatial consistency,
camera stability is also a factor (ideally, both camera views are completely static).

\medskip
\noindent\textbf{Rating scale:} Good (3) - Medium (2) - Bad (1)

\medskip
\noindent\textbf{General principles.}
\begin{itemize}\setlength{\itemsep}{2pt}\setlength{\parsep}{0pt}
  \item \textbf{Evaluation focus.}
    Consistency in motion, shape, color, and relative spatial position of
    objects/robot within the overlapping field of view visible to both camera
    angles; along with the stability of both views (ideally, both cameras are
    completely static).
  \item \textbf{Synchronized anomalies.}
    Synchronizations are evaluated \emph{without} regard to the semantic realism
    or physical plausibility of abnormal events (e.g., non-physical behaviors;
    objects deforming, appearing/disappearing, moving on their own).
    As long as these phenomena appear identically and simultaneously across both
    views, they are still considered consistent.
  \item \textbf{Out of scope.}
    Task completion rate, aesthetic appeal, and the accuracy or execution quality
    of the robot's actions are \emph{not} considered.
\end{itemize}

\medskip
\noindent\textbf{Evaluation criteria.}
\begin{center}
\begin{tabular}{@{} p{1.5cm} p{10cm} @{}}
\toprule
\textbf{Rating} & \textbf{Multi-view Consistency \& Camera Stability} \\
\midrule
\textbf{Good} & 
$\bullet$ Motion, shape, color, and relative spatial positioning across both views are virtually identical.\newline
$\bullet$ Camera is almost static. \\
\midrule
\textbf{Medium} & 
$\bullet$ Discrepancies appear between the two views (e.g., minor differences in motion, shape, or relative position), but both views still depict the same action.\newline
$\bullet$ Cameras may exhibit occasional jitter or small pans/drifts. \\
\midrule
\textbf{Bad} & 
$\bullet$ Inconsistencies are pronounced, severe, and persistent: the robots in the two views appear to perform different actions.\newline
$\bullet$ Cameras may shift, pan, or drift with large amplitudes. \\
\bottomrule
\end{tabular}
\end{center}
\end{tcolorbox}

\subsubsection{Annotation Interface \& Workflow}
\label{supp:human_ui}

\begin{figure*}[htbp]
  \centering
  \begin{subfigure}{0.8\textwidth}
    \centering
    \includegraphics[width=\textwidth]{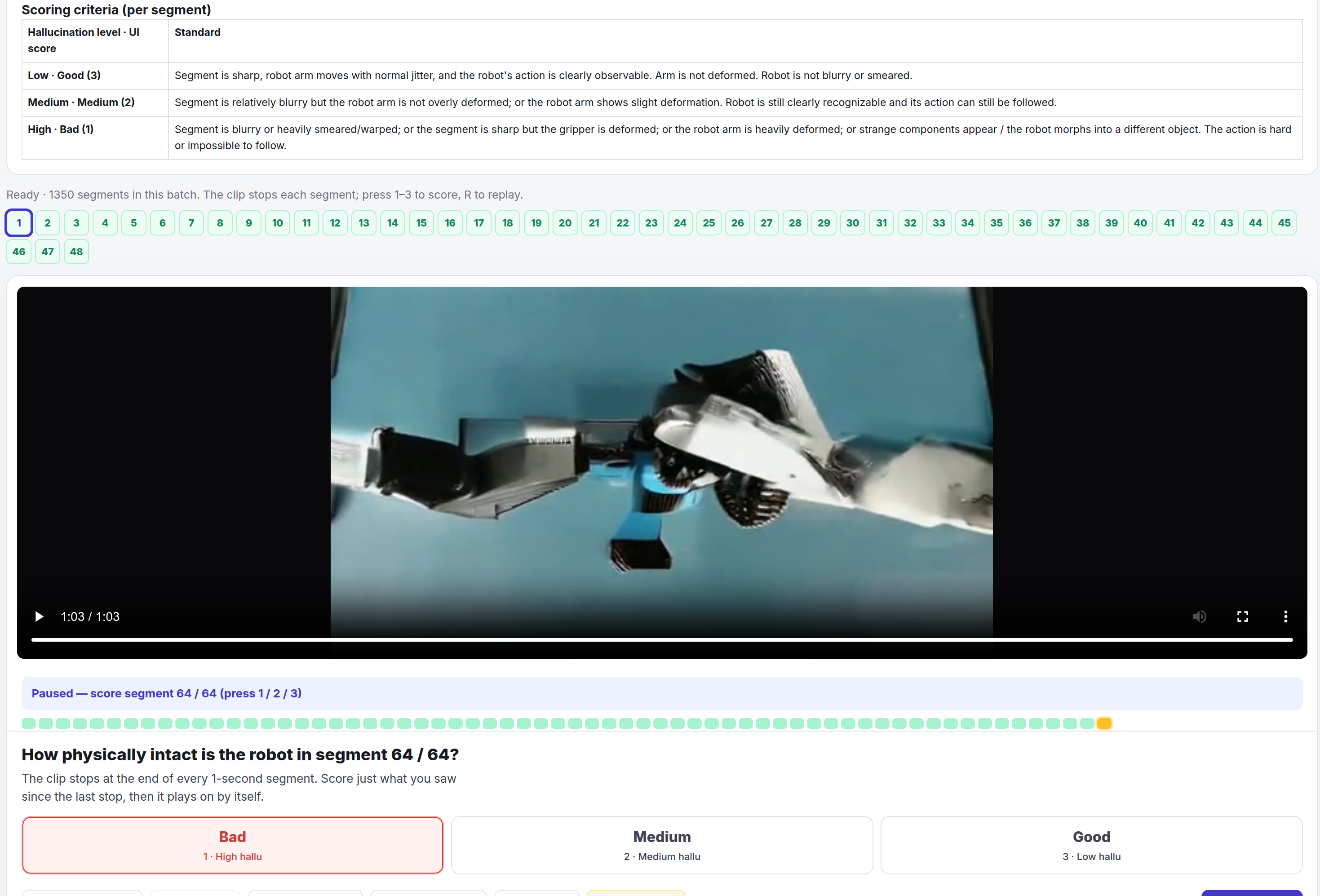}
    \caption{Embodied Physics Evaluation Interface.}
    \label{fig:supp_ui_kiva}
  \end{subfigure}
  
  \vspace{1.5em} 
  
  \begin{subfigure}{0.8\textwidth}
    \centering
    \includegraphics[width=\textwidth]{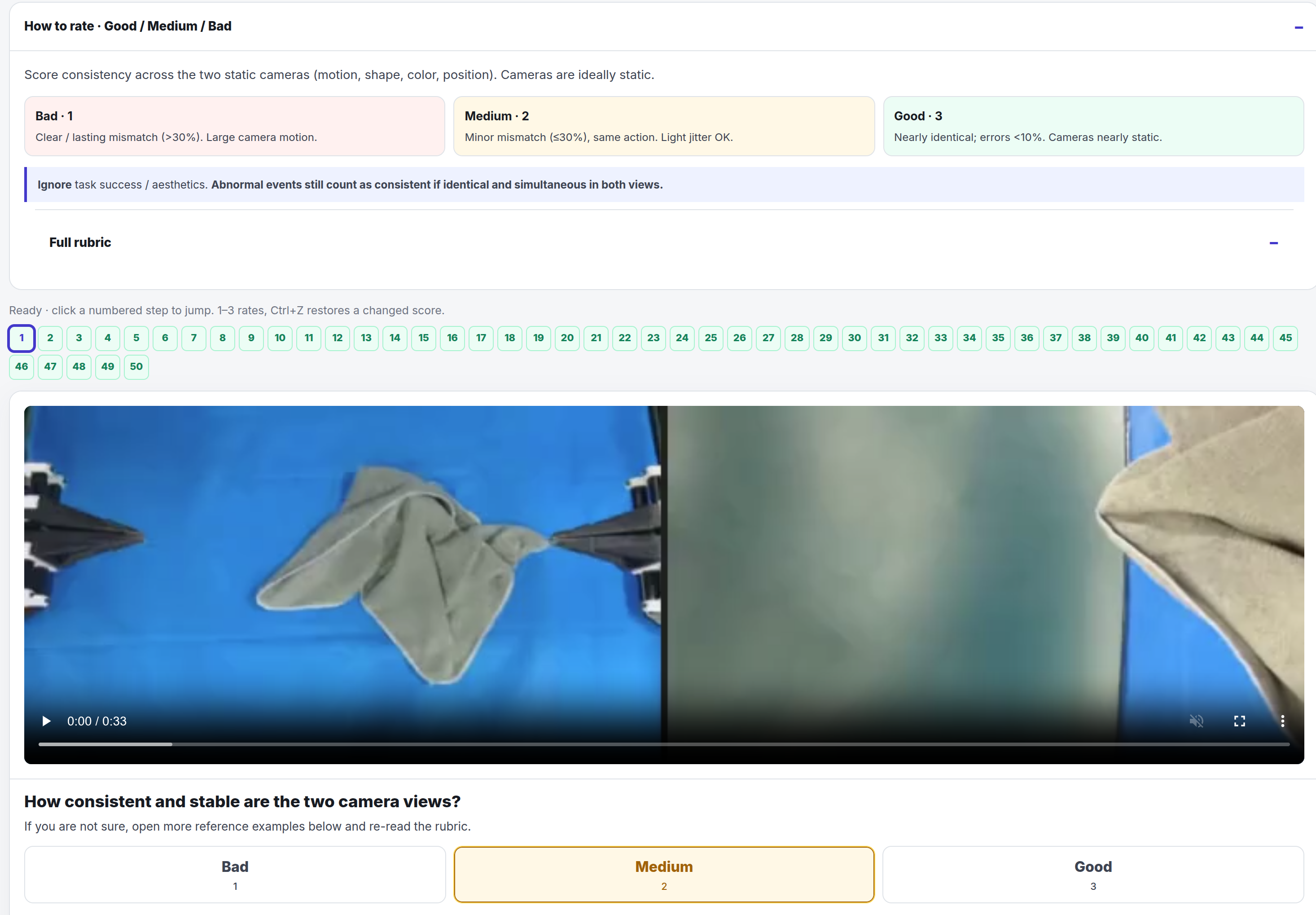}
    \caption{Multi-view Consistency Evaluation Interface.}
    \label{fig:supp_ui_rigis}
  \end{subfigure}
  
  \caption{\textbf{Human evaluation interfaces.} The web-based annotation dashboard showing tailored interfaces for multi-view consistency (side-by-side synchronized views with video-level scoring) and embodied physics (single-camera clips segmented into 3-second intervals with pause-and-prompt checks). Instructions, scoring rubrics, and reference examples remain persistently pinned on-screen to guide raters during evaluation.}
  \label{fig:supp_human_ui}
\end{figure*}

\paragraph{Annotation Interfaces.} Because the two evaluation criteria target different aspects of video quality, we designed tailored annotation interfaces optimized for each specific task:
\begin{itemize}
\item Multi-view Consistency: This interface displays two synchronized views side-by-side. Annotators review the entire video clip and assign a single video-level ordinal score using the buttons provided below.
\item Embodied Physics: This interface focuses on physical plausibility using a single camera view. The video is segmented into 3-second intervals where the playback pauses at the end of each segment to prompt the annotator for a score before proceeding to the next segment.
\end{itemize}
Before starting the evaluation, raters are provided with a video tutorial on scoring guidelines, accompanied by three curated example videos for each category combination appeared in Table~\ref{tab:supp_human_splits}. During the evaluation session, detailed rubrics and examples remain persistently visible on-screen as constant references. The detailed annotation interfaces for both metrics are captured in Fig.~\ref{fig:supp_human_ui}.

\paragraph{Adjudication.}
Each instance is rated by two independent scorers.
If the two scores agree, that value is the final label;
if they disagree, an expert checker assigns the final score.

\subsubsection{Human assessment results}
\label{supp:human_results}

\begin{figure*}[t]
  \centering

  \begin{subfigure}[b]{0.25\textwidth}
    \centering
    \includegraphics[width=\textwidth]{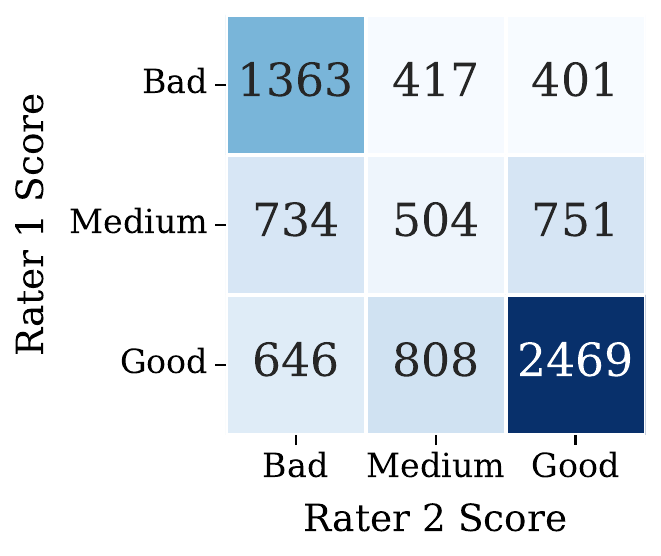}
    \caption{}
    \label{fig:supp_embodied_confusion}
  \end{subfigure}
  \hfill
  \begin{subfigure}[b]{0.25\textwidth}
    \centering
    \includegraphics[width=\textwidth]{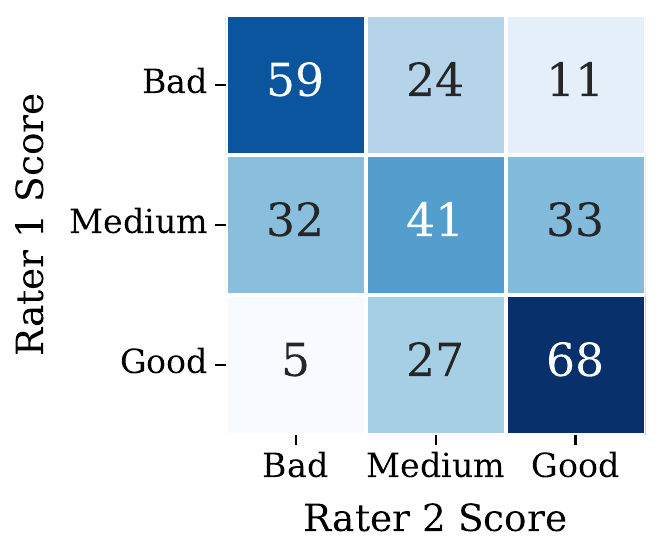}
    \caption{}
    \label{fig:supp_multiview_confusion}
  \end{subfigure}
  \hfill
  \begin{subfigure}[b]{0.48\textwidth}
    \centering
    \small
    \setlength{\tabcolsep}{4pt} 
    \begin{tabular}{lcc}
      \toprule
      Criterion & Instance & Agree (\%) \\
      \midrule
      Embodied Physics   & 8093 & 53.6\% \\ 
      Multi-view Consistency & 300 & 56.0\% \\
      \bottomrule
    \end{tabular}
    \vspace{2em} 
    \caption{}
    \label{tab:human_annotation_stats}
  \end{subfigure}

  \caption{\textbf{Human evaluation analysis.} 
  (a)~Embodied physics confusion matrix.
  (b)~Multi-view consistency confusion matrix. 
  (c)~Exact agreement rate between two raters across evaluation criteria.}
  \label{fig:supp_human_eval_combined}
\end{figure*}

Figure~\ref{fig:supp_human_eval_combined} reports the distribution of final human scores for multi-view consistency (video-level) and embodied physics (segment-level), reflecting how evaluators rated the generated robotic trajectories across criteria. To validate the reliability of these evaluations, Table~\ref{tab:human_annotation_stats} summarizes the agreement between the two primary raters including exact agreement rates and ordinal correlation metrics.

\subsection{VLM Judge Evaluation Setup}
\label{supp:vlm_setup}
While Vision-Language Models (VLMs) have been increasingly adopted as automated judges across general multimodal and video generation tasks~\citep{huang2024vbench,zheng2025vbench2,yang2026mirabench}, their capacity to reliably evaluate fine-grained physical aspects such as geometry and kinematics remains underexplored. To investigate this, we implement a VLM-based evaluation pipeline alongside human raters and our proposed metrics, revealing that current VLMs struggle significantly with capturing spatial-temporal nuances in multi-view consistency and embodied physics.

\subsubsection{VLM judge configuration and workflow}
\label{supp:vlm_config}

To comprehensively benchmark automated evaluation, we evaluate both closed-source and open-source Vision-Language Models, namely specifically Gemini 3.6 Flash, Qwen 3.6-Plus, Gemma 4 31B, and Qwen 3.6-35B-A3B, acting as automated judges via an \textit{OpenRouter}-compatible API. To maintain consistency with human evaluation, the VLMs assess the same evaluation split detailed in Appendix~\ref{supp:human_protocol} using a matching protocol.
For each evaluation variant (model $\times$ prompt type), every video is assessed in
\textbf{three independent turns} to mitigate evaluation variance and potential bias. 
Table~\ref{tab:supp_vlm_hparams} lists the detailed configuration. 

\begin{table}[t]
  \centering
  \small
  \caption{VLM judge configuration and model variants.}
  \label{tab:supp_vlm_hparams}
  \begin{tabular}{@{}ll@{}}
    \toprule
    Setting & Value \\
    \midrule
    API backend & OpenRouter \\
    Evaluated Models & \begin{tabular}[t]{@{}l@{}}
                       \textit{Closed-source:} Gemini 3.6 Flash, Qwen 3.6-Plus \\
                       \textit{Open-source:} Gemma 4 31B, Qwen 3.6-35B-A3B
                       \end{tabular} \\
    Prompt variants & Vanilla; +Chain-of-Thought \\
    Turns per video & 3 \\
    Aggregation & Majority vote over turns \\
    Score scale & \{Bad , Medium , Good\}  \\
    Input mode & \begin{tabular}[t]{@{}l@{}}
                 \textit{Multi-view Consistency:} Entire video \\
                 \textit{Embodied Physics:} Frames of a 3-second segment
                 \end{tabular} \\
    \bottomrule
  \end{tabular}
\end{table}

\subsubsection{Prompt design}
\label{supp:vlm_prompt}

The embedded rubric text is \emph{identical} to the human annotator guidelines at Box~\ref{box:supp_embodied_rubric} and \ref{box:supp_multiview_rubric}; we omit repeating it below. The \textbf{+CoT} variant differs only in the output instructions which is reasoning in \texttt{<think>\ldots</think>} before JSON.

\paragraph{System prompt (Vanilla).}

\begin{tcolorbox}[
  title={Embodied Physics - System Prompt (Vanilla)},
  colback=gray!4,
  colframe=black!35,
  fonttitle=\bfseries\small,
  breakable,
]
\footnotesize\ttfamily\raggedright
You watch one short segment of a robot video generated by a model and rate its hallucination level: artifacts that a video of a real robot would never contain. Robot quality covers whether the robot is deformed, blurred, color-bleeding, moving strangely, or moving in a way that violates ordinary physics.\\[0.4em]
Scale: low, medium, high. low is the best level (little to no deformation, the segment looks clean like a real video) and high is the most severe error level.\\[0.4em]
Focus on the robot itself:\\
-\, Shape: is the robot deformed, stretched, broken apart, missing its arm or gripper, blurred, or dissolving?\\
-\, Color and surface: any color bleeding, flickering, or abnormal color changes?\\
-\, Motion: is it smooth and physically plausible like a real robot, or jittery, teleporting between positions, or bending joints in impossible ways?\\
-\, Temporal consistency: is the robot in a later frame still the same robot as in earlier frames, with no missing details and no extra parts appearing?\\
-\, Camera stability: ideally the camera is fully static; noticeable shaking or abnormal rotation or translation counts as an error.\\[0.4em]
Abnormal events happening to the robot itself count as errors, including but not limited to: the robot growing an extra arm or part; losing its arm or gripper; suddenly moving or jumping to another position; strange or non-physical behavior; deforming, disappearing or appearing suddenly. Only the robot is considered, not other objects in the frame.\\[0.4em]
Do not consider:\\
-\, The robot's interaction with the environment (whether a grasp succeeds, whether the task is completed).\\
-\, Video compression quality or resolution.\\[0.4em]
Scoring criteria (per segment):\\
-\, low: Segment is sharp, robot arm moves with normal jitter, and the robot's action is clearly observable. Arm is not deformed. Robot is not blurry, not smeared.\\
-\, medium: Segment is relatively blurry but the robot arm is not overly deformed; or the robot arm shows slight deformation. Robot is still clearly recognizable and its action can still be followed.\\
-\, high: Segment is blurry, image appears smeared/warped (heavy distortion); or the segment is sharp but the gripper is deformed; or the robot arm is heavily deformed; or strange components appear / the robot morphs into a different object. The action is hard or impossible to follow.\\[0.4em]
Return strict JSON matching this schema:\\
\{\\
\phantom{xx}"evidence": "one sentence naming the visible robot artifact, or stating the robot looks clean",\\
\phantom{xx}"level": "low | medium | high"\\
\}\\[0.4em]
Output ONLY valid JSON. No markdown fences.
\end{tcolorbox}

\begin{tcolorbox}[
  title={Multi-view Consistency - System Prompt (Vanilla)},
  colback=gray!4,
  colframe=black!35,
  fonttitle=\bfseries\small,
  breakable,
]
\footnotesize\ttfamily\raggedright
You are an expert judge of multi-view consistency for generated robot videos.\\[0.4em]
You are given the FULL generated video. TWO camera views are placed SIDE BY SIDE
in every frame:\\
-\, LEFT half = View 1\\
-\, RIGHT half = View 2\\
Watch the entire clip for motion sync, shape/color/pose agreement, and camera stability.\\[0.4em]
Judge ONLY multi-view consistency and camera stability using the rubric below.\\
Do NOT judge task success, aesthetics, or physical realism of anomalies if they match across views.\\[0.4em]
=== RUBRIC ===\\
\emph{\rmfamily\upshape Identical to the human assessment rubric (Box~\ref{box:supp_multiview_rubric}).}\\
=== END RUBRIC ===\\[0.4em]
Map the rubric labels to integer scores as follows:\\
-\, Bad -> 1\\
-\, Medium -> 2\\
-\, Good -> 3\\[0.4em]
Respond with ONLY a compact JSON object (no markdown fences):\\
\{\\
\phantom{xx}"score": <integer 1, 2, or 3>,\\
\phantom{xx}"label": "<short label matching the score>",\\
\phantom{xx}"rationale": "<2-4 sentences citing what you observed across the video timeline>",\\
\phantom{xx}"error\_area\_pct\_estimate": <number 0-100 or null>,\\
\phantom{xx}"error\_duration\_pct\_estimate": <number 0-100 or null>\\
\}
\end{tcolorbox}

\paragraph{System prompt (+CoT).}
Same preamble, media blurb, rubric block, and score mapping as Vanilla.
Only the final output block is replaced by:

\begin{tcolorbox}[
  title={Embodied Physics - System Prompt (+CoT output block)},
  colback=gray!4,
  colframe=black!35,
  fonttitle=\bfseries\small,
  breakable,
]
\footnotesize\ttfamily\raggedright
\emph{\rmfamily\upshape The system prompt is byte-identical to Vanilla. CoT is enabled
through the decoding interface, not through prompt text:}\\[0.4em]
Open models: \texttt{chat\_template\_kwargs: \{enable\_thinking: true\}}, served by vLLM
with \texttt{-{}-reasoning-parser gemma4} (Gemma~4~31B) or \texttt{qwen3} (Qwen~3.6~35B-A3B).\\
Closed models: \texttt{reasoning: \{effort: high\}}.\\[0.4em]
The reasoning trace is returned in a separate \texttt{reasoning\_content} field, so the
message content still holds only the strict-JSON object
(\texttt{evidence}, \texttt{level}), decoded under the same JSON-schema constraint as Vanilla.
Budget: \texttt{max\_tokens} 2048 for CoT (256 for Vanilla).
\end{tcolorbox}

\begin{tcolorbox}[
  title={Multi-view Consistency - System Prompt (+CoT output block)},
  colback=gray!4,
  colframe=black!35,
  fonttitle=\bfseries\small,
  breakable,
]
\footnotesize\ttfamily\raggedright
Step 1. Think step by step about multi-view consistency and camera stability.\\
Consider: motion synchronization, shape/color/pose agreement, relative spatial positions
in the overlapping field of view, camera stability (jitter / pan / drift), and approximate
error area and duration relative to the rubric thresholds.\\
Do NOT judge task success, aesthetics, or physical realism of anomalies if they match across views.\\[0.4em]
Step 2. After your reasoning, output a compact JSON score report:\\
\emph{\rmfamily\upshape (same JSON schema as Vanilla; rationale cites observations
``across the video timeline''.)}\\[0.4em]
Wrap the reasoning between <think>...</think> so the final JSON is unambiguous.\\
Do not put markdown fences around the JSON.
\end{tcolorbox}

\paragraph{User message.}

\begin{tcolorbox}[
  title={Embodied Physics - User Message},
  colback=gray!4,
  colframe=black!35,
  fonttitle=\bfseries\small,
  breakable,
]
\footnotesize\ttfamily\raggedright
Rate this segment of \{robot\}: \{n\_frames\} frames at \{fps\} fps (\{duration\} s).\\[0.35em]
\emph{\rmfamily\upshape [16 JPEG frames of one segment, quality 95, sent as
\texttt{image\_url} data URLs before the text; \{robot\} is one of
``a two-arm robot'', ``a humanoid robot'', ``a single-arm robot'';
16 frames at 16 fps = 1.00 s]}
\end{tcolorbox}

\begin{tcolorbox}[
  title={Multi-view Consistency - User Message},
  colback=gray!4,
  colframe=black!35,
  fonttitle=\bfseries\small,
  breakable,
]
\footnotesize\ttfamily\raggedright
Evaluate this raw multi-view video file (\{video\_name\}).\\
Two views are merged horizontally (left=view1, right=view2).\\
Use only visible evidence from the video.\\[0.35em]
\emph{\rmfamily\upshape [video\_url: full side-by-side clip as data URL / file attach]}
\end{tcolorbox}

\subsubsection{VLM results summary}
\label{supp:vlm_results_summary}

Table~\ref{tab:vlm_eval_results} reports the self-consistency, internal score standard deviation, and average processing time per 1 video second of VLM judges across evaluation turns, where evaluation scores are mapped to a discrete numerical scale (Bad: 1, Medium: 2, Good: 3).

\begin{table}[t]
\centering
\small
\caption{\textbf{Agreement rate, score standard deviation, and processing latency of VLM judges.} \textit{Agree} denotes the percentage of videos where a model yields identical scores across evaluation turns, and \textit{Score Std} measures the average score standard deviation per video. Latency is measured as the average time in seconds required to process one video second. The bottom row reports the latency of our proposed RIGIS and KIVA metrics evaluated under identical settings.}
\label{tab:vlm_eval_results}
\begin{tabular}{lcccccc}
\hline
\multirow{2}{*}{Method} & \multicolumn{3}{c}{Embodied Physics} & \multicolumn{3}{c}{Multi-view Consistency} \\ \cline{2-7} 
                        & Agree (\%) $\uparrow$ & Score Std $\downarrow$ & Latency $\downarrow$ & Agree (\%) $\uparrow$ & Score Std $\downarrow$ & Latency $\downarrow$ \\ \hline
\multicolumn{7}{l}{\textit{Vanilla Zero-Shot}} \\ \hline
Gemini 3.6 Flash        & 87.2          & \textit{0.093}          & 2.781          & 63.0          & 0.240          & \textit{0.784} \\
Qwen 3.6 Plus           & \textbf{90.3} & \textbf{0.076} & 4.396          & \textbf{77.0} & \textbf{0.191} & 3.121 \\
Gemma 4 31B             & 84.9          & 0.097          & 2.000          & 57.7          & 0.291          & 1.054 \\
Qwen 3.6 35B            & 80.7          & 0.138          & \textit{0.750} & 38.7          & 0.471          & 1.154 \\ \hline
\multicolumn{7}{l}{\textit{Chain-of-Thought (CoT)}} \\ \hline
Gemini 3.6 Flash        & 84.9 & 0.112 & 4.631 & \textit{64.3} & \textit{0.227} & 0.804 \\
Qwen 3.6 Plus           & 63.2          & 0.252          & 17.440         & 33.7          & 0.591          & 2.818 \\
Gemma 4 31B             & 75.5          & 0.156          & 9.989          & 62.3          & 0.266          & 2.757 \\
Qwen 3.6 35B            & 38.3          & 0.344          & 15.438         & 33.0          & 0.507          & 1.110 \\ \hline 
\textit{Our Metrics}    & \multicolumn{2}{c}{\textbf{KIVA}} & \textbf{0.289} & \multicolumn{2}{c}{\textbf{RIGIS}} & \textbf{0.356} \\ \hline
\end{tabular}
\end{table}

As reported in Table~\ref{tab:vlm_eval_results}, the embodied physics evaluation reveals that vanilla zero-shot models establish high scoring reliability. Qwen 3.6 Plus achieves the highest agreement rate of 90.3 percent and the lowest score standard deviation of 0.076 while requiring a latency of 4.396 seconds. Implementing chain-of-thought prompting severely harms this performance across all evaluated models. For instance, Qwen 3.6 35B experiences an agreement drop to 38.3 percent and a score standard deviation increase to 0.344 while its processing latency surges to 15.438 seconds. In contrast, our proposed KIVA metric eliminates these generative overheads entirely and achieves a superior processing latency of 0.289 seconds.

For the multi-view consistency evaluation presented in Table~\ref{tab:vlm_eval_results}, the increased task difficulty leads to lower overall agreement rates and higher score standard deviations for baseline models. Qwen 3.6 Plus again leads the vanilla zero-shot configurations by securing a 77.0 percent agreement rate and a score standard deviation of 0.191. Chain-of-thought prompting introduces substantial instability in this setting as well, where models like Qwen 3.6 Plus drop to an agreement rate of 33.7 percent alongside a score standard deviation of 0.591. Meanwhile, our proposed RIGIS metric maintains high computational efficiency by processing video inputs in 0.356 seconds, outperforming all baseline alternatives by a wide margin.

\subsection{Supplementary Results for KIVA}
\label{supp:human_kiva}

\begin{figure}[t]
\centering
\includegraphics[width=\linewidth]{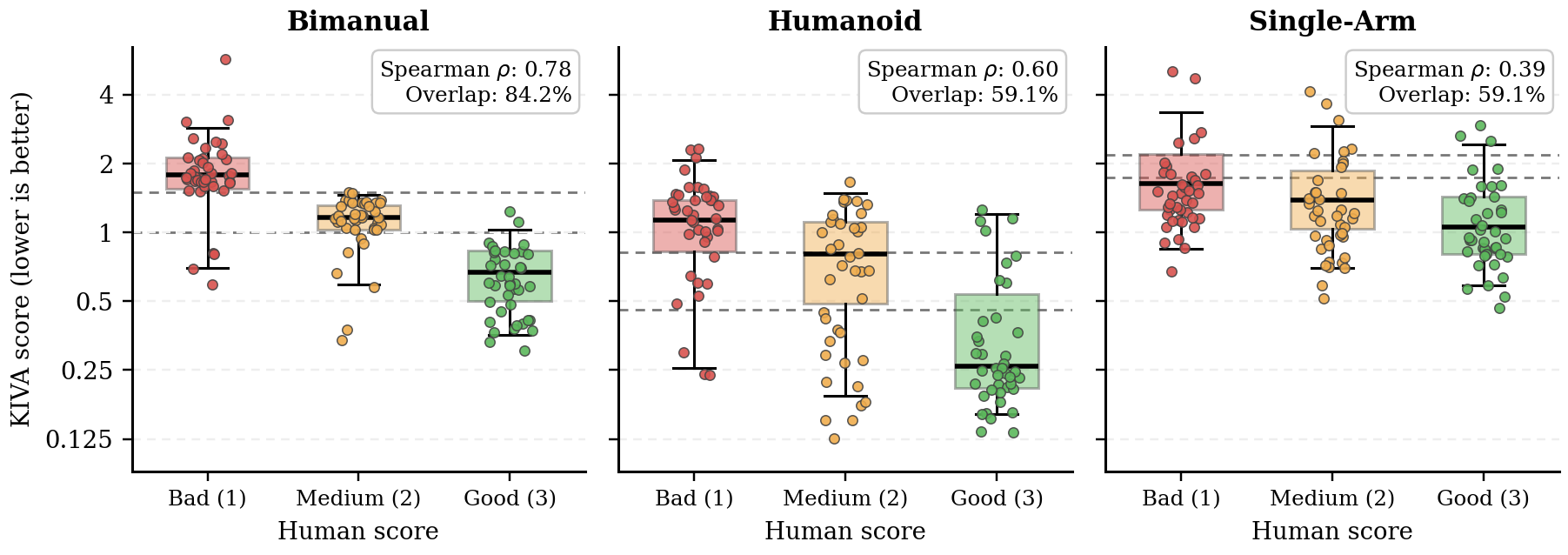}
\caption{\textbf{KIVA scores against human ratings.} Each panel shows KIVA's score (log scale, lower is better) for every labelled 16-frame segment, grouped by the human label. Boxes span the quartiles and whiskers the 5th to 95th percentile of all segments at that level; dots are 40 random segments per level. Dashed lines are the calibrated band cuts. Each panel reports Spearman's $\rho$ on the continuous score and the calibrated overlap.}
\label{fig:kiva_human}
\end{figure}

\begin{table}[t]
\centering
\small
\caption{\textbf{Agreement of KIVA and VLM judges with human labels across robot morphologies.} Spearman's $\rho$ is computed per 16-frame segment between the human label (Bad 1, Medium 2, Good 3) and each method's score: the mean level over three runs for a VLM, the continuous score for KIVA. Overlap is the percentage of segments whose three-level verdict matches the human label; a VLM's verdict is the majority over its runs. KIVA's two band cuts are calibrated on nine videos per embodiment (three per quality level) and applied to the remaining videos; its overlap is the mean over 1000 random draws of the calibration videos. Macro is the mean over the three embodiments; Micro pools all segments. Bold: best; italic: best VLM.}
\label{tab:kiva_vlm_human}
\begin{tabular}{llccccc}
\multicolumn{7}{c}{(a) Spearman correlation ($\rho\uparrow$)} \\
\toprule
Method & Strategy & Bimanual & Humanoid & Single Arm & Macro & Micro \\
\midrule
Gemini 3.6 Flash & \multirow{4}{*}{Vanilla} & \textit{0.465} & 0.396 & 0.565 & 0.475 & 0.480 \\
Qwen 3.6 Plus &  & 0.380 & 0.472 & 0.589 & 0.481 & 0.447 \\
Gemma 4 31B &  & 0.399 & 0.494 & 0.628 & 0.507 & 0.478 \\
Qwen 3.6 35B &  & 0.343 & 0.334 & 0.516 & 0.398 & 0.372 \\
\midrule
Gemini 3.6 Flash & \multirow{4}{*}{CoT} & 0.438 & 0.377 & 0.527 & 0.447 & 0.450 \\
Qwen 3.6 Plus &  & 0.423 & 0.460 & 0.624 & 0.502 & 0.489 \\
Gemma 4 31B &  & 0.438 & \textit{0.506} & \textbf{0.631} & \textit{0.525} & \textit{0.503} \\
Qwen 3.6 35B &  & 0.413 & 0.363 & 0.476 & 0.417 & 0.389 \\
\midrule
\textbf{KIVA (Ours)} &  & \textbf{0.781} & \textbf{0.603} & 0.390 & \textbf{0.591} & \textbf{0.573} \\
\bottomrule
\addlinespace[6pt]
\multicolumn{7}{c}{(b) Overlap (\%\,$\uparrow$)} \\
\toprule
Method & Strategy & Bimanual & Humanoid & Single Arm & Macro & Micro \\
\midrule
Gemini 3.6 Flash & \multirow{4}{*}{Vanilla} & 51.5 & 51.4 & 71.3 & 58.1 & 56.7 \\
Qwen 3.6 Plus &  & 49.4 & 55.4 & 72.3 & 59.1 & 56.5 \\
Gemma 4 31B &  & 50.8 & \textit{58.8} & \textbf{74.0} & \textit{61.2} & 58.2 \\
Qwen 3.6 35B &  & 45.0 & 47.7 & 67.6 & 53.5 & 51.5 \\
\midrule
Gemini 3.6 Flash & \multirow{4}{*}{CoT} & 51.3 & 51.8 & 69.9 & 57.7 & 56.3 \\
Qwen 3.6 Plus &  & 50.9 & 56.7 & 71.6 & 59.8 & 57.4 \\
Gemma 4 31B &  & \textit{52.0} & 57.4 & 72.8 & 60.7 & \textit{58.4} \\
Qwen 3.6 35B &  & 49.1 & 49.8 & 47.3 & 48.7 & 48.7 \\
\midrule
\textbf{KIVA (Ours)} &  & \textbf{84.2} & \textbf{59.1} & 59.1 & \textbf{67.5} & \textbf{73.5} \\
\bottomrule
\end{tabular}
\end{table}

\paragraph{Alignment with human ratings.} Fig.~\ref{fig:kiva_human} plots KIVA's score against the human label for each morphology. On all three the median score falls from Bad to Good. On Bimanual and Humanoid the levels are well separated and the calibrated cuts fall between them (Spearman $\rho$ of 0.78 and 0.60). On Single-Arm the medians still fall in order, but the three distributions overlap heavily and $\rho$ drops to 0.39.

\paragraph{Band calibration.} KIVA's score is continuous, and a three-level verdict needs two cuts on it. With the default cuts (1.0 and 1.5 times the calibrated threshold), the overlap with human labels is 85.5\%, 50.2\% and 46.8\% on Bimanual, Humanoid and Single-Arm. We refit the two cuts per morphology on nine labelled videos, three from each quality level, and score KIVA on the remaining videos. Averaged over 1000 random draws of the nine videos, the overlap on Humanoid rises from 50.2\% to 59.1\% and on Single-Arm from 46.8\% to 59.1\%, while Bimanual keeps its default cuts and moves from 85.5\% to 84.2\%. Macro overlap goes from 60.8\% to 67.5\% ($\pm$2.8 across draws) and micro overlap from 69.5\% to 73.5\% ($\pm$2.6). Calibration only moves the cuts, so it leaves Spearman's $\rho$ unchanged.

\paragraph{Comparison with VLM judges.} Table~\ref{tab:kiva_vlm_human} compares KIVA with the eight VLM configurations. On Bimanual, KIVA reaches $\rho = 0.781$ and 84.2\% overlap, against at most 0.465 and 52.0\% for a VLM. On Humanoid it has the highest Spearman (0.603 against 0.506) and matches the best VLM on overlap (59.1\% against 58.8\%). Over the three morphologies, KIVA reaches a macro $\rho$ of 0.591 and a macro overlap of 67.5\%, against 0.525 and 61.2\% for the best VLM; pooled over all segments, its overlap is 73.5\% against 58.4\%. Across the 1000 calibration draws, KIVA's micro overlap is above that of Gemma 4 31B in 99.5\% of draws. CoT prompting changes a VLM's macro Spearman by less than 0.03.

\paragraph{Implementation of Few-Shot VLM Prompting.}
To push the evaluation limits of vision-language models, we configure Gemma 4 31B with a structured few-shot prompting pipeline augmented by Chain-of-Thought (CoT) reasoning. The prompt establishes an expert roboticist persona, defines the $1\text{--}3$ embodied physics scale, and includes annotated video exemplars with ground-truth human scores, enforcing sequential reasoning for contact stability and physical plausibility before scoring. 


\subsection{Supplementary Results for RIGIS}
\label{supp:human_rigis}

\paragraph{Alignment with Human Ratings.}
\begin{figure*}[htb]
  \centering
  \includegraphics[width=\textwidth]{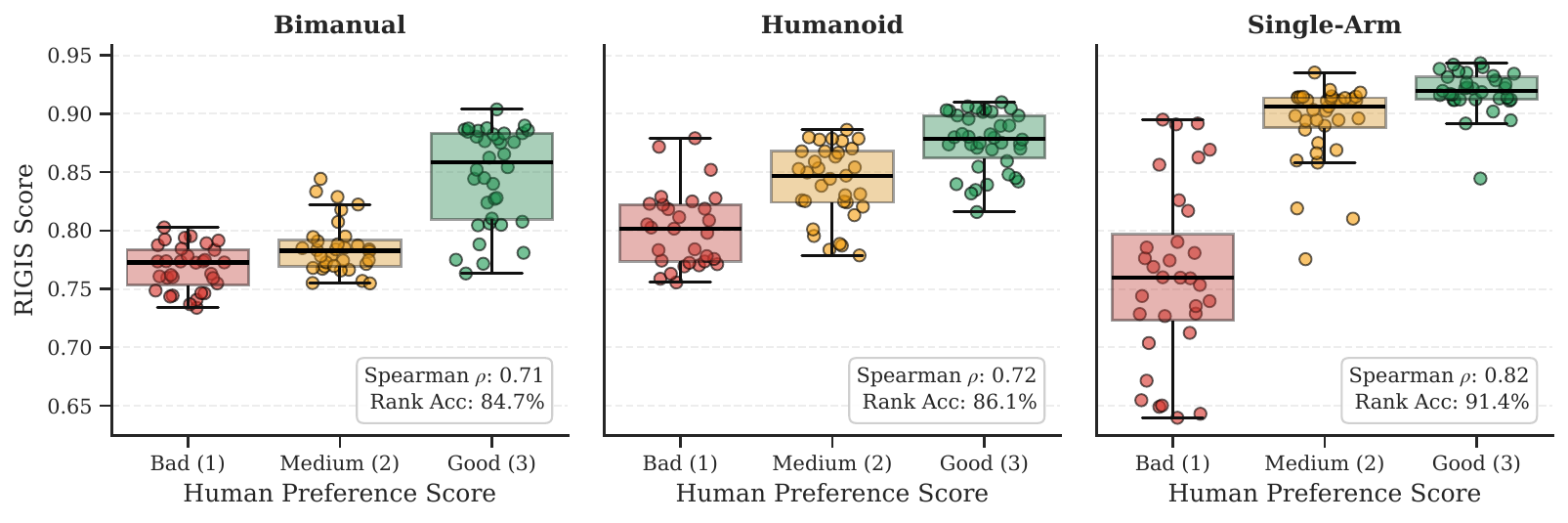}
  \caption{\textbf{Alignment of RIGIS scores with human preference ratings.} Distribution of RIGIS scores across human ratings for Bimanual, Humanoid, and Single-Arm morphologies.}
  \label{fig:supp_rigis_scatter}
\end{figure*}
To visually assess the agreement between RIGIS and human judgments, Fig.~\ref{fig:supp_rigis_scatter} illustrates the distribution of RIGIS scores across discrete human preference levels for three morphologies.
Across all types, RIGIS demonstrates a clear, monotonic alignment with human ratings. As depicted by the box and strip plots, the median RIGIS score exhibits a distinct, step-wise upward trend from rating $Bad$ through $Good$, accompanied by minimal interquartile range overlap between adjacent preference tiers.
Specifically, \textit{Single-Arm} systems present the most compact distributions and tightest point clustering.
For higher-DoF configurations such as \textit{Bimanual} and \textit{Humanoid} platforms, the vertical margins between rating tiers remain sharp and well-separated.
This distinct visual margin across all score levels confirms that RIGIS effectively evaluates multi-view consistency quality across diverse settings.

\paragraph{Comparison with VLM Judges.}
\begin{table}[htb]
\centering
\small
\caption{\textbf{Agreement of RIGIS and VLM judges with human labels across robot morphologies.} Spearman $\rho$ and pairwise rank accuracy $RankAcc$ are computed per embodiment between human labels and each method's scores, utilizing the majority verdict over three runs for VLMs and raw continuous scores for RIGIS, and subsequently averaged across embodiment. Pairwise rank accuracy specifically measures the fraction of human-ordered video pairs whose relative order is preserved by the evaluation method, with prediction ties treated as errors.}
\label{tab:supp_rigis_corr}

\begin{subtable}{\linewidth}
\centering
\caption{Spearman Correlation ($\rho \uparrow$)}
\label{tab:supp_rigis_spearman}
\begin{tabular}{lc cccc}
\toprule
Method & Strategy & Bimanual & Humanoid & Single Arm & Mean \\
\midrule
Gemini 3.6 Flash & \multirow{4}{*}{Vanilla} & \textit{0.379} & 0.154 & 0.288 & 0.274 \\
Qwen 3.6 Plus    &                          & 0.022          & 0.233 & 0.234 & 0.163 \\
Gemma 4 31B      &                          & 0.313          & 0.327 & 0.305 & 0.315 \\
Qwen 3.6 35B     &                          & 0.343          & 0.146 & 0.024 & 0.171 \\
\midrule
Gemini 3.6 Flash & \multirow{4}{*}{CoT}     & 0.183          & 0.055 & 0.252 & 0.163 \\
Qwen 3.6 Plus    &                          & 0.337          & 0.173 & 0.163 & 0.224 \\
Gemma 4 31B      &                          & 0.291          & 0.342 & \textit{0.344} & \textit{0.326} \\
Qwen 3.6 35B     &                          & 0.144          & \textit{0.359} & 0.298 & 0.267 \\
\midrule
\multicolumn{2}{l}{\textbf{RIGIS (Ours)}}   & \textbf{0.706} & \textbf{0.724} & \textbf{0.816} & \textbf{0.749} \\
\bottomrule
\end{tabular}
\end{subtable}

\vspace{0.6em}

\begin{subtable}{\linewidth}
\centering
\caption{Rank Accuracy ($\text{RankAcc} \uparrow$, \%)}
\label{tab:supp_rigis_rankacc}
\begin{tabular}{lc cccc}
\toprule
Method & Strategy & Bimanual & Humanoid & Single Arm & Mean \\
\midrule
Gemini 3.6 Flash & \multirow{4}{*}{Vanilla} & \textit{47.9} & 20.3 & \textit{42.4} & 36.9 \\
Qwen 3.6 Plus    &                          & 9.8          & 7.5 & 35.9          & 17.7 \\
Gemma 4 31B      &                          & 32.4          & 46.9 & 39.7          & 39.6 \\
Qwen 3.6 35B     &                          & 37.4          & 35.9 & 28.8          & 34.0 \\
\midrule
Gemini 3.6 Flash & \multirow{4}{*}{CoT}     & 39.4          & 21.7 & 40.2          & 33.8 \\
Qwen 3.6 Plus    &                          & 29.2          & 29.6 & 32.2          & 30.3 \\
Gemma 4 31B      &                          & 30.1          & 47.8 & 41.6          & 39.8 \\
Qwen 3.6 35B     &                          & 35.4          & \textit{50.3} & 40.2 & \textit{42.0} \\
\midrule
\multicolumn{2}{l}{\textbf{RIGIS (Ours)}}   & \textbf{84.7} & \textbf{86.2} & \textbf{91.4} & \textbf{87.4} \\
\bottomrule
\end{tabular}
\end{subtable}
\end{table}
As summarized in Table~\ref{tab:supp_rigis_corr}, off-the-shelf VLMs struggle significantly to evaluate robotic video generation quality, primarily due to an inherent inability to comprehend cross-view geometry consistency. Under the Vanilla zero-shot setting, all baseline models exhibit weak alignment with human preferences, with the highest-performing VLM (Gemma 4 31B) reaching a mean Spearman $\rho$ of only $0.315$ and a Rank Accuracy of $39.6\%$.
While prompting VLMs with CoT reasoning yields marginal gains for certain models, where Gemma 4 31B reaching $\rho = 0.326$ and Qwen 3.6 35B reaching $\text{RankAcc} = 42.0\%$, their overall performance remains fundamentally inadequate. This confirms that general-purpose VLMs lack the necessary spatial and geometric grounding to detect multi-view physical and structural inconsistencies across synthesized camera angles.

In contrast, RIGIS substantially outperforms all VLM judges across every morphology and metric. Overall, RIGIS achieves a mean Spearman $\rho$ of 0.749, representing a 138\% relative improvement over the best VLM baseline, Gemma 4 31B CoT, which reaches $0.326$. Furthermore, RIGIS obtains a mean Rank Accuracy of $87.4\%$, whereas the highest VLM result, achieved by Qwen 3.6 35B CoT, is only $42.0\%$.
Crucially, while VLMs exhibit severe performance instability across different robot configurations due to complex camera layouts and visual occlusions, as exemplified by Qwen 3.6 35B dropping to a Spearman $\rho$ of $0.024$ on Single Arm, RIGIS maintains uniformly high alignment across all embodiments, consistently achieving a Spearman $\rho$ of at least $0.706$ and a Rank Accuracy of at least $84.7\%$.

\begin{figure*}[htb]
  \centering
  \includegraphics[width=\textwidth]{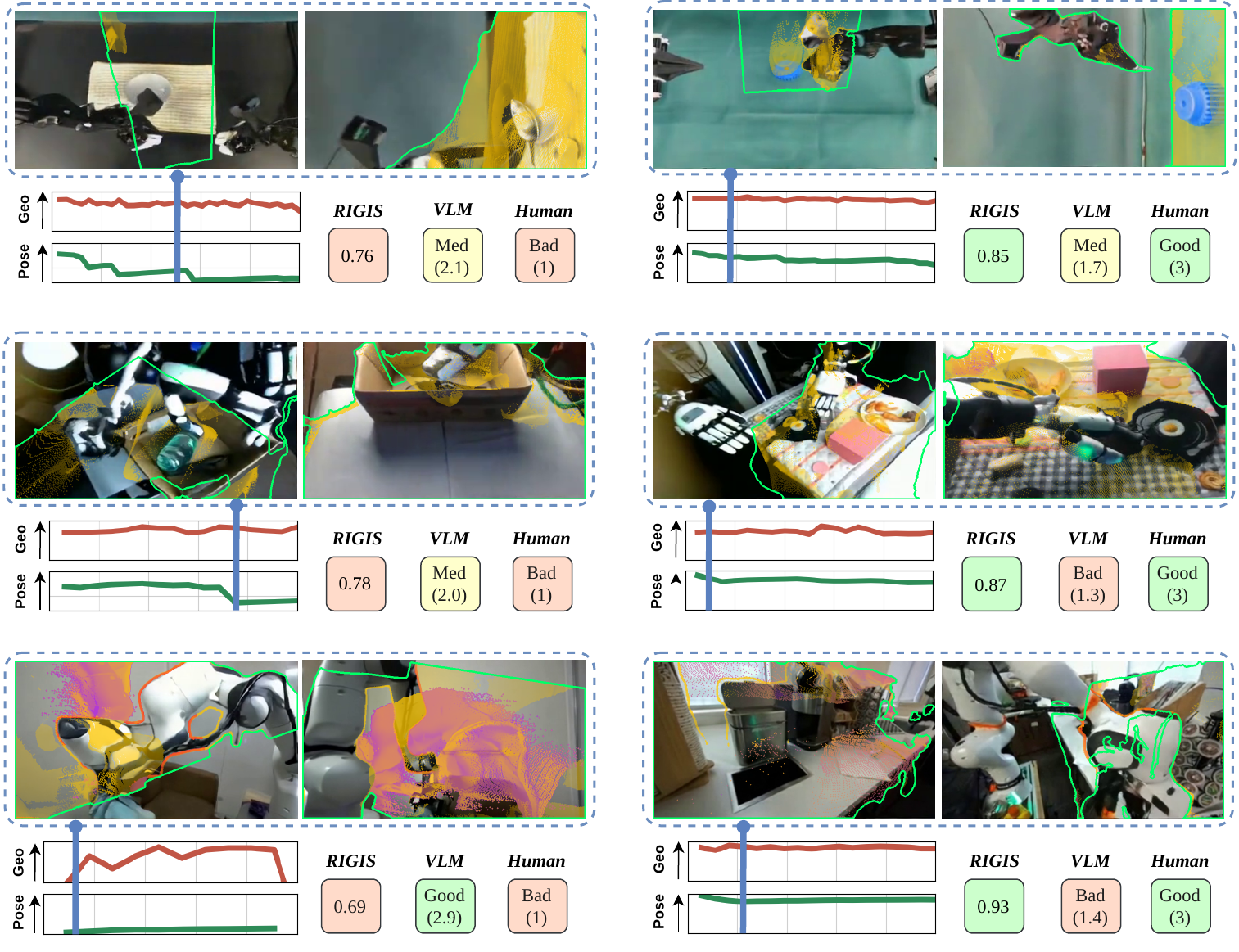}
  \caption{\textbf{Qualitative comparison of multi-view consistency evaluation between RIGIS, VLM judges, and human consensus.} Each panel displays synchronized multi-view video frames alongside temporal trajectories, where the synchronized trajectories denote the geometry and rig stability quality. The VLM's score is averaged across all methods, turns, and prompting strategies. These examples highlight notable discrepancies and failure modes where VLM judges misalign with human assessments and automated metrics.}
  \label{fig:supp_rigis_vlm_qualitative}
\end{figure*}

Fig.~\ref{fig:supp_rigis_vlm_qualitative} showcases qualitative failure modes of VLM judges compared to human evaluations and RIGIS. The synchronized trajectories reveal that VLMs fail to properly leverage geometric and rig stability cues, leading to significant score discrepancies relative to ground-truth human ratings. These findings demonstrate that standard VLMs are unequipped to reason about cross-view geometric fidelity, reinforcing the necessity of RIGIS as a dedicated, geometry-aware surrogate metric for robotic video world model evaluation.

\clearpage
\section{Using and Extending \ours}
\label{supp:extension}

\begin{figure}[t]
\centering
\includegraphics[width=\linewidth]{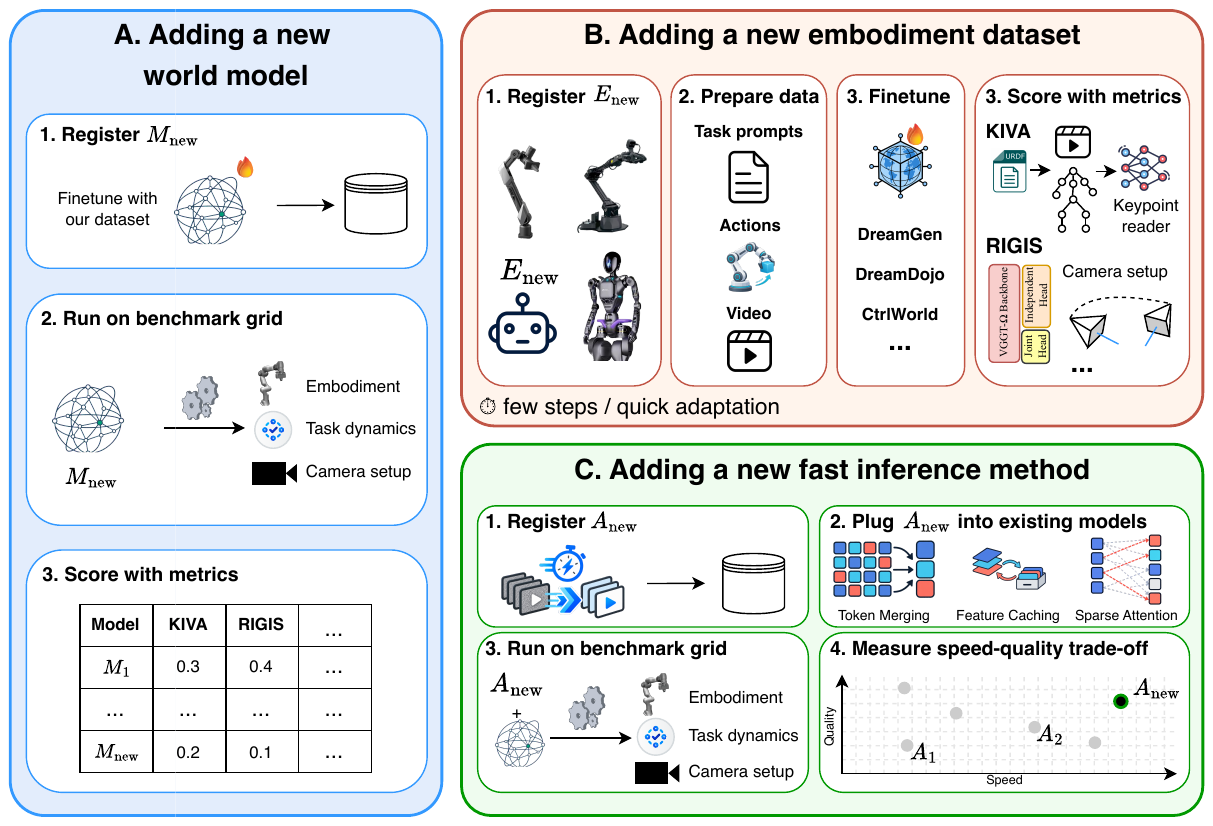}
\caption{\textbf{The three scaling dimensions for \ours.} Path A integrates a new world model $M_{\text{new}}$ fine-tuned on the released corpus, evaluated across the standard grid of embodiment, task dynamics, and camera setup, and scored with identical metrics to populate a new table row. Path B incorporates a new embodiment or dataset $E_{\text{new}}$ by adding a table column, registering camera mounts and task dynamics, formatting data into native representations, fine-tuning across three backbones, and instituting physical audits including a KIVA keypoint reader and a RIGIS camera setup. Path C introduces a new acceleration method $A_{\text{new}}$ as an intervention family applied to existing checkpoints without retraining, evaluated against dense rollouts on the standard grid, and plotted on the speed and quality plane alongside previously audited methods. Only Path B requires training, whereas Paths A and C reuse released episodes, conditioning, and checkpoints.}
\label{fig:supp_extension_flow}
\end{figure}

\ours{} supports extension under a standardized protocol across three dimensions. New world models or acceleration methods integrate into the existing evaluation grid, producing metrics directly comparable to current results without re-establishing evaluation assets. Conversely, a new embodiment or dataset establishes a new table column, requiring explicit declaration of physical invariants like camera placement and motion criteria before evaluation. Fig.~\ref{fig:supp_extension_flow} illustrates these dimensions, and Appendix~\ref{supp:extension_reporting} outlines the shared reporting checklist.

\subsection{Evaluating a New World Model}
\label{supp:extension_worldmodel}

\paragraph{What the benchmark requires of a model.}
\ours{} audits rollouts rather than architectures, requiring a minimal interface where a generator accepts a benchmark conditioning prefix and emits a frame-aligned RGB rollout at a declared frame rate. Because evaluation depends solely on this interface, structurally diverse models ranging from U-Nets to diffusion transformers are seamlessly compared. Consequently, models employing custom conditioning modalities remain fully admissible provided they satisfy the standard input and output specifications.

\paragraph{Conditioning and action interface.}
\ours{} releases three conditioning signals per episode, allowing models to use any subset:
\begin{itemize}
    \item \textbf{Observation prefix:} Contains conditioning frames at the declared resolution and frame rate.
    \item \textbf{Instruction string:} Verbatim text from the source corpus, scored in the Text Alignment column where text-ignoring models receive a \texttt{None} score rather than zero.
    \item \textbf{Proprioception:} Frame-aligned state and action vectors with synchronized indices. Action dimensionality is model-defined using released vectors truncated to standard $14$-D, $8$-D, or $7$-D interfaces, with custom truncation explicitly declared.
\end{itemize}
Two strict constraints apply: normalization must use fixed per-dimension statistics shared within each table cell, and conditioning prefixes must remain identical across compared methods to ensure valid paired attribution.

\paragraph{Fine-tuning is expected, not optional.}
Evaluating a released checkpoint zero-shot measures domain shift rather than acceleration. In-domain fine-tuning significantly improves performance, moving PSNR by $+5.94$ and $+6.76$\,dB on Ctrl-World configurations and by $+4.09$ and $+4.14$\,dB on DreamGen (see Appendix~\ref{supp:impl_finetuning}), whereas un-finetuned failures cause severe scene collapse (see Fig.~\ref{fig:supp_finetune_qual}). Because dense rollouts serve as the normalization reference for all acceleration results, an un-finetuned baseline invalidates retention scores. Consequently, new models must be fine-tuned on the released training corpus using the pipeline in Appendix~\ref{supp:dataset_processing} and report dense quality prior to acceleration. Unmatched fine-tuning recipes are fully permissible since the benchmark evaluates accelerators relative to the generator input.

\paragraph{Generating the rollouts.}
New models integrate by implementing the released three-stage pipeline consisting of data preparation, inference rollout, and execution wrappers. 

Follow these generation conventions:
\begin{itemize}
    \item Match rollout lengths to ground-truth episodes rather than fixed horizons to capture long-horizon behavior (Sec.~\ref{sec:result_rq4}).
    \item Discover episodes dynamically from annotation directories.
    \item Make execution passes resumable by skipping existing outputs.
    \item Mirror dense rollout subtrees with accelerated variants to enable straightforward paired comparisons keyed on episode paths.
\end{itemize}

\paragraph{Scoring and where the new model lands.}
Following rollout generation, evaluation proceeds uniformly without per-model customization by mapping clips to the standard analysis grid (Appendix~\ref{supp:implementation_details}): removing conditioning frames, striding to approximately $5$\,fps, cropping the target view, and resampling to $256\times192$ before computing metrics (Sec~\ref{supp:metrics}). This standardization ensures models generating at different native frame rates face no disadvantage, provided the declared rate is accurate. Furthermore, reference-free columns and physical audits require no dense counterpart, allowing new models to be evaluated on KIVA, RIGIS, and reference-free metrics independently of acceleration. Reference-based columns evaluate accelerators relative to the model's own dense rollout, rendering them inapplicable to standalone dense entrants. Finally, Visual Integrity remains barred from model ranking while serving in acceleration comparisons, and Text Alignment supports ranking claims only as a paired delta (Appendix~\ref{supp:metrics_other}).

\paragraph{Partial coverage is admissible; partial settings are not.}
New models need not cover the entire grid, as demonstrated by existing configurations spanning subsets. However, any covered cell must be evaluated completely across both task-dynamics regimes, all source cameras, and every curated episode, since partial evaluations compromise comparability. Furthermore, single-view FVD is referenced to each configuration's own ground truth rather than a shared pool, meaning new single-view models must report FVD against their own baseline and cannot be FVD-ranked against other single-view rows. Speedup measurements and physical audits carry no such restriction.

\subsection{Adding a New Embodiment or Dataset}
\label{supp:extension_embodiment}

\paragraph{Why this path costs more than the other two.}
Adding a world model or accelerator consumes benchmark invariants, whereas adding an embodiment supplies them. Physical audits are not generic appearance measures: KIVA verifies link length consistency and plausible jerk budgets for the embodiment, while RIGIS checks whether simultaneous multi-view feeds are consistent with a single camera rig geometry. Both audits require explicit domain-specific metadata including link connectivity, normal morphology motion bounds, and camera extrinsics. Because scoring relies on this closed-world assumption, registering a new embodiment requires declaring these invariants rather than inferring them at evaluation time. Consequently, embodiment extension includes a dedicated audit instantiation stage following preparation and fine-tuning, as shown in Fig.~\ref{fig:supp_extension_flow}-b.

\paragraph{What a source corpus must provide.}
\ours{} collects no new data, meaning a new domain integrates using the standard ingestion pipeline detailed in Appendix~\ref{supp:dataset_sources}. 

Ensure the source corpus satisfies these requirements:
\begin{itemize}
    \item Provide synchronized RGB video across all rig cameras with a declared frame rate and resolution, time-aligned per-frame proprioception, and a verbatim instruction string per episode.
    \item Maintain consistent camera counts and stability within each evaluation cell.
    \item Ensure episode lengths accommodate generator training windows, such as the $93$-frame requirement for DreamGen.
    \item Confirm corpora are redistributable or reproducibly obtainable, applying no success or minimum-length filters to allow episodes to span from $4$ to $3{,}000$ frames.
\end{itemize}

\paragraph{Mapping the task taxonomy and the dynamics regimes.}
Assign task taxonomies and dynamics regimes strictly before generating rollouts. Parse task instruction clauses and map leading verbs to canonical manipulation primitives, ensuring unmatched verbs are explicitly reported rather than silently bucketed across the $507$ instructions and $14$ primitives detailed in Appendix~\ref{supp:dataset_taxonomy}. Classify tasks as non-Markovian when correct continuation requires historical context invisible in the current frame (Appendix~\ref{supp:dataset_sources}). Assign this regime label from the task definition prior to rollout generation to prevent model behavior from biasing the classification and to preserve axis integrity, as these regimes dictate rollout depth (Appendix~\ref{supp:metrics_chunk_depth}).

\paragraph{Declaring the camera setup.}
While video geometry models such as VGGT-$\Omega$ \citep{wang2026vggtomega} enable camera pose estimation, allowing RIGIS to operate on generated videos without rigid camera assumptions, providing explicit camera parameters is strongly recommended. Supplying these ground-truth calibration profiles establishes an optimal reference for multi-view consistency evaluation.

\paragraph{Preprocessing.}
Follow the preparation in Appendix~\ref{supp:dataset_processing} while retaining four inherited decisions:
\begin{itemize}
    \item Resize frames to native training resolutions via bilinear resampling without letterboxing or aspect ratio preservation, restricting cross-cell metric comparisons.
    \item Apply integer-striding temporal resampling to match model training rates, measuring realized rates directly on generated clips.
    \item Compose multi-view settings into a single canvas where unused slots are filled with black frames, marked inactive in the prompt, and excluded from all metrics.
    \item Compute proprioception dimensions and per-dimension percentile normalization from prepared episodes to establish fixed cell-wide normalization parameters, and drop episodes missing required streams rather than padding them.
\end{itemize}

\paragraph{Calibration that must be redone rather than inherited.}
Never reuse another morphology's thresholds, as doing so misattributes structural differences as hallucination rates. Calibrate KIVA violation thresholds specifically for the new source corpus using clip-level band cuts of $1.00$ and $1.50$ derived from the $95$th percentile of real data rather than human-fitted values (Appendix~\ref{supp:metrics_hallucination}). 

Follow these calibration requirements:
\begin{itemize}
    \item Perform dedicated real-segment calibration passes before scoring generated segments.
    \item Implement a morphology-specific keypoint reader for rigidity and jerk computations (Appendix~\ref{supp:metrics_hallucination}), as explicitly named in Fig.~\ref{fig:supp_extension_flow}-b.
    \item Provide co-located real-video anchors for RIGIS and a real video pool keyed by embodiment, camera, and regime for FVD, validated via split-half exchangeability checks (Appendix~\ref{supp:metric_fvd}).
\end{itemize}
Avoid human-label fitting; optimizing thresholds for raw agreement severely distorts class balance, reducing minority class emissions from $18.6\%$ to $3.0\%$ and yielding a worse detector despite superficially improved scores.

\paragraph{What quick adaptation does and does not mean.}
Recognize that the minimal fine-tuning stage designated in Fig.~\ref{fig:supp_extension_flow}-b is relative. Adapting the three supported backbones to a new domain is inexpensive only because training recipes are publicly released, use full fine-tunes without secondary adapters, and share the preparation pipeline. True effort concentrates in registration and calibration rather than training, as the keypoint reader, real-segment thresholds, mount declarations, and FVD pool must be established before scoring any generated clip.
\subsection{Evaluating a New Acceleration Method}
\label{supp:extension_acceleration}

\paragraph{Registration.}
This is the cheapest extension because it reuses existing episodes, conditioning, and checkpoints, making acceleration training-free and focused entirely on inference. Each method follows a simple pattern: a configuration module with an argument group and a builder that creates the runtime setup. Inference scripts forward these arguments, so methods are enabled via flags without modifying model code or forking architectures. 

\paragraph{The matched dense baseline.}
Always compute speedups against a dense baseline re-run inside the same cell, on identical episodes, and under the same configuration. Never use stored constants or pre-computed summaries for the $1.00\times$ reference. 

To ensure a valid comparison:
\begin{itemize}
    \item Execute the identical command line with the method flags disabled using the built-in toggle switch.
    \item Verify that control and candidate runs differ only by the intervention to eliminate configuration drift and sampling noise.
\end{itemize}

\paragraph{Tuning to the knee, and where the sweep output goes.}
Benchmark methods at their performance knee on each cell rather than using default author settings. The knee is the most aggressive setting where quality remains indistinguishable from the cell dense rollout.

Follow these steps for knee tuning:
\begin{itemize}
    \item Vary the primary quality-speed knob while holding warm-up parameters constant.
    \item Score each point using PSNR, SSIM, and LPIPS against the dense rollout of the same episode.
    \item Inspect the final frame manually to detect multi-view smearing, compression artifacts, and metric collapse points where washed-out frames artificially outscore structurally incorrect high-frequency details.
    \item Publish the complete curve and configurations that caused visible degradation alongside the final shipped configuration.
    \item Write all sweep outputs to scratch directories rather than the canonical rollout tree to prevent mixing tuning artifacts with released rollouts, and record the shipped configuration per cell rather than per method.
\end{itemize}

\paragraph{Per-cell tuning is mandatory, and the reason is measured.}
Never tune a method once and transfer the settings across settings. Backbone dispersion causes a median absolute disagreement of $0.102$ in aggregate retention across accelerator-cell pairs, with nearly half of the pairs falling on opposite sides of the quality threshold ($Q=0.95$). 

Follow this rule:
\begin{itemize}
    \item Perform tuning independently for every cell because optimal operating points depend entirely on backbone runtime composition and sequence length.
    \item Avoid global defaults, which lead to mis-tuned configurations that falsely classify methods as lossless on one backbone while causing visible damage on another (see Appendix~\ref{supp:impl_acceleration}).
\end{itemize}


\paragraph{Timing protocol.}
Measure all speedups strictly sequentially on a single idle A100-SXM4-80GB GPU per run, with context parallelism disabled. 

Follow these measurement instructions:
\begin{itemize}
    \item Refuse execution on any busy GPU card to prevent co-tenancy overheads from inflating generation times by $15\%$ to $20\%$.
    \item Read generation times exclusively from each family's authoritative log line rather than process wall clock, avoiding distortions from model loading, video muxing, or metric computations.
    \item Sample each cell using episodes from both regimes to account for length-dependent savings.
    \item Report initial segment performance separately from the steady-state ratio to capture one-off warm-up costs, such as attention profiling or $k$-means initialization, preventing distortions in the baseline comparison.
\end{itemize}

\paragraph{Reporting the quality cost as paired deltas.}
Report every quality claim as a paired delta representing the accelerated clip minus its own dense clip on the same episode, view, and matched frame grid. 

Follow these reporting instructions:
\begin{itemize}
    \item Provide the measured speedup alongside $\Delta$KIVA (in percentage points of invalid segments), $\Delta$RIGIS (for multi-view settings), and paired deltas for the four reference-free columns.
    \item Include the valid sample size ($n$), exclusion rate, and ensure the claim gate from Appendix~\ref{supp:extension_reporting} is applied.
    \item Use the aggregate retention $Q$ defined in Appendix~\ref{supp:impl_acceleration} solely as a ranking statistic, avoiding it for effect sizes due to its compression factor of $2.8$ to $3.2$.
    \item Recognize that audit improvements under acceleration do not indicate superior rollouts, and account for baseline divergence where high dense hallucination rates (e.g., $59.4\%$ KIVA versus a $10.1\%$ real-video baseline) constrain the interpretation of accelerated values.
\end{itemize}

\paragraph{Negative results are results.}
Report all negative results, including methods that fail to accelerate and fall below $1\times$ speed, rather than omitting them or replacing them with stronger alternatives from the same family. 

Follow these reporting rules:
\begin{itemize}
    \item Retain negative rows (such as sparse attention on the Ctrl-World U-Net or slower-than-dense configurations on DreamDojo) to preserve full benchmark information.
    \item Distinguish clearly between measured failures and unmeasured configurations.
    \item Never interpolate missing cell values from neighboring configurations.
\end{itemize}

\subsection{Standardised Evaluation and Reporting Protocol}
\label{supp:extension_reporting}

\paragraph{The claim gate.}
Apply the claim gate to convert any measurement into a formal statement across all extension paths. Aggregate rollout values for a cell as a median with a $95\%$ bootstrap confidence interval within a single stratum (model, embodiment, camera mount, and regime), reporting both the valid sample size $n$ and the exclusion rate.

Follow these verification requirements:
\begin{itemize}
    \item Promote a difference to a valid claim only when the confidence interval strictly excludes zero, the Wilcoxon signed-rank test yields $p < 0.05$, and $n \ge 8$. 
    \item Report any cell with $n < 8$ as untested rather than showing no effect.
    \item Ensure paired deltas originate from the exact same measurement rather than merely the same episode, requiring identical source frame rates, effective frame rates, and lengths matching within $10\%$.
\end{itemize}
This strict gate is mandatory; for instance, it successfully rejected $226$ out of $1{,}836$ acceleration pairs, including an entire cell where differing frame rates and lengths between dense and accelerated rollouts manufactured spurious continuity gains (Sec.~\ref{sec:evaluation_protocol}).

\paragraph{Reporting checklist.}
A submission to \ours{} must include the items listed below, grouped by extension path with a shared block applying to all submissions.

\begin{itemize}
\item \textbf{Shared.} (i) Covered grid settings, including task-dynamics regimes, all rig cameras, and episode counts compared against curated cell totals. (ii) Unchanged analysis grid confirmation (conditioning frames dropped, stride adjusted to $\approx\!5$\,fps, requested view cropped, resampled to $256\times192$, and clips under three surviving frames excluded from scoring). (iii) Per-cell median, $95\%$ bootstrap confidence interval, valid $n$, and exclusion rate for every reported column, ensuring the claim gate is applied and untested settings are explicitly labeled. (iv) Unmeasurable quantities recorded as \texttt{None} rather than $0$ to prevent conflating missing data with zero motion. (v) Strict adherence to per-column restrictions (no model ranking on Visual Integrity, no unpaired Text Alignment median, and no per-cell claims on FVD where effective $n=1$). (vi) Clear identification of single-sample values versus means.

\item \textbf{New world model.} (vii) Checkpoint identifiers per cell, fine-tuning recipes, training corpora, and disclosures of any label or frame-rate defects. (viii) Declared frame rates, resolutions, conditioning-prefix lengths, action truncations, and fixed normalizers. (ix) Pre-accelerator dense quality per cell as the foundational reference. (x) Seeding details and single-sample indicators. (xi) Self-referenced FVD for single-view settings without cross-configuration ranking.

\item \textbf{New embodiment or dataset.} (xii) Corpus provenance, licenses, and reproducible acquisition paths. (xiii) Per-camera mount classes, body-relative extrinsics, and world-rigidity flags. (xiv) Pre-assigned task-dynamics regime labels. (xv) Taxonomy mappings with unmapped verbs reported. (xvi) Realized frame rates, counts, and durations measured directly on generated clips. (xvii) Morphology-specific KIVA keypoint readers and un-fitted real-segment threshold calibrations. (xviii) RIGIS camera-set declarations or explicit cross-view limitations. (xix) FVD real pools and split-half exchangeability validation.

\item \textbf{New acceleration method.} (xx) Intervention families, full per-cell shipped configurations, warm-up parameters, and execution environments. (xxi) Complete sweep curves including degraded configurations, or explicit statements regarding inherited untuned settings. (xxii) Matched dense controls executed via identical command lines with method flags disabled on identical episodes. (xxiii) Timing conditions (idle single GPU, disabled context parallelism, authoritative log lines, dual-regime sampling, and separated warm-up versus steady-state ratios). (xxiv) Paired quality deltas with $\Delta$KIVA, $\Delta$RIGIS, and reference-free columns. (xxv) Retained negative speedups and clear distinctions between unmeasured and failing configurations.
\end{itemize}

\paragraph{Two reporting surfaces, kept separate.}
Preserve the strict separation between the two reporting surfaces utilized throughout this benchmark. Absolute per-configuration medians describe a configuration against its own dense baseline and must never be compared vertically down a column across different configurations. Paired deltas carry no absolute level and serve as the exclusive surface for ranking acceleration methods. Maintaining both surfaces requires no extra scoring effort, as pairing is performed via a simple join and subtraction at table time, ensuring new methods remain immediately comparable to existing benchmark entries without re-scoring.

\subsection{Additional Use Cases of RoboWorm}
\label{supp:extension_usecases_roboworm}
RoboWorM serves as a critical filter for identifying high-fidelity video world models that can be reliably deployed across diverse downstream robotic pipelines. High-ranking models can function as robust interactive simulators, e.g., Ctrl-World~\citep{guo2026ctrl}, to \textbf{(i) support non-Markovian task execution}, providing the temporal consistency required for object-centric memory states, semantic-graph planning~\citep{vo2026codegraphvlp,hanyu2026slotvla,chung2026rethinking}. Additionally, these models facilitate robust policy adaptation by \textbf{(ii) synthesizing novel, high-quality rollout videos} from sparse demonstrations and text prompts to support few-shot frameworks~\citep{nguyen2026foca,ha2026equivla}. For post-training phases, top-tier models provide the envisioned trajectories needed to align generative control policies through progress-guided preference generation~\citep{luu2026progress} and rolling imagination frameworks~\citep{zhou2026rolling}.

When extended to multi-view generation, these ranked models can \textbf{(iii) synthesize 3D demonstrations to train spatial policies} like PointWorld~\citep{huang2026pointworld}; however, managing this high-dimensional data requires strict architectural optimization, such as minimizing necessary point cloud tokens~\citep{tran2026many}, restricting fine-tuning to essential layers~\citep{nguyen2026finetuning,liu2025vla}, or speed-up techniques~\citep{tran2024accelerating,zhao2026faster,nguyen2026vla} to maintain viable inference speeds. Finally, deploying these filtered models in (\textbf{iv) highly cluttered environments}~\citep{nguyen2026glovla,liu2026hsc,vo2025clutter} necessitates coupling the generated visuals with explicit geometric grounding and automated scene generation (e.g., SimFoundry~\citep{ranawaka2026simfoundry}) to ensure object-centric spatial reasoning remains accurate under occlusion.


\end{document}